\documentclass[Afour,sageh,times]{sagej}
\usepackage[table]{xcolor}
\usepackage{graphicx}
\usepackage{caption}
\usepackage{subcaption}
\usepackage{moreverb,url}
\usepackage{xcolor}
\usepackage[colorlinks,bookmarksopen,bookmarksnumbered,citecolor=red,urlcolor=red]{hyperref}
\usepackage{bm}
\usepackage{algorithm}
\usepackage[noend]{algpseudocode}
\usepackage{dsfont} 
\usepackage{array}
\usepackage{adjustbox}

\newcommand{\ciwn}[1]{{\raisebox{1pt}{\textcircled{\raisebox{-.9pt} {#1}}} }}

\newcolumntype{C}{>{$}c<{$}}

\newcommand\BibTeX{{\rmfamily B\kern-.05em \textsc{i\kern-.025em b}\kern-.08em
T\kern-.1667em\lower.7ex\hbox{E}\kern-.125emX}}

\def\volumeyear{2016}
\begin{document}

\runninghead{Willibald and Lee}

\title{Hierarchical Task Decomposition for Execution Monitoring and Error Recovery: Understanding the Rationale Behind Task Demonstrations}

\author{Christoph Willibald\affilnum{1} and Dongheui Lee\affilnum{1}\affilnum{2}}

\affiliation{\affilnum{1}Institute of Robotics and Mechatronics, German Aerospace Center (DLR), Germany\\
\affilnum{2}Technische Universität Wien (TU Wien), Chair of Autonomous Systems, Vienna, Austria}

\corrauth{Christoph Willibald, Institute of Robotics and Mechatronics, German Aerospace Center (DLR), 82234 Wessling, Germany.}

\email{Christoph.Willibald@dlr.de}

\begin{abstract}
Multi-step manipulation tasks where robots interact with their environment and must apply process forces based on the perceived situation remain challenging to learn and prone to execution errors. Accurately simulating these tasks is also difficult. Hence, it is crucial for robust task performance to learn how to coordinate end-effector pose and applied force, monitor execution, and react to deviations. To address these challenges, we propose a learning approach that directly \textbf{infers} both \textbf{low- and high-level task representations} from user demonstrations on the real system. We developed an \textbf{unsupervised task segmentation} algorithm that combines \textbf{intention recognition} and \textbf{feature clustering} to infer the skills of a task. We leverage the inferred characteristic features of each skill in a novel \textbf{unsupervised anomaly detection} approach to identify deviations from the intended task execution. Together, these components form a comprehensive framework capable of incrementally learning task decisions and new behaviors as new situations arise. Compared to state-of-the-art learning techniques, our approach significantly reduces the required amount of training data and computational complexity while efficiently learning complex in-contact behaviors and recovery strategies. Our proposed task segmentation and anomaly detection approaches outperform state-of-the-art methods on force-based tasks evaluated on two different robotic systems.

\end{abstract}


\maketitle

\section{Introduction}
\label{sec:Introduction}
\begin{figure*}[th]
  \centering
   \includegraphics[scale=0.948]{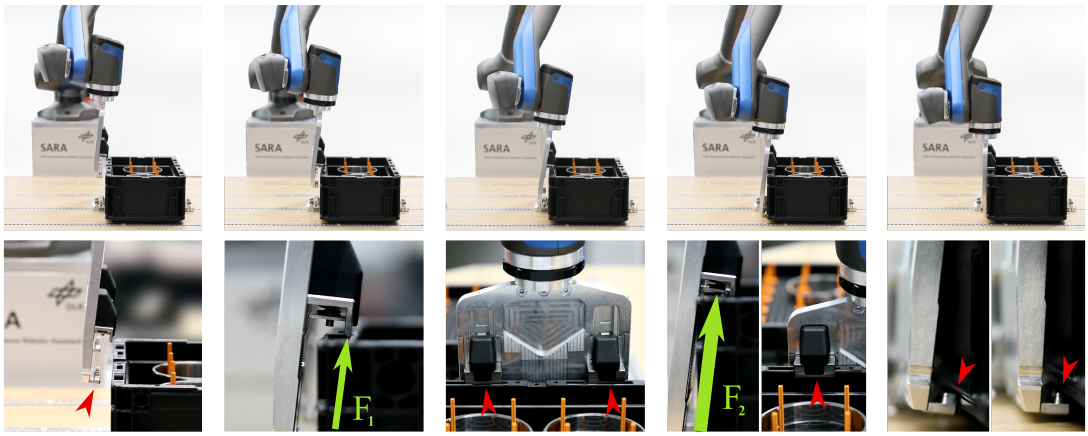}
  \caption{The different phases and challenges of the box grasping and locking task. The upper row shows EEF configurations in the subgoal region of the respective skills during the execution which trigger a transition to the successor skills. The bottom row highlights the difficulties during each skill. In the first skill, the lower part of the gripper must not collide with the box, while the movable slides are positioned above the box. In the next phases, the robot must apply force via the front part of the slides, without pushing the locking pin inside the slides, and maintain contact with the side wall of the box until the third subgoal configuration is reached. After that, the locking pin must be pushed to compress the springs in the slides while rotating the gripper into the vertical configuration. If there is not enough force exerted to push the gripper down, the tight clearance between the box and the box locks on the lower part of the gripper causes a collision. A video of the task learning and execution can be seen in Extension 3.}
  \label{fig:Teaser_Box_grasping}
\end{figure*}

Recent advancements in robot learning leverage large-scale general-purpose models trained on diverse datasets to learn policies capable of handling a broad range of different tasks \citep{Padalkar2023OpenXR, reed2022generalist}. Other works learn end-to-end policies for motion planning \citep{fishman2023motion}, or employ pre-trained vision-language models to detect anomalies during task execution \citep{du2023vision, zhang2023grounding, driess2023palm}. While these deep-learning models perform well for common tasks with many available training examples, such as pick-and-place or free-space motions, they struggle with specialized contact tasks where precise coordination of end-effector pose and applied force is important (see Fig.~\ref{fig:Teaser_Box_grasping}). Generating training data for such tasks is expensive as it involves sensing contact forces and torques in the real world, which requires a significantly higher effort compared to obtaining natural language and image datasets and thus leads to less availability of such data. Reinforcement Learning in simulation with sim-to-real transfer attempts to address this challenge and has shown impressive results for learning e.g. legged locomotion policies \citep{gangapurwala2022rloc} or dexterous in-hand manipulation \citep{pitz2023dextrous}. However, to focus policy learning on physically accurate environment configurations, it is still necessary to precisely identify and simulate the system. Especially for complex multi-point contact simulations, physics engines are still limited when calculating friction and contact forces \citep{yoon2023comparative, liao2023performance}. To avoid the required modeling effort for every new setup and to enable learning of specialized skills, we propose a framework capable of learning contact tasks from only a few demonstrations on the real system.

 Learning from Demonstration (LfD) is a popular method for transferring task knowledge from humans to robots. A user provides task examples, e.g. via remote control of the robot or by kinesthetic teaching, from which the robot learns how to perform a given task, even if the environment conditions change. Modern lightweight robots, like the DLR SARA \citep{iskandar2021collision}, can sense the contact forces during demonstrations and reproduce them during execution. To accommodate different environment conditions and to make task decisions online, it is insufficient to simply replay a demonstration. Movement Primitives \citep{ijspeert2002movement, pervez2018learning, calinon2016tutorial, huang2019kernelized} or Dynamical Systems \citep{hersch2008dynamical, khansari2011learning, calinon2016tutorial} are established approaches to react to low-level perturbations and allow to generalize trajectories to varying start and end points while preserving the nature of the movement. However, they can not capture the high-level structure of a task, which prohibits adaptive behavior required for making task decisions or recovering from anomalies. Therefore, we propose to learn a hierarchical task representation from demonstrations. Our approach decomposes the task into more manageable sub-problems by first inferring the individual skills that comprise the task before structuring them in a task graph that can be incrementally extended if required.
 
 Our framework is designed to learn complex multi-step contact tasks requiring specialized skills, such as the box grasping task illustrated in Fig.~\ref{fig:Teaser_Box_grasping}. In such tasks, precise coordination of end-effector pose and commanded force is essential for successful execution. Since the observed motion and applied force during the grasping sequence are unique for this task, we can not use prior knowledge about commonly used skills to identify the sub-problems of the task. To address this, we developed an unsupervised task segmentation algorithm, combining inverse reinforcement learning with probabilistic clustering to identify a sequence of unique skills based on their individual subgoals and feature constraints. We assume that each skill intends to reach a specific subgoal, represented by a region of end-effector (EEF) configurations relative to important objects. Reaching the subgoal is considered the intention of a skill and is a postcondition for its successful termination. Further, a skill is governed by constraints that must be satisfied during its execution. The constraints can be described by multimodal features, such as applied force or the relative distances of the end-effector to objects. After segmentation, our framework uses the inferred feature constraints for unsupervised anomaly detection during autonomous skill execution. The anomaly detection approach determines the confidence for its predictions based on the availability of the training data and reacts accordingly. If a deviation from the intended skill execution is detected, a higher-level decision-making mechanism takes over. This mechanism, guided by the task graph and the nature of the anomaly, determines the appropriate recovery action. We make the following contributions:
 
\begin{itemize}
    \item Unsupervised task segmentation approach \textbf{integrating inverse reinforcement learning-based intention recognition and probabilistic feature clustering}.
    \item Unsupervised anomaly detection method leveraging each skill's \textbf{probabilistic representation of expected features to make decisions based on epistemic and aleatoric uncertainty}.
    \item \textbf{Framework for incremental learning of hierarchical task representations}.
\end{itemize}

Throughout this article, we repeatedly mention incremental hierarchical task learning. With hierarchical learning, we refer to the simultaneous process of learning both the low-level data-driven models for motion generation and anomaly detection for each skill, and a more abstract model of the entire task. The abstract model is represented as a Task Graph, which organizes skills at a higher level. Incremental learning, also referred to as continual learning \citep{lesort2020continual}, describes the strategy of learning an initial model that can later be refined \citep{simonivc2021analysis} or extended for novel conditions or situations that were not anticipated in the beginning. The incremental approach applies both to high- and low-level learning.
\section{Related Work}
\label{sec:related_work}
\subsection{Task Segmentation}
\label{sec:segmentation}
Task segmentation is often a necessary step in robotic task learning approaches to break down the complexity into more manageable sub-problems.
\subsubsection{Supervised Approaches}
\label{sec:supervised}
Probabilistic models are employed in \cite{eiband2023online}, \cite{wang2018masd}, \cite{kulic2012incremental}, and \cite{meier2011movement} to detect known skills in a task demonstration. \cite{kulic2012incremental} incrementally learn probabilistic models of motion primitives, whereas \cite{meier2011movement} assume a known DMP library for online movement recognition and segmentation. Support Vector Machines (SVM) with a sliding time window are used to predict the most likely skill at each time step from video data in \cite{wang2018masd} or using the robot's proprioceptive sensor measurements in \cite{eiband2023online}. 

Semantic skill recognition is used in \cite{eiband2023online}, \cite{wachter2015hierarchical}, \cite{ramirez2017transferring} and \cite{steinmetz2019intuitive}. In \cite{eiband2023online} and \cite{steinmetz2019intuitive}, a world model is queried to compare the current semantic object relations to pre- and post-conditions of skills formulated with the PDDL convention \citep{ghallab1998pddl}.
\cite{eiband2023online} and \cite{wachter2015hierarchical} use multi-step segmentation processes to refine the first rough result in a downstream step. In \cite{wachter2015hierarchical}, semantic segmentation is followed by a sub-segmentation based on the acceleration of the demonstrated trajectory, while \cite{eiband2023online} utilize an SVM-based segmentation to further refine identified contact skills.
\subsubsection{Unsupervised Approaches}
\label{sec:unsupervised}
Unsupervised methods aim to find characteristic points, similarities between demonstrations, or apply generative models to data to perform segmentation, all without requiring explicit knowledge of the observed skills. A Gaussian Mixture Model (GMM) expresses the joint probability distribution of high-dimensional training data as a weighted sum of independent Gaussian components. \cite{krishnan2017transition} utilize hierarchical clustering based on several GMMs to identify similar skill transition states across repeated task demonstrations. \cite{lee2015autonomous} apply Principal Component Analysis to one single task demonstration prior to fitting a GMM on the training data. Segmentation points are determined at the intersection of adjacent Gaussian components. \cite{karlsson2019segmentation} extend this approach by incorporating force measurements, where sudden changes in the interaction force are leveraged to verify or extend existing segmentations. In \cite{kruger2012imitation} and \cite{figueroa2018physically} a Bayesian nonparametric approach is used to fit a GMM onto demonstrations. In \cite{kruger2012imitation}, every mixture ensures global asymptotic stability for point-to-point motions. In \cite{figueroa2018physically}, only physically consistent clusters are generated by considering a similarity metric based on the distance of samples and direction of the velocity. In our approach, we employ a Bayesian nonparametric GMM (BN-GMM) to cluster demonstrated states but quantify action similarity in terms of reaching a common state subgoal using Q-values. In contrast to \cite{figueroa2018physically}, this allows us to infer the underlying intentions of actions by comparing the demonstrated actions with the optimal actions to reach a subgoal and thus can identify more complex behavior.

A Hidden Markov Model (HMM) can describe a process that evolves over time and has underlying unobservable modes. The modes can be interpreted as skills in the context of task segmentation. Bayesian nonparametric extensions of the HMM are used in \cite{niekum2012learning, grigore2017discovering, chi2017learning} where a beta process (BP) prior is leveraged to infer the number of active modes per demonstration. \cite{grigore2017discovering} combine a BP-HMM with a clustering approach to determine the appropriate level of granularity for identified motion primitives based on clustering performance. The Beta Process Autoregressive HMM relaxes the conditional independence of observations by describing time dependencies between observations as a Vector Autoregressive process \citep{niekum2012learning, chi2017learning}. \cite{kroemer2015towards} incorporate state dependency in the HMM's phase transition probability, which allows the model to learn regions, where phase transitions are more likely. In \cite{hagos2018segmenting}, the  phase transition probability depends on the measured interaction force, to account for contact changes between the robot and the environment that indicate a phase transition. 

In \cite{sugawara2023unsupervised}, segmentation points of contact-rich tasks are detected based on the time derivatives of force and torque measurements. To reduce over-segmentation due to sensor noise, Bayesian online change point detection \citep{adams2007bayesian} is used to identify true positive segmentation lines. If a robot's end-effector enters or leaves the proximity area of an object \citep{caccavale2019kinesthetic}, or the distance relations between objects change \citep{wachter2013action}, new segmentation lines are detected. Using pre-and post-conditions, the segments are then associated with semantic skills. \cite{shi2023waypoint} propose an algorithm to automatically extract a demonstration's minimal set of waypoints for which the trajectory reconstruction error lies below a specified threshold when linearly interpolating between the waypoints. In contrast to the approaches mentioned in this paragraph, our proposed approach does not segment each task demonstration individually but finds a combined segmentation result over all demonstrations. It leverages the similarities across different demonstrations, leading to a more consistent result compared to segmenting each demonstration individually. In \cite{ureche2015task} the variance of task variables within one, and across several demonstrations is analyzed to determine task constraints. Segmentation lines are drawn when the relevant task constraints change. \cite{manschitz2020learning} focus on segmentation for point-to-point motions, where the demonstrations are first intentionally over-segmented using Zero Velocity Crossing \citep{fod2002automated} before again combining segments converging to the same attractor. Similarly, \cite{lioutikov2017learning} iteratively eliminate false positive segmentation lines using a probabilistic segmentation approach with motion primitives as the generative skill models. 

In our approach, the observed skill sequence can be inferred from a single or several task demonstrations, where each skill's intention and feature constraints are used as grouping mechanisms in the data for segmentation. Our method combines a GMM in feature space with Inverse Reinforcement Learning to capture the intention and feature similarities of every state-action observation in a joint mixture model, where every mixture component represents a skill.

\subsection{Hierarchical Inverse Reinforcement Learning}
\label{sec:IRL}
Another approach to breaking down the complexity of a task into smaller sub-problems is Hierarchical Inverse Reinforcement Learning (HIRL). IRL as proposed by \cite{ng2000algorithms} avoids the cumbersome process of manually designing a reward function for a given task by representing the problem as a Markov Decision Process (MDP) with unknown reward function and learning the reward function from expert demonstrations. It may be difficult, however, or even impossible to represent a complex task with a single reward function. That is why HIRL finds segments of corresponding data points in the demonstration and solves the IRL problem per segment. To that end,  \cite{michini2012bayesian}, \cite{michini2015bayesian} propose Bayesian nonparametric IRL (BN-IRL). Using a Dirichlet Process mixture model as the prior over segments, BN-IRL divides an observed task demonstration into a set of smaller subtasks, so that each subtask can be described by a simple subgoal-based reward function. This approach was extended to Constraint-based BN-IRL (CBN-IRL) \citep{park2020inferring} to also infer parts of the demonstration where local feature constraints are active. This allows the model to represent complex behavior with a fewer number of segments as in BN-IRL, but requires predefined constraint boundaries for every feature and significantly increases the computational complexity. Since active feature constraints are indirectly determined by changing the restriction of the state transition function in CBN-IRL, only constraints with respect to the end-effector position and orientation can be considered. This approach does not allow to constrain contact forces or torques. Furthermore, changing the constraint boundary of one feature requires expensive recomputation of Q-values. The number of Q-value recomputations grows exponentially with the number of features when altering the boundaries independently. That is why CBN-IRL only distinguishes between constrained segments, where all features lie within the boundaries, and unconstrained segments. Our approach solves the problem of computational complexity by separating the constraint inference from the Q-value-based intention recognition. As a result, the number of required Q-value calculations is independent of the number of considered features. We infer individual feature constraint regions for each segment by modeling them as multivariate Gaussian distributions. This not only reduces the computational complexity but also allows our model to detect correlations between multimodal features across several task demonstrations.

Leveraging the option framework for describing temporally extended actions, the HIRL approaches in \cite{surana2014bayesian}, \cite{ranchod2015nonparametric}, and \cite{fox2017multi} can recover complex reward functions or policies from the demonstrations. On the higher level, \cite{surana2014bayesian} model the task as a switched MDP, and \cite{ranchod2015nonparametric} as a BP-HMM with emissions from different MDPs. \cite{fox2017multi} infer high-level meta policies and low-level options based on Deep Q-networks, which requires many training examples of the task. \cite{krishnan2016hirl, krishnan2019swirl} suggest a multi-step learning process consisting of sequence-, reward- and policy learning. The task segmentation is performed using transition state clustering similar to \cite{krishnan2017transition}. The transition states are then used to infer simpler rewards per segment via Maximum Entropy IRL \citep{ziebart2008maximum} to finally learn a policy via forward RL. In contrast to \cite{krishnan2016hirl, krishnan2019swirl}, our approach solves the segmentation problem with IRL. We identify subgoals by analyzing the actions observed within segments, which are directed toward reaching these subgoals.

\subsection{High-Level Task Structuring and Decision Making}
\label{sec:high-level}
As discussed in the previous sections, robotic tasks often consist of modular skills designed to address specific sub-problems of the task. Task-level decision-making determines how to apply those skills, taking into account the current context or the outcomes of prior skills. In current robotic approaches, the organization of low-level skills and the handling of high-level decisions are often achieved through the use of Behavior Trees (BT), or Task Graphs (TG), which are variants of Finite State Machines. Originally developed for decision-making in computer games, Behavior Trees have gained attention in robotics and have proven their effectiveness in various applications including machine tending \citep{guerin2015framework}, polishing and assembly \citep{paxton2017costar, rovida2018motion, mayr2021learning}, and autonomous navigation \citep{de2023autonomous}. BTs and TGs share close relationships, and the high-level structure of a task can often be effectively represented using either. However, due to their design, BTs can transition more flexibly between skills, which in turn makes it more difficult to infer their structure from demonstrations.

The Learning from Demonstration (LfD) approaches in \cite{manschitz2020learning}, \cite{caccavale2019kinesthetic}, \cite{willibald2022multi}, \cite{willibald2020collaborative}, \cite{kappler2015data}, \cite{niekum2015learning}, \cite{konidaris2012robot}, \cite{su2018learning} propose the use of Task Graphs to organize skills at a higher level of abstraction. This structure can be defined manually by an operator \citep{kappler2015data} or learned from demonstration. For the latter case, multiple task demonstrations are segmented individually and transformed into a TG with \cite{konidaris2012robot}, \cite{niekum2015learning}, \cite{su2018learning}. \cite{konidaris2012robot} employ an iterative merging approach starting from the final segment, while \cite{su2018learning} cluster segments based on their final configuration and connect them based on transition frequencies. \cite{niekum2015learning} additionally split up nodes in the TG based on groupings of the node's parents. \cite{willibald2020collaborative} follow a different approach and add decision states with recovery behaviors at fixed time steps when anomalies are detected. The approach presented in this paper involves learning a common skill sequence from multiple task demonstrations, which can be incrementally extended with new skills to flexibly address task decisions or recovery behaviors.

Unlike approaches that do not incorporate higher-level organization \citep{pastor2012towards, Eiband2019}, TG-based methods restrict the number of candidate skills for transitions based on the currently executed skill. This restriction addresses the perceptual aliasing problem, where a task decision can not be determined solely based on current sensor readings but might require that the robot has already reached certain high-level goals. While \cite{caccavale2019kinesthetic}, \cite{willibald2020collaborative}, \cite{niekum2015learning}, \cite{konidaris2012robot} and \cite{su2018learning} allow transitions only at the end of a skill, \cite{kappler2015data} introduce an online decision-making system employing supervised classification to determine when and to which successive skill to switch, however their approach does not autonomously detect and resolve new anomalies but requires human supervision for that. \cite{denivsa2015synthesis} cluster segments of task demonstrations and store them in a hierarchical skill database that can be queried at runtime to generate new movements by recombining partial paths that were not demonstrated together. While this approach reduces the number of required task demonstrations it does not explicitly encode the temporal sequence of partial paths. In \cite{manschitz2020learning}, a sequence graph synchronizes independent motion primitives for the end-effector pose, force, and finger configuration, where each node has at most one successor and a classifier determines when to transition to the next node. We employ an unsupervised approach to identify deviations in the current execution. When such deviations are detected, we transition to a suitable recovery behavior within the TG, based on the identified failure mode, after which the robot can continue with the intended task execution. Additionally, when the skill's subgoal is reached, we seamlessly transition to the next task-flow node in the TG. This approach offers the combined advantages of skill transitioning flexibility found in BTs and the candidate skill restriction of TGs.

\subsection{Multimodal Anomaly Detection}
\label{sec:anomaly}
Robust anomaly detection and recovery are vital for autonomous robotic systems. To detect anomalies, we employed time-based Gaussian Mixture Regression (GMR) in previous works \citep{willibald2020collaborative, Eiband2019, eiband2023collaborative} to compute the Mahalanobis distance between the robot's measured and expected proprioceptive sensor values. The probabilistic modeling allows the approach to scale the anomaly detection sensitivity depending on the current timestep. \cite{romeres2019anomaly} use Gaussian Process Regression to learn the expected force profile and epistemic uncertainty during insertion tasks in combination with a predefined anomaly threshold for anomaly detection. A Hidden Markov Model is used to detect multimodal anomalies, either with predefined anomaly thresholds and a sliding time window \citep{azzalini2020hmms} or with probabilistic threshold estimation based on execution progress \citep{park2019multimodal}. \cite{Chernova2007} encode a simple policy using basic symbolic actions via a GMM, where the observation likelihood of unseen states is used to detect outliers. Stereotypical sensor traces along with movement primitives are used in \cite{kappler2015data} and \cite{Pastor2011} for supervised anomaly detection.

We employ GMR where expected feature values and allowed deviations are predicted by conditioning on the measured end-effector pose relative to the relevant coordinate system for the current skill. Conditioning anomaly detection on time or task progress would require consistent feature profiles across demonstrations, i.e. an exact replication of the situation in all runs. We argue that important features, such as contact forces, are influenced by interaction dynamics between the robot and the environment rather than time. Additionally, conditioning on the relative end-effector pose allows our anomaly detection approach to distinguish between epistemic and aleatoric uncertainty and to handle both cases individually. Notably, this makes our approach unique with respect to the state of the art. Epistemic uncertainty arises from incomplete model information and can be reduced by collecting additional data, while aleatoric uncertainty represents the variability in the underlying distribution. While the approach in \cite{silverio2019uncertainty} acknowledges the two interpretations of variance, it equally modulates the robot's control gains to obtain a compliant robot in cases of high uncertainty and variability. \cite{Maeda2017} utilize the epistemic uncertainty of a Gaussian Process in the context of trajectory generation to quantify the generalization capability of the model to unseen goal positions. Our approach, however, leverages the epistemic uncertainty to determine the confidence in the anomaly detection, while the aleatoric uncertainty is used to scale the anomaly detection sensitivity.

\cite{park2018multimodal}, \cite{pol2019anomaly}, \cite{azzalini2021minimally} employ Variational Autoencoders (VAEs) for anomaly detection. VAEs use a decoder network to reconstruct the original input from a compressed latent space representation, learned from normal data instances. Anomalous data leads to a higher reconstruction error, which can be used to identify anomalies. Other works based on deep neural networks use sensor streams including RGB-D images to detect anomalies during robotic manipulation \citep{inceoglu2024multimodal, altan2022clue, yoo2021multimodal}. Recent works apply pre-trained vision-language models (VLMs) or large vision-language models (LVLMs) to robotics by treating anomaly detection as a visual question-answering problem \citep{du2023vision, zhang2023grounding, driess2023palm, agia2024unpacking}. ConditionNET \citep{sliwowski2024conditionnet} is a VLM designed to predict preconditions and effects of skills. It frames anomaly detection as a state prediction problem, where an anomaly is detected if the predicted and expected state do not match. While deep-learning-based approaches have shown improved performance over comparable state-of-the-art techniques, they need large annotated training datasets. Our anomaly detection approach is focused on challenging in-contact manipulation tasks where the force between the robot and the environment plays a crucial role for the task's success (see Fig.~\ref{fig:Teaser_Box_grasping}). The vast amount of required training data for deep-learning-based approaches is not available and even hard to obtain in simulation for such scenarios. That is why we propose a Learning from Demonstration setup, using data-efficient methods, that are capable of identifying anomalies with just a few task demonstrations. We demonstrate robust anomaly detection capabilities with only up to three user demonstrations.
\section{Incremental Task Learning Framework}
\label{sec:procedure}
\begin{figure*}[thpb]
  \centering
   \includegraphics[scale=0.945]{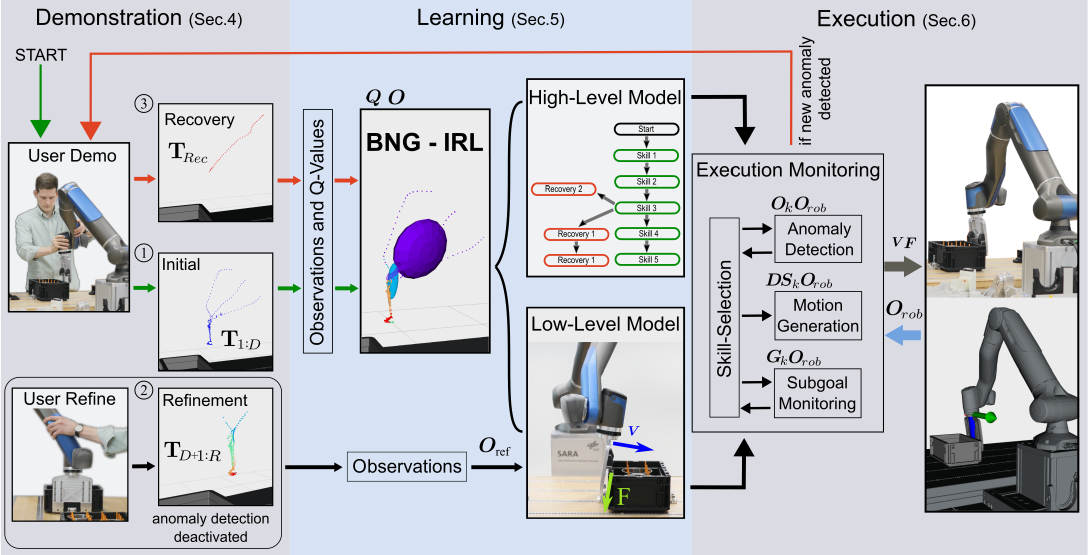}
  \caption{Our proposed incremental high- and low-level task learning framework. The process of learning a new task starts with an initial teaching sequence \ciwn{1}(Sec.~\ref{sec:initial-ts}), where the user provides several demonstrations of the intended task that are segmented using BNG-IRL to learn an initial task model. The low-level skills of this initial model can then be further refined during the skill refinement teaching sequence \ciwn{2}(Sec.~\ref{sec:skill-refinement-ts}), depicted at the bottom. During this phase, the robot performs the initial task, while the user supervises and assists in phases, where the skills need refinement. The recorded training data is used to update the skills. The last teaching sequence \ciwn{3}(Sec.~\ref{sec:task-decision-ts}) is triggered if the execution module identifies a new anomaly (see Sec.~\ref{sec:motion-generation-anomaly-detection} and \ref{sec:skill-selection}). Similar to the initial teaching sequence, the user provides a demonstration that shows how to recover from the anomaly, which is segmented and appended to the skill during which the anomaly occurred.}
  \label{fig:Overview}
\end{figure*}
In this section, we introduce the overall framework of our incremental task learning approach before describing the individual components in more detail in Sec.~\ref{sec:Learning} and Sec.~\ref{sec:Execution}. The framework, as depicted in Fig.~\ref{fig:Overview}, consists of three sequential phases: \textit{Demonstration}, \textit{Learning}, and \textit{Execution}. These phases form a teaching sequence, which can be triggered multiple times throughout the incremental task-learning process. However, the teaching sequences differ depending on the event that triggered them.
\subsection{Initial Teaching Sequence}
\label{sec:initial-ts}
The task learning process begins with the first teaching sequence, aimed at learning the intended task flow model. The task model comprises both low-level skill representations and a high-level Task Graph that organizes the skills on a higher level of abstraction. To start the process, a user demonstrates the strategy to solve the task with the robot multiple times. Our experiments demonstrate a good learning performance for a contact-rich task with as few as three initial user demonstrations. The recorded data from these user demonstrations is passed to the learning component, which segments the demonstrations into a skill sequence using our proposed Bayesian Nonparametric Gaussian Inverse RL (BNG-IRL) segmentation approach detailed in Sec.~\ref{sec:Segmentation}. The Task Graph is initially based on the inferred skill sequence from the first user demonstrations. The learned task model can already be executed, however the anomaly detection component is not activated during the first few executions on the robot to allow the user to refine the learned low-level skill representations.
\subsection{Skill Refinement Teaching Sequence}
\label{sec:skill-refinement-ts}
In this teaching sequence, the user monitors the robot collecting additional training data while performing the task autonomously. Since the robot is impedance controlled, the user can help the robot during parts where the learned task model is not yet optimal, providing additional training input. The corrective support data, combined with the data collected by the robot during autonomous execution is used to refine each skill's low-level motion generation and anomaly detection model. Depending on the application and learning setup, our approach can be combined with various motion generation methods capable of learning policies from sparse training data, based on Movement Primitives, or Dynamical Systems. Once the initial task model achieves sufficient robustness, anomaly detection is activated, and high-level task learning can be initiated.
\subsection{Task Decision Teaching Sequence}
\label{sec:task-decision-ts}
In the final teaching sequence, high-level decisions are learned that can not be resolved by the low-level motion generation approach but require to change the strategy. These decisions include normal task decisions and recovery behaviors, which may arise due to diverse environment conditions or unintended task interferences. As illustrated in Fig.~\ref{fig:Overview}, the execution monitoring module of our framework has access to the task model and consists of submodules for skill selection, anomaly detection, motion generation, and subgoal monitoring. Detailed descriptions of these modules are provided in Sec.~\ref{sec:Execution}. The skill selection module handles high-level task decisions and communicates the selected skill to the other execution modules. Task decisions can be triggered by either the anomaly detection or subgoal monitoring module. When a skill's subgoal is reached, the selection module transitions to the intended subsequent skill in the Task Graph, learned from the initial teaching sequence. In contrast, if an anomaly or a different environment condition is detected, the skill selection module must determine the appropriate skill for that situation. As the initial task model only comprises the skill sequence for the intended task flow, recovery skills must be acquired from user demonstrations. To facilitate this, our framework employs an incremental learning approach. When an anomaly is detected and the skill selection module cannot find a suitable skill for recovery in the Task Graph, a new teaching sequence is initiated. During this sequence, the user demonstrates how to resolve the situation with the robot. The recorded data is segmented into a skill sequence, as described for the first teaching sequence. However, this skill sequence is appended to the skill in the Task Graph where the anomaly was detected. In subsequent executions, the robot can autonomously apply the newly learned recovery behavior to address similar anomalies. Thanks to our subgoal and feature-based skill inference, the robot can flexibly reuse the newly acquired recovery behavior throughout a skill, without being limited to the specific time step where the anomaly occurred or needing an exact replication of the anomaly.
\section{Demonstration}
\label{sec:Demonstration}
Each teaching sequence begins with a demonstration phase to gather new training data. As shown in Fig.~\ref{fig:Overview}, the mode of demonstration depends on whether the goal is to learn a new skill sequence (Sec.~\ref{sec:initial-ts} and \ref{sec:task-decision-ts}) or to refine existing low-level skills (Sec.~\ref{sec:skill-refinement-ts}). We distinguish between two modes: \textit{User Demonstration} and \textit{User Refinement}.

\subsection{User Demonstration}
\label{sec:User_Demo}
In the User Demonstration mode, we adopt a conventional kinesthetic teaching setup for high- and low-level skill learning as described for the initial task model- and the task decision teaching sequence. During this mode, the robot compensates for its own weight, while the user hand-guides the robot to perform the task. The task can be demonstrated either once or repeatedly to increase the variation of the training data across varying environmental conditions. Throughout the demonstration, the robot captures proprioceptive sensor measurements, and we also track the poses of objects within the robot's workspace using an external camera setup. The recorded data includes the 6D object and end-effector (EEF) poses, the 6D vector of contact forces and torques at the EEF, and, if applicable, data related to an active tool, such as gripper finger distance and grasp status.

We record the robot's EEF trajectories from $D$ task demonstrations and construct a state-action sequence $\bm{\mathrm{T}}_d$ for each demonstration. Each sequence 
\begin{gather*}
  \bm{\mathrm{T}}_d = [(\bm{s}_1, \bm{a}_1), ..., (\bm{s}_{N_d}, \bm{a}_{N_d})]
\end{gather*}
comprises a variable number of $N_d$ state-action pairs, where $d \in [1, ..., D]$, $\bm{s}_i \in \bm{S}$ and $\bm{a}_i \in \bm{A}$. The state space $\bm{S}$ includes all reachable robot EEF poses, while actions $\bm{a}_i$ are defined as the EEF velocities. For every state-action pair, we compute a feature vector $\bm{f}_i \in \bm{\mathcal{F}}$. The feature space $\bm{\mathcal{F}}$ consists of a set of generic multimodal features, including but not limited to contact forces and torques, relative distances and orientations between the robot and objects, audio signals, and tool information. This results in a total number of $N = \sum_{d=1}^D N_d$ data points across all demonstrations, termed observation set
\begin{gather*}
  \bm{O} = \{(\bm{f}_i, \bm{s}_i, \bm{a}_i){\rbrace}_{i=1}^N.
\end{gather*}
 
\subsection{User Refinement}
\label{sec:User_Support}
The second demonstration mode is designed to refine low-level models of previously learned skills by gathering additional training data with the robot. In this phase, the robot employs the initial motion generation policy of each skill from the previous teaching sequence and generalizes it to varying environment configurations. While the robot performs the task, the user supervises and provides manual support if the initial model fails to achieve specific task subgoals.

To prevent false positive anomaly detections, caused by external forces applied by the user during corrective support, the anomaly detection mechanism is deactivated during this phase. However, contact forces between the robot's EEF and the environment, as well as the sensor measurements described in Sec.~\ref{sec:User_Demo}, are recorded and used to update the low-level skill models.

For this purpose, we compile an observation set $\bm{O}_{\mathrm{ref}}$ containing $N_{\mathrm{ref}}$ data points from the user refinement phase. Each data point contains a skill assignment parameter $z_i \in [1, ..., K]$, which maps each observation $\bm{o}_{i, \mathrm{ref}}$ to its respective skill $k$ out of $K$ skills:
\begin{gather*}
  \bm{O}_{\mathrm{ref}} = \{(z_i, \bm{o}_{i, \mathrm{ref}}){\rbrace}_{i=1}^{N_{\mathrm{ref}}} = \{(z_i, \bm{f}_i, \bm{s}_i, \bm{a}_i){\rbrace}_{i=1}^{N_{\mathrm{ref}}} \\
\end{gather*}
Each observation set $\bm{O}_k$ used to model the low-level behavior for motion generation and anomaly detection of the associated skill is updated with the observations $\{\bm{o}_{i, \mathrm{ref}}  \vert z_i=k\}$. After that, the robot can fully autonomously execute the refined skills and robustly detect deviations from the intended behavior. 
\section{Hierarchical Task Learning}
\label{sec:Learning}
Each demonstration phase is followed by a learning phase to incorporate the newly collected training data into the task model. As different demonstration modes serve to learn different aspects of the model, the learning phases vary accordingly: 
\begin{enumerate}
    \item After new \textit{User Demonstrations}, the observation sets need to be segmented into skills before they can be added to the task graph (see upper rows in Fig.~\ref{fig:Overview}).
    \item In the case of \textit{User Refinements}, observations are already assigned to skills and can be directly used as new training data for low-level skill refinement (bottom row in Fig.~\ref{fig:Overview}).
\end{enumerate}
In our task learning approach, we represent demonstrations as sequences of skills, where each skill comprises a motion primitive, a subgoal, and a feature constraint region. To infer these properties along with the skills themselves, we introduce a novel task segmentation algorithm called Bayesian Nonparametric Gaussian Inverse Reinforcement Learning (BNG-IRL) (see Fig.~\ref{fig:BNG-IRL_Overview}). This algorithm combines probabilistic clustering of observed demonstration states in feature space with intention recognition based on inverse reinforcement learning.

\subsection{Unsupervised Task Segmentation BNG-IRL}
\label{sec:Segmentation}
In this section, we first introduce the components of the task segmentation approach that are integrated in the probabilistic model. After that, the model parameter inference is explained.
\subsubsection{Subgoal-Driven Intention Recognition}
\begin{figure}[thpb]
  \centering
  \begin{subfigure}{0.5\textwidth}
      \includegraphics[trim={0 0 0 0}, clip, width=\textwidth]{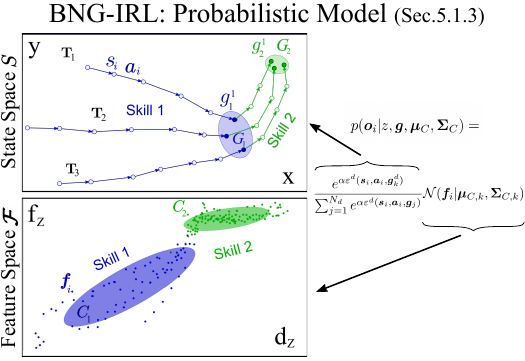}
      \caption{Three demonstrations of a task consisting of two different skills. Every demonstration $\bm{T}_d$ reaches each skill's subgoal region $\bm{G}_k$ (upper image and Sec.~\ref{sec:subgoal-driven-intention}), while the recorded features of the demonstrations lie within the characteristic constraint regions $\bm{C}_k$ of the skills (lower image and Sec.~\ref{sec:feature clustering}). The observation likelihood~(\ref{eq:obs_llh}) incorporates both influences.}
      \label{fig:BNG-IRL_Model}
  \end{subfigure}
  \begin{subfigure}{0.5\textwidth}
      \includegraphics[width=\textwidth]{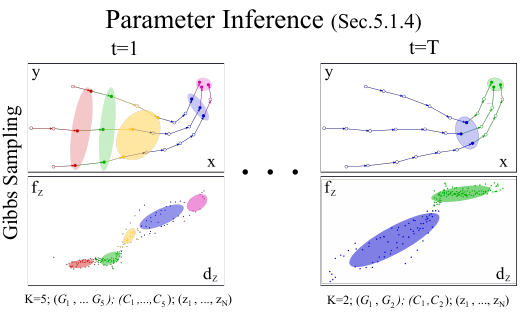}
      \caption{BNG-IRL uses Gibbs Sampling to infer the latent variables and parameters of the probabilistic model: the number of skills $K$, the skill assignment $z$ for every observation represented by the color of the samples, the optimal state subgoal $\bm{g}_k^d$ per skill and demonstration as well as the subgoal region $\bm{G}_k=\mathcal{N}(\bm{\mu}_{G,k}, \bm{\Sigma}_{G,k})$ per skill across all demonstrations, and the constraint region $\bm{C}_k=\mathcal{N}(\bm{\mu}_{C,k}, \bm{\Sigma}_{C,k})$ of every skill in feature space.}
      \label{fig:BNG-IRL_Inference}
  \end{subfigure}
\caption{Model and parameter inference of our proposed unsupervised task segmentation approach BNG-IRL.}
\label{fig:BNG-IRL_Overview}
\end{figure}
\label{sec:subgoal-driven-intention}
The subgoal of a skill serves as a post-condition that must be met at the end of a skill for successful completion. Across all task demonstrations $\bm{\mathrm{T}}_d$ shown in the upper row of Fig.~\ref{fig:BNG-IRL_Model}, each skill $k$ intends to reach a subgoal region $\bm{G}_k$. This concept is closely related to low-level motor intentions of cognitive science defined in \cite{pacherie2008phenomenology}. For the $k$-th skill of the $d$-th demonstration, there exists one specific state subgoal $\bm{g}_k^d$ that the user reaches with the robot. Given the assumption that the demonstrator reaches each skill's subgoal during every demonstration, we can limit the subgoal candidates for one demonstration to the observed states during that demonstration $\bm{s}_i \in \bm{\mathrm{T}}_d $.

To determine how good an action is in terms of reaching a demonstration's state subgoal, we utilize the Q-function. The Q-function is also referred to as the state-action value function and is the expected accumulated reward for taking action $\bm{a}$ in state $\bm{s}$ and then following policy $\pi$. More specifically, we evaluate the state-action value function $Q^{\pi^{\bm{T}_d}}(\bm{s}_i, \bm{a}_i, \bm{g}_k^d)$ for taking the observed action $\bm{a}_i$ in state $\bm{s}_i$ and then following the demonstrated policy from $\bm{\mathrm{T}}_d$ until reaching subgoal $\bm{g}_k^d$. To compare this result with the optimal policy $\pi^*$ for reaching $\bm{g}_k^d$ from $\bm{s}_i$ we also compute the result for the optimal value function $V^{\pi^*}(\bm{s}_i, \bm{g}_k^d)$. For simplicity, we will refer to the above-mentioned functions as $Q^{\bm{T}_d}$ and $V^*$. Dividing $Q^{\bm{T}_d}$ by $V^*$ provides a measure of optimality $\varepsilon^d$ for the observed policy, action $\bm{a}_i$ and subgoal candidate $\bm{g}_k^d$, where $0 \leq \varepsilon^d = \frac{Q^{\bm{T}_d}}{V^*} \leq 1$. $Q^{\bm{T}_d}$ and $V^*$ are computed through the Bellman Equation (\ref{eq:Q_obs}) and the Bellman Optimality Equation (\ref{eq:V_opt}), respectively.

\begin{equation}
\begin{split}
\ & \ Q^{\bm{T}_d}(s, a, g) \\
\ = & \sum_{s' \in S} P(s'|s,a) (R_g(s,a, s') + \gamma V^{\bm{T}_d}(s', g)) \\
\ = & \ R_g(s,a) + \gamma V^{\bm{T}_d}(s', g)
\label{eq:Q_obs}
\end{split}
\end{equation}
In both Equations (\ref{eq:Q_obs}) and (\ref{eq:V_opt}), we assume the same discount factor $\gamma$ and a deterministic transition from state $\bm{s}$ to $\bm{s'}$ when taking action $\bm{a}$, hence we can simplify the equations according to the last line.
\begin{equation}
\begin{split}
\ & V^*(s, g) = \max_{a \in A(s)} Q^*(a, s, g)\\
= & \max_{a \in A(s)} \sum_{s' \in S} P(s'|s,a) (R_g(s,a,s')+ \gamma V^*(s', g)) \\
= & \max_{a \in A(s)} (R_g(s,a)+ \gamma V^*(s', g))
\label{eq:V_opt}
\end{split}
\end{equation}
Similar to \cite{michini2015bayesian} and \cite{park2020inferring}, we use a sparse reward function $R_g(\bm{s}, \bm{a}, \bm{s}')=\mathds{1}(\bm{s}'=\bm{g})$ that returns the complete reward when transitioning to subgoal state $\bm{g}$. This reflects our initial assumption that all actions of a skill are targeted toward reaching the skill's subgoal $\bm{g}_k^d$. However, $\bm{g}_k^d$ is unknown and has to be inferred from the demonstrations. 
\begin{figure}[thpb]
  \centering
  \begin{subfigure}{0.49\textwidth}
      \includegraphics[trim={0 1cm 0 0}, clip, width=\textwidth]{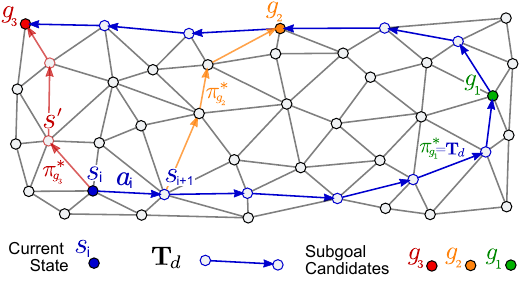}
      \caption{Demonstrated trajectory (blue) and random subgoals $g_1, g_2, g_3$ with respective trajectories following the optimal policy to reach them from state $\bm{s}_i$.}
      \label{fig:Value_Fct_leg}
  \end{subfigure}
  \begin{subfigure}{0.49\textwidth}
      \includegraphics[width=\textwidth]{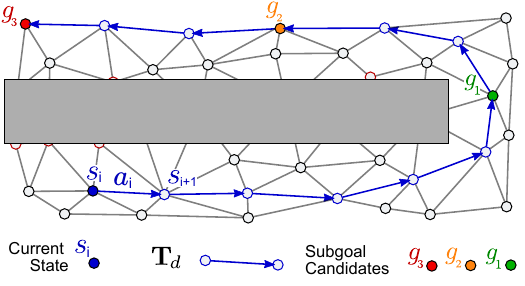}
      \caption{Same setup as in (a), but an introduced obstacle changes the optimal trajectories to reach the subgoals. The demonstrated trajectory becomes now the optimal trajectory to reach the subgoals.}
      \label{fig:Value_Fct_leg_obs}
  \end{subfigure}
\caption{Simplified 2D example of subgoal-driven intention recognition based on IRL for a single demonstration.}
\label{fig:Value_Fct_general}
\end{figure}

We illustrate the advantages of our optimality score $\varepsilon^d(\bm{s}_i, \bm{a}_i, \bm{g}_k^d) = \frac{ Q^{\bm{T}_d}(\bm{s}_i, \bm{a}_i, \bm{g}_k^d)}{V^*(\bm{s}_i, \bm{g}_k^d)}$ with the help of a simplified 2D example in Fig.~\ref{fig:Value_Fct_general}. From the setup in Fig.~\ref{fig:Value_Fct_leg} it becomes clear that the demonstrated action $\bm{a}_i$ in state $\bm{s}_i$ is not optimal to reach $\bm{g}_3$, which leads to a low optimality score. However, for subgoal $\bm{g}_1$, $\bm{a}_i$ and the subsequent demonstration follows the optimal policy, resulting in an optimality score tending towards 1. Now, we look at subgoal $\bm{g}_2$ in Figure~\ref{fig:Value_Fct_leg}. Compared to the optimal action $\bm{a} \in A(\bm{s}_i)$, that maximizes Eq.~(\ref{eq:V_opt}), the observed action $\bm{a}_i$ in $\bm{s}_i$ appears to be targeted to reaching this subgoal. However, the demonstrated trajectory after that action until reaching $\bm{g}_2$ is not optimal with regard to that subgoal, which reduces $Q^{\bm{T}_d}(\bm{s}_i, \bm{a}_i, \bm{g}_2)$, and hence the optimality score. Different from \cite{michini2015bayesian} and \cite{park2020inferring}, the observed policy from the demonstration after $\bm{a}_i$ is also taken into account when computing the optimality score with our approach. With this, we extend the time horizon for judging the intention of the demonstration from only one timestep to the duration until reaching the subgoal. This is advantageous for our approach since we aim to find a sequence of concise skills, where the intention changes with every skill, but not with every timestep (see Fig.~\ref{tab:comparison_box_pushing}). Another advantage can be seen from Fig.~\ref{fig:Value_Fct_leg_obs}, where an obstacle blocks the direct way to reach subgoals $\bm{g}_2$ and $\bm{g}_3$ from $\bm{s}_i$. Compared to using the velocity vector as an indicator for clustering similar observations, as proposed in \cite{figueroa2018physically}, the formulation based on the Q-function allows us to infer more complicated behaviors between the robot and the environment. Even though $\bm{a}_i$ is moving away from $\bm{g}_3$, we can infer that the action is targeted toward reaching that subgoal by comparing it to the optimal policy for reaching $\bm{g}_3$ in this scenario, resulting in a high optimality score. We obtain the probability distribution
\begin{equation}
    p(\bm{s}_i, \bm{a}_i\vert\bm{g}) = \frac{e^{\alpha \varepsilon^d(\bm{s}_i, \bm{a}_i, \bm{g})}}{\sum_{j=1}^{N_d}e^{\alpha \varepsilon^d(\bm{s}_i, \bm{a}_i, \bm{g}_j)}},
\end{equation}
by computing the softmax function of the optimality score $\varepsilon^d(\bm{s}_i, \bm{a}_i, \bm{g})$ for all subgoal candidates $\bm{g}_j$ of demo $d$. The parameter $\alpha$ describes the degree of confidence in the user to provide an optimal demonstration.
\subsubsection{Probabilistc Feature Clustering}
\label{sec:feature clustering}
Another source of information that we consider to cluster the different skills that comprise a task are the recorded feature values $\bm{f}_i \in \bm{\mathcal{F}}$ of the demonstrations, which complete the observation set $\bm{O}$ used as training data for our segmentation algorithm. We model the expected feature region of a skill and correlations among the features as a multivariate Gaussian distribution in feature space $\bm{\mathcal{F}}$, where $\bm{f}_i \sim \mathcal{N}(\bm{\mu}_{C,k}, \bm{\Sigma}_{C,k}) | z_i = k$, with mean $\bm{\mu}_{C,k}$ and covariance matrix $\bm{\Sigma}_{C,k}$ for every skill $k$ (see bottom row of Fig.~\ref{fig:BNG-IRL_Model}). With this, we can account for multi-modal and task-relevant information such as force readings, relative distances, and tool information to cluster different skills. That allows the algorithm to distinguish between in-contact and free-space motions or skills with different interaction force profiles. When loading or tensioning a spring, for example, the interaction force increases linearly with distance, whereas other tasks, such as opening a drawer, might require a constant force profile. By considering the observation set $\bm{O}$, containing the samples from all task demonstrations, we formulate the problem of segmentation as fitting a Gaussian Mixture Model to the data, where each mixture component corresponds to one skill. In the next section, we present a method to unify probabilistic feature clustering with subgoal-driven intention recognition in one probabilistic model.

\subsubsection{Probabilistic Model}
\label{sec:segmentation_model}
Combining the aspects of feature clustering and subgoal-driven action generation, we define the following observation likelihood for every $\bm{o}_i \in \bm{O}$
\begin{equation} 
\begin{split}
\label{eq:obs_llh}
&\ p(\bm{o}_i\vert z, \bm{g}, \bm{\mu}_C, \bm{\Sigma}_C) = p(\bm{s}_i, \bm{a}_i\vert\bm{g}_k^d) \ p(\bm{f}_i\vert \bm{\mu}_{C, k}, \bm{\Sigma}_{C, k})\\
=&\ \frac{e^{\alpha \varepsilon^d(\bm{s}_i, \bm{a}_i, \bm{g}_k^d)}}{\sum_{j=1}^{N_d}e^{\alpha \varepsilon^d(\bm{s}_i, \bm{a}_i, \bm{g}_j)}}\mathcal{N}(\bm{f}_i\vert \bm{\mu}_{C,k}, \bm{\Sigma}_{C,k}),
\end{split}
\end{equation}
where the first term represents the subgoal-driven policy under the skill-specific reward function $R_{g_k}$, and the second term represents the feature clustering of skill $k$ as described in \ref{sec:subgoal-driven-intention} and \ref{sec:feature clustering}, and depicted in Fig~\ref{fig:BNG-IRL_Model}.

Due to the described underlying grouping mechanism of the task demonstrations into skills with an a priori unknown number of skills, a Bayesian nonparametric Mixture Model with observation likelihood~(\ref{eq:obs_llh}) is used as the generative model to explain the observation set $\bm{O}$. Every mixture component corresponds to a skill with its own subgoal-based reward function and feature constraints. This leads to the following latent variables and model parameters, which have to be inferred from data:
\begin{itemize}
    \item The optimal \textbf{number of observed skills} $K$.
    \item An optimal \textbf{state subgoal} $\bm{g}_k^d$ per demonstration and skill, as well as the \textbf{subgoal region} $\bm{G}_k$ per skill across all demonstrations represented as a multivariate Gaussian distribution $\mathcal{N}(\bm{\theta}_{G,k})$ in state space $\bm{S}$ with parameters $\bm{\theta}_{G,k} = (\bm{\mu}_{G,k}, \bm{\Sigma}_{G,k})$.
    \item An optimal \textbf{constraint region} $\bm{C}_k$ for every skill represented as a multivariate Gaussian distribution $\mathcal{N}(\bm{\theta}_{C,k})$ in feature space $\bm{\mathcal{F}}$ with parameters $\bm{\theta}_{C,k} = (\bm{\mu}_{C,k}, \bm{\Sigma}_{C,k})$.
    \item The \textbf{skill assignment} $z_i=k$ for every observation $\bm{o}_i$ of the demonstrations, which determines the skill membership.
\end{itemize}The joint probability distribution over $\mathbf{O}$, $\mathbf{z}$, $\bm{g}$ and $\bm{\theta}_C$ is thus given by
\begin{equation}
\begin{split}
\label{eq:Generative_P}
&\ p(\bm{O}, \bm{z}, \bm{g}, \bm{\theta}_C) \\
= & \prod_{i=1}^{N} \underbrace{p(\bm{o}_i\vert z_i, \bm{g}, \bm{\theta}_C)}_\text{observation llh.} \underbrace{p(z_i\vert \bm{z}_{\setminus i})}_\text{z prior (CRP)} \prod_{k=1}^{K}\underbrace{p(\bm{g}_k)}_\text{$g$ prior}\underbrace{p(\bm{\theta}_{C, k})}_\text{$\theta_C$ prior}
\end{split}
\end{equation}
A Chinese Restaurant Process (CRP) with concentration parameter $\eta$ is used as a prior for the component assignment $z_i$, which allows to scale the number of mixture components with a growing number of observations up to a potentially infinite number.
\begin{equation}
\label{eq:CRP}
p(z_i=k\vert \bm{z}_{\setminus i}, \eta) =
\begin{cases}
  \frac{N_k}{N-1+\eta}, & \text{if}\  k\leq K\\
  \frac{\eta}{N-1+\eta}, & \text{if}\  k=K +1,
\end{cases}
\end{equation}
$N$ is the number of observations, $N_k$ the number of observations assigned to mixture component $k$, $K$ the number of current mixture components and $\bm{z}_{\setminus i}$ all assignment parameters excluding $z_i$. A Normal Inverse Wishart (NIW) distribution with hyperparameters $\bm{\beta} = \{\bm{m}_0, \kappa_0, \bm{S}_0, \nu_0\} $ is chosen as the prior for the parameters $\bm{\theta}_{G,k}$ and $\bm{\theta}_{C,k}$ of subgoal and constraint region for every mixture component. State subgoals are drawn from the subgoal region's posterior according to the following process
\begin{equation}
\begin{split}
\label{eq:Process_Subgoal_Sampling}
\bm{\beta}_G &= \{\bm{m}_{0,G}, \kappa_{0,G}, \bm{S}_{0,G}, \nu_{0,G}\}\\
{\bm{\mu_G}, \bm{\Sigma_G}} &\sim NIW(\bm{\beta}_G)\\
\bm{g}_k^d&\sim p(\bm{g}_{k\setminus d}, \bm{\mu_G}, \bm{\Sigma_G}).\\     
\end{split}
\end{equation}
The posterior predictive distribution
\begin{equation}
    p(\bm{g}_k^d = \bm{s}_n \vert \bm{g}_{k\setminus d}, \bm{\beta}_G) = \frac{p(\bm{g}_k^d, \bm{g}_{k\setminus d}, \bm{\beta}_G)}{p(\bm{g}_{k\setminus d}, \bm{\beta}_G)}
\end{equation}
has the form of a multivariate Student-T distribution and is chosen as the state subgoal prior, where $\bm{g}_k^d$ is the state subgoal of skill $k$ and demonstration $d$ and $\bm{g}_{k\setminus d}$ are all other state subgoals of skill $k$ excluding $\bm{g}_k^d$. This prior favors state subgoals that are similar to the other subgoals $\bm{g}_{k\setminus d}$ of the same skill from the other demonstrations and reflects our assumption that state subgoals are similar across demonstrations.

\subsubsection{Parameter Inference}
The optimal model parameters, conditioned on the observation, are obtained by MAP estimation $(\bm{z}^*, \bm{g}^*,\bm{\theta}_C^*) = \arg\!\max_{\bm{z}, \bm{g}, \bm{\theta}_C}p(\bm{z}, \bm{g}, \bm{\theta}_C\vert \bm{O})$. Since direct inference from the joint posterior is intractable, but sampling from the conditional distributions (\ref{eq:conditional_z}) and (\ref{eq:conditional_g}) is possible, we use collapsed Gibbs sampling \citep{geman1984stochastic} to approximate the joint posterior. As depicted in Fig.~\ref{fig:BNG-IRL_Inference}, the samples $\bm{z}^{(1:T)}$ and $\bm{g}^{(1:T)}$ drawn from the conditional distributions will converge to the samples from the true posterior (see Algorithm~\ref{alg:collapsed_GS}). 

Due to the conjugacy of its prior, we can marginalize the parameter $\bm{\theta}_C$ and the full conditional of $z_i$ simplifies to
\begin{equation}
\begin{split}
\label{eq:conditional_z}
&\ p(z_i=k\vert \mathbf{O}, \mathbf{z}_{\setminus i}, \bm{g}, \bm{\beta}_C, \eta)\\
\propto&\ p(z_i=k\vert \mathbf{z}_{\setminus i}, \eta)\ p(\mathbf{o}_i\vert \mathbf{O}_{k\setminus i}, \bm{g}_{k}^d, \bm{\beta}_C)\\
=&\ p(z_i=k\vert \mathbf{z}_{\setminus i}, \eta)\ p(\bm{s}_i, \bm{a}_i\vert\bm{g}_k^d)
\ p(\bm{f}_i\vert \mathbf{O}_{k\setminus i}, \bm{\beta}_C)
\end{split}
\end{equation}
where $\mathbf{O}_{k\setminus i}$ = $\{\mathbf{o}_j\vert z_j\!=\!k, j\!\neq\!i\}$ are all observations assigned to cluster $k$, except $\mathbf{o}_i$. The term $p(\bm{f}_i\vert\mathbf{O}_{k\setminus i}, \bm{\beta}_C)$ is the feature clustering's posterior predictive for $\mathbf{o}_i$ of mixture component $k$. The full conditional for $\bm{g}_k^d$ can be expressed by
\begin{equation}
\begin{split}
\label{eq:conditional_g}
&\ p(\bm{g}_k^d=\mathbf{s}_n\vert \mathbf{O}, \mathbf{z}, \bm{\theta}_C)\\
\propto&\ p(\bm{g}_k^d = \bm{s}_n \vert \bm{g}_{k\setminus d}, \bm{\beta}_G)\ p(\mathbf{O}_k^d\vert \bm{g}_k^d=\mathbf{s}_n, \mathbf{z}, \bm{\theta}_C)\\
\propto&\ p(\bm{g}_k^d = \bm{s}_n \vert \bm{g}_{k\setminus d}, \bm{\beta}_G)\sum_{i\in I_k^d}p(\bm{o}_i\vert \bm{g}_k^d=\mathbf{s}_n, \bm{\theta}_{C,k}),
\end{split}
\end{equation}
with the index set $I_k^d = \{i \vert \bm{s}_i \in \bm{\mathrm{T}}_d \wedge  z_i=k\}$. Every $\bm{o}_i$ from the $k$-th skill of the $d$-th demonstration is independent of other skills' subgoals and feature constraints. For a more complete derivation of the above equations, please refer to \cite{murphy2012machine} and \cite{michini2015bayesian}.

\begin{algorithm}
\caption{Inference using collapsed Gibbs sampling}\label{alg:collapsed_GS}
\hspace*{\algorithmicindent}\!\textbf{Input:} $\mathbf{O}$, $\mathbf{z}^0$\\
\hspace*{\algorithmicindent}\!\textbf{Output:}\ samples\ $\mathbf{z}^{(1:T)}$, $\bm{g}^{(1:T)}$
\begin{algorithmic}[1]
\State $\mathbf{z} \gets \mathbf{z}^0$
\For {$i = 1$ to $N$}
    \State $\bm{g} \gets \mathbf{s}_i$
    \State Precompute $Q(\cdot,\cdot, R_g)$, where $R_g(\mathbf{s}_n)=\mathds{1}(\mathbf{s}_n=\bm{g})$
\EndFor
\For {$t = 1$ to $T$}
    \For {$d = 1$ to $D$ \textit{in random order}} \label{l:start_g}
        \For{$k = 1$ to $K$}
            \For {all $\bm{s}_n \in \bm{\mathrm{T}}_d$}
                \State Compute $p(\bm{g}_k^d = \mathbf{s}_n\vert \mathbf{O}, \mathbf{z}, \bm{\theta}_C)$ \Comment{(\ref{eq:conditional_g})}
            \EndFor
            \State Normalize $p(\bm{g}_k^d\vert \mathbf{O}, \mathbf{z}, \bm{\theta}_C)$
            \State Sample $\bm{g}_k^d \sim p(\bm{g}_k^d\vert \mathbf{O}, \mathbf{z}, \bm{\theta}_C)$ \label{l:end_g}
        \EndFor
    \EndFor
    \For {$i = 1$ to $N$ \textit{in random order}} \label{l:start_z}
        \State Remove $\mathbf{o}_i$'s statistics from old cluster $z_i$
        \For{$k = 1$ to $K$}
            \State Compute $p(z_i=k\vert \mathbf{O}, \mathbf{z}_{\setminus i}, \bm{g}, \bm{\beta}, \eta)$ \Comment{(\ref{eq:conditional_z})}
        \EndFor
        \State Compute $p(z_i=K+1\vert \mathbf{O}, \mathbf{z}_{\setminus i}, \bm{g}, \bm{\beta}, \eta)$ \label{l:new_cluster}
        \State Normalize $p(z_i\vert \mathbf{O}, \mathbf{z}_{\setminus i}, \bm{g}, \bm{\beta}, \eta)$
        \State Sample $z_i \sim p(z_i\vert \mathbf{O}, \mathbf{z}_{\setminus i}, \bm{g}, \bm{\beta}, \eta)$ \label{l:end_z}
        \State Add $\mathbf{o}_i$'s statistics to new cluster $z_i$
        \State If any cluster is empty, remove it \& decrease $K$
    \EndFor
\EndFor
\State \textbf{return} samples $\mathbf{z}^{(1:T)}$, $\bm{g}^{(1:T)}$
\end{algorithmic}
\end{algorithm}

Algorithm~\ref{alg:collapsed_GS} starts with initializing $\bm{z}^0$ by assigning every observation to a random mixture component of an initial number of mixture components and precomputing the Q-values for all $N$ subgoal candidates to speed up the subsequent Gibbs sampling process. After that, the inference algorithm iteratively samples state subgoals for the current mixture components and then the component assignment for every observation, conditioned on the latest values of the other parameter. Algorithm~\ref{alg:collapsed_GS} iterates over all demonstrations $D$ and mixture components $K$ in line \ref{l:start_g}-\ref{l:end_g} to evaluate equation (\ref{eq:conditional_g}) for all potential subgoal candidates. For every combination of demonstration $d$ and mixture component $k$, a state subgoal is sampled from the normalized conditional with support over all states in $\bm{\mathrm{T}}_d$. In line \ref{l:start_z}-\ref{l:end_z}, the component assignment for every observation $\bm{o}_i \in \bm{O}$ is sampled from the normalized conditional of $z_i$ with support over all mixture components $K$ and a new mixture component $K+1$. In each sampling iteration, every $\bm{o}_i$ can thus be assigned to a new mixture $K+1$ with a random subgoal with a probability determined by line~\ref{l:new_cluster}. If a mixture has no observation from one of the demonstrations assigned to it, this mixture component is removed. After each iteration, a post-processing step is performed if the number of mixture components changed.

\subsection{Task Model Learning}
\label{sec:Models}
As described in the previous sections, we propose to learn both the low- and high-level models of a task. Decomposing the task into its individual skills and structuring them in a task graph has several advantages. 

Especially when learning low-level policies from demonstration, segmenting the task into skills reduces the problem of accumulating errors caused by training data that are irrelevant for the current part of a task \cite{shi2023waypoint}. For dynamical system-based motion generation and reinforcement learning policies, the decomposition additionally has the advantage that the same state input can be mapped to different actions if different skills are active. Without the decomposition, more complicated policies would be required to achieve that behavior. The BNG-IRL segmentation algorithm provides us with a subgoal region $\mathcal{N}(\bm{\theta}_{G,k})$ and a constraint region $\mathcal{N}(\bm{\theta}_{C,k})$ for every skill $k$. Together with the observation set $\bm{O}_k=\{\bm{o}_i \vert z_i=k\}$, which is used to learn the motion primitive for a skill, this is all the information needed to execute the low-level skills on a robot. The mechanism for the monitored execution will be explained in Sec.~\ref{sec:Execution}. As described in Sec.~\ref{sec:skill-refinement-ts} and Sec.~\ref{sec:User_Support}, the parameters $\bm{\theta}_{C,k}$ and the motion primitives of every skill can be further refined with new training data collected during the refinement phase.

At the higher level, combining the task graph structure with a subgoal for every skill enables the system to monitor task progress. The next skill in the task graph is only scheduled once the subgoal of the current skill is reached. If an anomaly is detected, the system can trace it back to the specific skill where the error occurred, enhancing the explainability of the issue and allowing the robot to respond automatically. The task graph also narrows the margin for error in high-level decision-making by limiting choices to those relevant for the current skill. After new user demonstrations were segmented with BNG-IRL, the skill sequence is used to update the task graph. The skills inferred from the initial teaching sequence are used to create an initial task graph. This skill sequence can be incrementally extended with task decisions and recovery behaviors as described in Sec.~\ref{sec:task-decision-ts}. If the unsupervised anomaly detection approach recognizes deviations from the skill's intended execution, the skill selection module needs to determine if an appropriate recovery behavior has already been demonstrated to automatically resolve the situation, or if a new task decision teaching sequence needs to be triggered. The different components of the execution module for this functionality are explained in the following section.

\section{Autonomous Task Execution}
\label{sec:Execution}
As depicted on the right in Fig.~\ref{fig:Overview}, the execution monitoring module combines several submodules for motion generation, unsupervised anomaly detection, subgoal monitoring, and skill selection. The skill selection module acts as an interface between the task model and the execution modules. We present the individual modules and their interplay in this section. 
\subsection{Motion Generation and Unsupervised Anomaly Detection}
\label{sec:motion-generation-anomaly-detection}
In principle, all classical approaches for learning motion primitives, learning policies via Reinforcement Learning, motion planners or even a combination of them are suited to be used in the motion generation module. We present a data-efficient dynamical systems-based approach using Gaussian Mixture Models (GMM) and Gaussian Mixture Regression (GMR) to learn motion primitives and feature constraints for every contact skill from a few demonstrations. We argue that the behavior of a skill depends on the configuration of the robot relative to important objects to interact with and not on time. The aspect of time or sequentiality plays rather a role in the higher-level representation of the task, where subgoals have to be reached one after another in order to continue with the next skill. That is why we propose to generate a velocity and force command based on the robot's measured end-effector pose.

For every skill, we thus construct a training data set $\{\bm{s}_i,\bm{\xi}_i{\rbrace}_{i=1}^{N_{k}}$ from all $\bm{o}_i \in \bm{O}_k$. $\bm{s}_i$ is the end-effector pose and $\bm{\xi}_i=[\bm{a}_i, \bm{f}_i]$ the corresponding vector of translational and rotational end-effector velocity and contact force. The training data set is encoded as a GMM, estimating the joint probability distribution of the data as a weighted sum of $E$ independent Gaussian components
\begin{equation}
\begin{bmatrix}
\bm{s} \\ \bm{\xi}
\end{bmatrix} \sim \sum_{e=1}^{E}\pi_e\mathcal{N}(\bm{\mu}_e, \bm{\Sigma}_e), 
\end{equation}
where $\pi_e, \bm{\mu}_e, \bm{\Sigma}_e$ are the prior probability, mean, and covariance matrix of the $e$-th Gaussian component. We can decompose the $e$-th mean vector and covariance matrix
\begin{equation}
    \bm{\mu}_e = \begin{bmatrix} \bm{\mu}_e^s \\ \bm{\mu}_e^{\xi} \end{bmatrix},
    \bm{\Sigma}_e = \begin{bmatrix} \bm{\Sigma}_e^{ss} & \bm{\Sigma}_e^{s\xi} \\ \bm{\Sigma}_e^{\xi s} & \bm{\Sigma}_e^{\xi\xi} \end{bmatrix}
\end{equation}
into input and output components corresponding to $\bm{s}$ and $\bm{\xi}$, respectively. GMR predicts the most likely output vector for a given input by computing the posterior probability distribution $P(\hat{\bm{\xi}}_n|\bm{s}_{n}) = \mathcal{N}(\hat{\bm{\xi}}_n|\hat{\bm{\mu}}^{\xi}(\bm{s}_n), \hat{\bm{\Sigma}}^{\xi\xi}(\bm{s}_n))$ conditioned on the input. Using the measured end-effector pose $\bm{s}_n$ as input during execution, GMR determines the expected output end-effector velocity and force $\mathbb{E}(\hat{\bm{\xi}}_n|\bm{s}_n) = \hat{\bm{\mu}}^{\xi}(\bm{s}_n)$ along with a covariance matrix $\hat{\bm{\Sigma}}^{\xi\xi}(\bm{s}_n)$. For simplicity, we will refer to the predicted mean and covariance matrix as $\hat{\bm{\mu}}_n^{\xi}$ and $\hat{\bm{\Sigma}}_n^{\xi\xi}$, respectively. The hat symbol indicates variables that are dynamically computed at each cycle. The expected output vector can be computed with
\begin{equation}
\label{eq:output_vetor}
    \hat{\bm{\mu}}_n^{\xi} = \sum_{e=1}^{E} \hat{h}_e(\bm{s}_n)\underbrace{\left(\bm{\mu}_e^{\xi} + \bm{\Sigma}_e^{\xi s}(\bm{\Sigma}_e^{ss})^{-1}(\bm{s}_n - \bm{\mu}_e^{s})\right)}_{\hat{\bm{\mu}}_e^{\xi}(\bm{s}_n)},
\end{equation}
where
\begin{equation}
\label{eq:hc}
    \hat{h}_e(\bm{s}_n) = \frac{\pi_e \mathcal{N}(\bm{s}_n \vert \bm{\mu}_e^s, \bm{\Sigma}_e^{ss})}{ \sum_{j=1}^{E}\pi_j \mathcal{N}(\bm{s}_n \vert \bm{\mu}_j^s, \bm{\Sigma}_j^{ss})}.
\end{equation}
Starting with an initial end-effector pose $\bm{s}_0$, we can thus predict an initial velocity and force command $\hat{\bm{\mu}}_0^{\xi}$ and send it to the robot for execution. In the next cycle $n+1$, we measure the end-effector pose again and repeat the process. This results in a dynamical system with state-based force overlay. 

We leverage the conditional covariance matrix $\hat{\bm{\Sigma}}_n^\mathrm{\xi\xi}$ for unsupervised anomaly detection. The covariance matrix is computed with
\begin{gather*}
    \hat{\bm{\Sigma}}_n^{\xi\xi} = \sum_{e=1}^{E} \hat{h}_e(\bm{s}_n)\left(\Tilde{\bm{\Sigma}}_e^{\xi\xi}+\hat{\bm{\mu}}_e^{\xi}(\bm{s}_n)\hat{\bm{\mu}}_e^{\xi}(\bm{s}_n)^T\right)-\hat{\bm{\mu}}_n^{\xi}(\hat{\bm{\mu}}_n^{\xi})^T,
\end{gather*}
where $\Tilde{\bm{\Sigma}}_e^{\xi\xi} = \bm{\Sigma}_e^{\xi s}(\bm{\Sigma}_e^{ss})^{-1}\bm{\Sigma}_e^{s\xi}$. Using the measured end-effector velocity and force vector $\bm{\xi}_{n}$ in cycle $n$, we compute the Mahalanobis distance to quantify its deviation towards the commanded output vector~(\ref{eq:output_vetor}) with
\begin{equation}
    \hat{D}_{\mathrm{M}}(\bm{\xi}_{n}) = \sqrt{(\bm{\xi}_{n}-\hat{\bm{\mu}}_n^{\xi})^\mathrm{T}(\hat{\bm{\Sigma}}_n^{\xi\xi})^{-1}(\bm{\xi}_{n}-\hat{\bm{\mu}}_n^{\xi})}.
\label{eq:anomaly_mahal_dist}
\end{equation}
To determine if the deviation is within the expected constraint region for the skill or if it constitutes an anomaly, we set
\begin{equation}
\label{eq:P_A}
\hat{P}_n^\mathrm{A}= \left\{
\begin{array}{cl}
    \text{anomaly}, & \text{if } \hat{D}_{\mathrm{M}}(\bm{\xi}_{n}) > D_{\mathrm{M, max}}\\
    \text{no anomaly}, & \text{if } \hat{D}_{\mathrm{M}}(\bm{\xi}_{n}) \leq D_{\mathrm{M, max}} \\
\end{array}\right.
\end{equation}
where $D_{\mathrm{M, max}} = \max_{\bm{\xi}_i \in \bm{O}_k} D_{\mathrm{M}}(\bm{\xi}_i)$ is the maximum Mahalanobis distance observed in the non-anomalous training data set $\bm{O}_k$. However, this model is not suited to produce anomaly predictions for end-effector poses $\bm{s}$ far away from the training data. To quantify the confidence in the anomaly prediction, we therefore use $\hat{P}(\bm{s}_n) = \sum_{e=1}^{E}\pi_e\mathcal{N}(\bm{s}_n \vert \bm{\mu}_e^s, \bm{\Sigma}_e^{ss})$, which decreases as the model is queried for inputs $\bm{s}_n$ further away from the training data and is already computed in (\ref{eq:hc}) as part of GMR in cycle $n$. Similar to determining an anomaly, we set 
\begin{equation}
\label{eq:P_C}
\hat{P}_n^\mathrm{C}= \left\{
\begin{array}{cl}
    \text{confident}, & \text{if } \hat{P}(\bm{s}_n) \geq P_\mathrm{min}(\bm{s})\\
    \text{not confident}, & \text{if } \hat{P}(\bm{s}_n) < P_\mathrm{min}(\bm{s}) \\
\end{array}\right.,
\end{equation}
where $P_\mathrm{min}(\bm{s}) = \min_{\bm{s}_i \in \bm{O}_k} P(\bm{s}_i)$.

This results in a two-step process for detecting anomalies. Only if the algorithm determines with $\hat{P}_n^\mathrm{C}$ that the robot is within the known region where an anomaly can be identified confidently, the prediction of $\hat{P}_n^\mathrm{A}$ is considered. If $\hat{P}_n^\mathrm{A}$ is confidently classified as \textit{anomaly} for $\varepsilon$ consecutive cycles, an anomaly is detected. Otherwise, if $\hat{P}_n^\mathrm{C}$ reports \textit{not confident} for $\varepsilon$ consecutive cycles, the robot switches to the refinement phase as described in Sec.~\ref{sec:skill-refinement-ts}, where it continues with the task execution under the supervision of the user who can assist or stop the robot at any time. The collected training data is then used to refine the skill model. It is worth noting that this approach distinguishes between two different sources of uncertainty in the two-step process. The first step considers the uncertainty caused by the lack of knowledge due to missing training data, also referred to as epistemic uncertainty. This uncertainty can be reduced by collecting additional training data during the refinement phase. The second step considers the aleatoric uncertainty, which refers to the uncertainty caused by variation in the training data. $\hat{\bm{\Sigma}}_n^\mathrm{\xi\xi}$ encodes the variability between the demonstrations and correlations among the elements of the output vector $\bm{\xi}$ at $\bm{s}_n$. Using the Mahalanobis distance together with the fixed decision threshold $D_{\mathrm{M, max}}$ causes the anomaly detection algorithm to be more sensitive in areas with a low variance of $\bm{\xi}$ in the training data set, expressed by small values on the diagonal of $\hat{\bm{\Sigma}}_n^\mathrm{\xi\xi}$. Differentiating between the epistemic and aleatoric uncertainty during the anomaly detection and reacting according to the source of the uncertainty makes our approach unique compared to state-of-the-art anomaly detection methods that fail to distinguish between the two sources \citep{kiureghian2009aleatoric}.
\subsection{Subgoal Monitoring}
\label{sec:subgoal-monitoring}
The last low-level execution module takes care of the subgoal monitoring. It runs in parallel to the other modules and constantly checks if the skill has reached its subgoal region $\bm{G}_k=\mathcal{N}(\bm{\mu}_{G,k}, \bm{\Sigma}_{G,k})$, as defined in Sec.~\ref{sec:segmentation_model}. We use a similar idea as in the anomaly detection approach to check if the end-effector pose $\bm{s}_n$ is within the expected goal region. We compute the Mahalanobis distance
\begin{equation}
    \hat{D}_{\mathrm{M}}^{G_k}(\bm{s}_n) = \sqrt{(\bm{s}_n-\bm{\mu}_{G,k})^\mathrm{T}(\bm{\Sigma}_{G,k})^{-1}(\bm{s}_n-\bm{\mu}_{G,k})}
\end{equation}
and set
\begin{equation}
\hat{P}_n^{G_k}= \left\{
\begin{array}{cl}
    \text{subgoal reached}, & \text{if } \hat{D}_{\mathrm{M}}^{G_k}(\bm{s}_n) \leq D_{\mathrm{M, max}}^{G_k}\\
    \text{subgoal not reached}, & \text{if } \hat{D}_{\mathrm{M}}^{G_k}(\bm{s}_n) > D_{\mathrm{M, max}}^{G_k} \\
\end{array}\right.
\label{eq:subgoal_reached_condition}
\end{equation}
where $D_{\mathrm{M, max}}^{G_k} = \max_{d \in D} D_{\mathrm{M}}^{G_k}(\bm{g}_k^d)$ is the maximum Mahalanobis distance of the state subgoals $\bm{g}_k^d$ inferred for the skill. If the subgoal region is reached, the skill selection component takes care of transitioning to the subsequent skill in the task graph.
\subsection{Skill Selection}
\label{sec:skill-selection}
The skill selection module is triggered by events from either the anomaly detection or subgoal monitoring module and determines how to react to the incoming events based on the task model. If the subgoal of a skill $k$ is reached, the skill selection module forwards the corresponding low-level skill parameters of the next skill $k+1$ of the intended task flow from the task graph to the execution components or terminates the execution successfully if skill $k$ is a termination state in the task graph.

\begin{algorithm}
\caption{Determine type of anomaly sequence ${\lbrace}\bm{\xi}_n{\rbrace}_{n=1}^{\varepsilon}$}\label{alg:determine_anomaly_label}
\hspace*{\algorithmicindent}\!\textbf{Input:} ${\lbrace}\bm{\xi}_n{\rbrace}_{n=1}^{\varepsilon}$, $\bm{\Xi}_k^\mathrm{A}$, $L_k$\\
\hspace*{\algorithmicindent}\!\textbf{Output:} anomaly label $l\in L_k$
\begin{algorithmic}[1]
\If {$ \bm{\Xi}_k^\mathrm{A} == \emptyset $}
        \State \textbf{set} $\bm{\Xi}_k^\mathrm{A} = {\lbrace}\bm{\xi}_n{\rbrace}_{n=1}^{\varepsilon}$
        \State \textbf{set} $l=1$, $L_k=\{l\}$
        \State \textbf{return} $l$
    \EndIf
\State $c = 0$
\For {$n = 1$ to $\varepsilon$}
    \State compute $P(\bm{\xi}_n) = \sum_{m=1}^{M}\pi_m\mathcal{N}(\bm{\xi}_n \vert \bm{\mu}_m^\mathrm{A}, \bm{\Sigma}_m^\mathrm{A})$
    \If {$P(\bm{\xi}_n) < P_\mathrm{min}(\bm{\xi})$}
        \State \textbf{set} $c = c+1$
    \EndIf
\EndFor
\If {$c > \varepsilon/2$}
    \State query user to confirm new anomaly
    \If {new anomaly}
    \State initiate new \textit{Task Decision} teaching sequence
    \State \textbf{set} $l=\max_{l \in L_k} l +1 $, $L_k= L_k \cup l$
    \EndIf
\Else
    \For {$n = 1$ to $\varepsilon$}
    \State predict $l=f(\bm{\xi}_n)$ and increase count of $l$
\EndFor
\State \textbf{set} $l = l$ with max count
\EndIf
\State \textbf{set} $\bm{\Xi}_k^\mathrm{A} = \bm{\Xi}_k^\mathrm{A} \cup {\lbrace}\bm{\xi}_n{\rbrace}_{n=1}^{\varepsilon}$
\State update GMM with $\bm{\Xi}_k^\mathrm{A}$ and classifier with ${\lbrace}\bm{\xi}_n{\rbrace}_{n=1}^{\varepsilon}$, $l$
\State \textbf{return} $l$
\end{algorithmic}
\end{algorithm}

In case an anomaly is detected during skill $k$, the skill selection module stops the current execution and needs to determine if an appropriate recovery behavior has already been learned for that skill and anomaly case or if a new recovery behavior for a new type of anomaly needs to be learned from demonstration (see Algorithm~\ref{alg:determine_anomaly_label}). To check if the detected anomaly is not within the region of known anomalies for skill $k$, we train a GMM: $\bm{\xi} \sim \sum_{m=1}^{M}\pi_m\mathcal{N}(\bm{\mu}_m^\mathrm{A}, \bm{\Sigma}_m^\mathrm{A})$ on the set $\bm{\Xi}_k^\mathrm{A} = {\lbrace}{\lbrace}\bm{\xi}_{i,j}{\rbrace}_{i=1}^{\varepsilon}{\rbrace}_{j=1}^J$ of samples recorded during the last $\varepsilon$ cycles before an anomaly was triggered from all previous anomaly occurrences $J$. To incrementally refine and extend the knowledge about possible anomalies, the set $\bm{\Xi}_k^\mathrm{A}$ is extended when new anomalies are detected during skill $k$. We adopt step one of the anomaly detection strategy (\ref{eq:P_C}) based on the notion of epistemic uncertainty to detect new anomaly cases. We compute $P(\bm{\xi}_n) = \sum_{m=1}^{M}\pi_m\mathcal{N}(\bm{\xi}_n \vert \bm{\mu}_m^\mathrm{A}, \bm{\Sigma}_m^\mathrm{A})$ for all $\bm{\xi}_n, n \in \{1, \hdots, \varepsilon\}$ that were confidently classified as anomalies with (\ref{eq:P_A}). If an anomaly is not within the region of known anomalies, hence $P(\bm{\xi}_n) < P_\mathrm{min}(\bm{\xi})$ for the majority of samples $\bm{\xi}_n$ with $P_\mathrm{min}(\bm{\xi}) = \min_{\bm{\xi}_i \in \bm{\Xi}_k^A} P(\bm{\xi}_i)$, the user is queried to confirm whether it is a new anomaly. In that case, a new recovery behavior is demonstrated and appended to skill $k$. In future occurrences of the same type of anomaly, the recovery behavior can be leveraged to autonomously recover. To distinguish the known anomaly cases, the observations $\bm{\xi}_n$ during the last $\varepsilon$ cycles before the anomaly is triggered, are used to learn a classifier $f(\bm{\xi})=l$. The classifier assigns a sample $\bm{\xi}_n$ to a class $l \in L_k$ of the known anomaly types $L_k$ of skill $k$. We use a Support Vector Machine with a sliding time window for that purpose in our experiments.

\section{Experiments}
\label{sec:Experiments}
We conducted experiments in simulation and on two different robots to evaluate the individual aspects of our proposed approach. First, we implemented a simple box-pushing task in simulation to compare our segmentation approach to other state-of-the-art methods. The other experiments are contact-based manipulation tasks conducted on real robots, where we increase the complexity of the task to show the applicability of the entire framework, including autonomous detection and recovery from anomalies in a real-world scenario. Supplementary videos for the experiments are provided in Extensions 1-3.
\subsection{Box Pushing in Simulation}
\label{sec:Box_Pushing_Exp}
In this experiment, we conduct an ablation study to examine in detail the influences of the IRL-based intention recognition and GMM-based feature clustering on our BNG-IRL segmentation approach.
\subsubsection{Setup and Task Description}
\label{sec:setup_box_pushing}
\begin{figure}[thpb]
    \centering
    \includegraphics{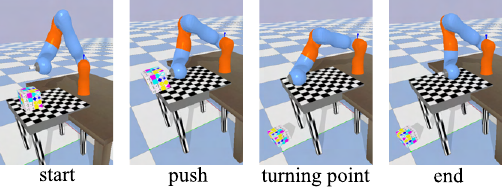}
    \captionsetup{belowskip=-1pt}
    \caption{Steps of the box pushing task demonstration.}
    \label{fig:Push_Box}
\end{figure}

\begin{figure*}[thpb]
  \centering  
  \begin{tabular}{cccc}
  \vspace{-10pt}
    & \large BNG-IRL (ours) & \large{BNGMM} & \large{BN-IRL} \\
    \vspace{-7pt}
    \rotatebox{90}{\qquad\quad\parbox{3cm}{Cartesian EEF \\Trajectory}} &
    \includegraphics[clip, trim=0 5 0 15,scale=0.24]{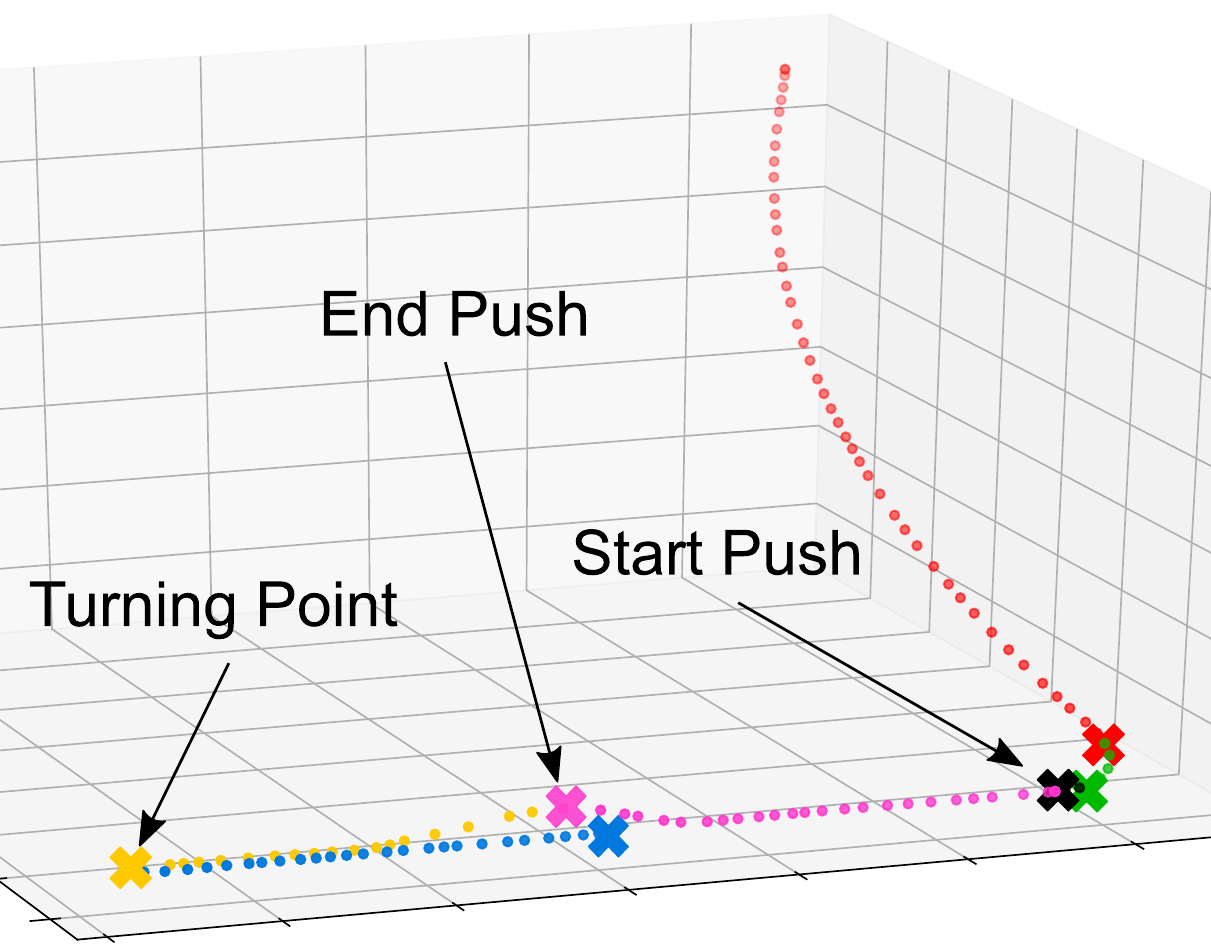} & 
    \includegraphics[clip, trim=0 5 0 10,scale=0.24]{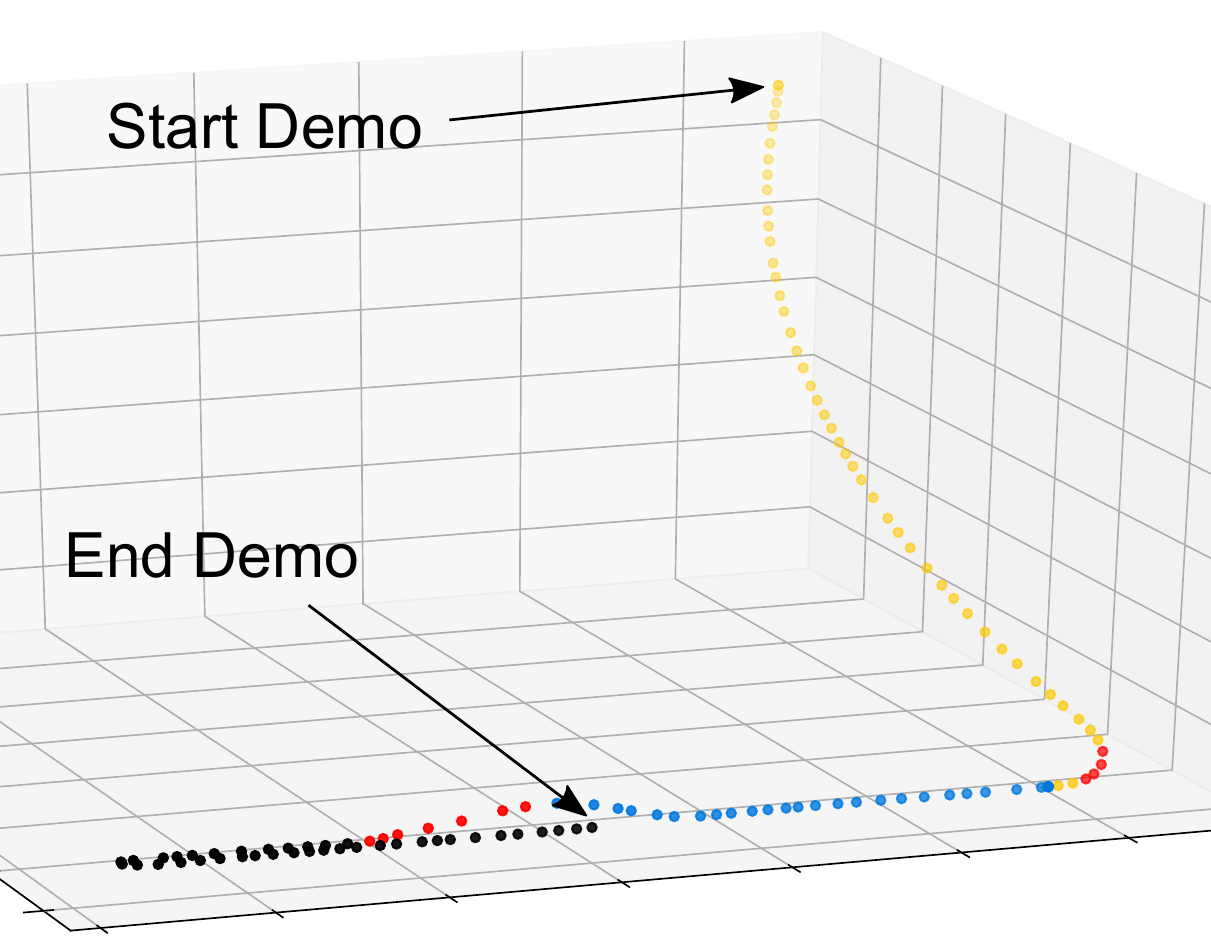} & 
    \includegraphics[clip, trim=0 5 0 15,scale=0.24]{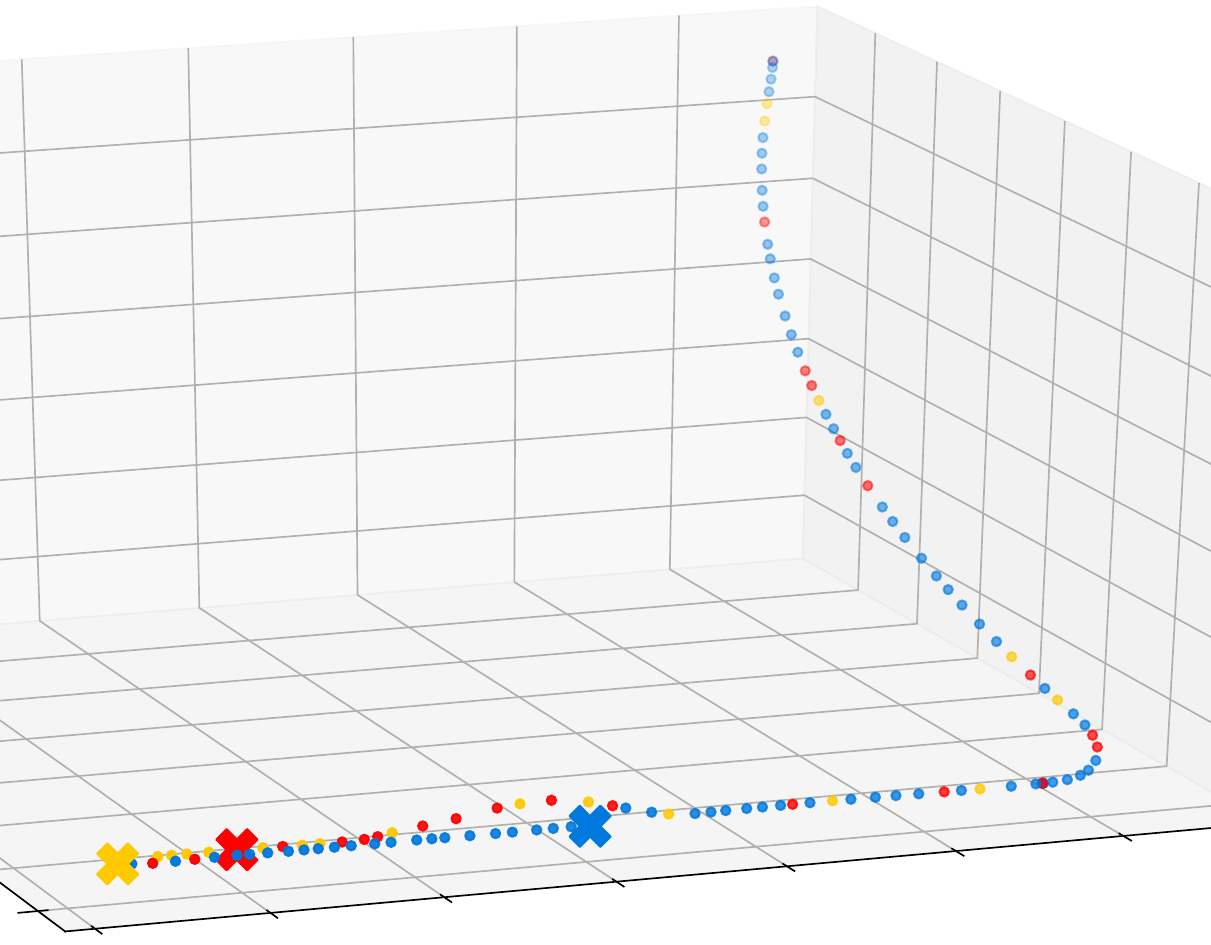} \\
    \rotatebox{90}{\quad Distance to Edge}&
    \includegraphics[clip, trim=120 0 135 80,scale=0.115]{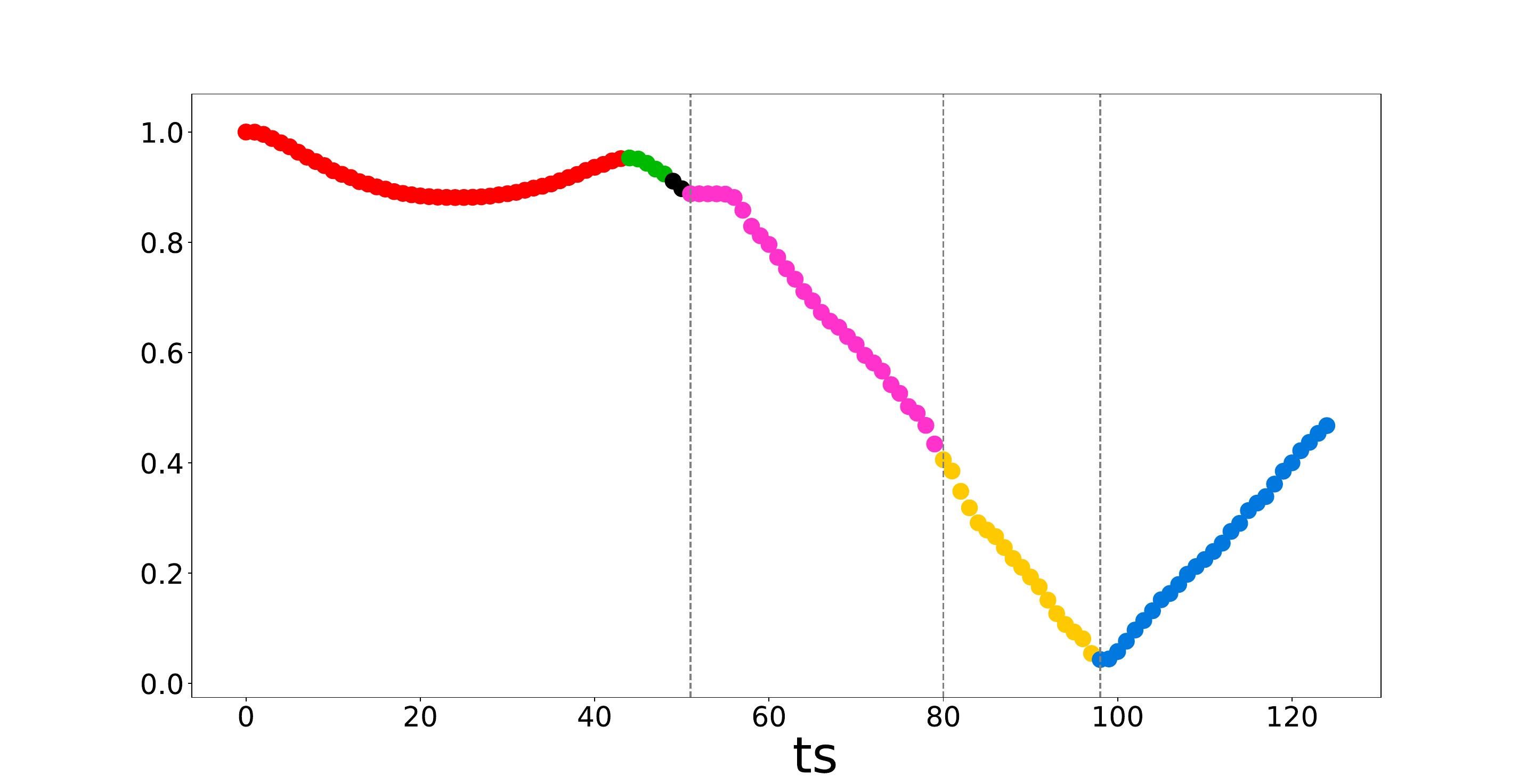} & 
    \includegraphics[clip, trim=120 0 135 80, scale=0.115]{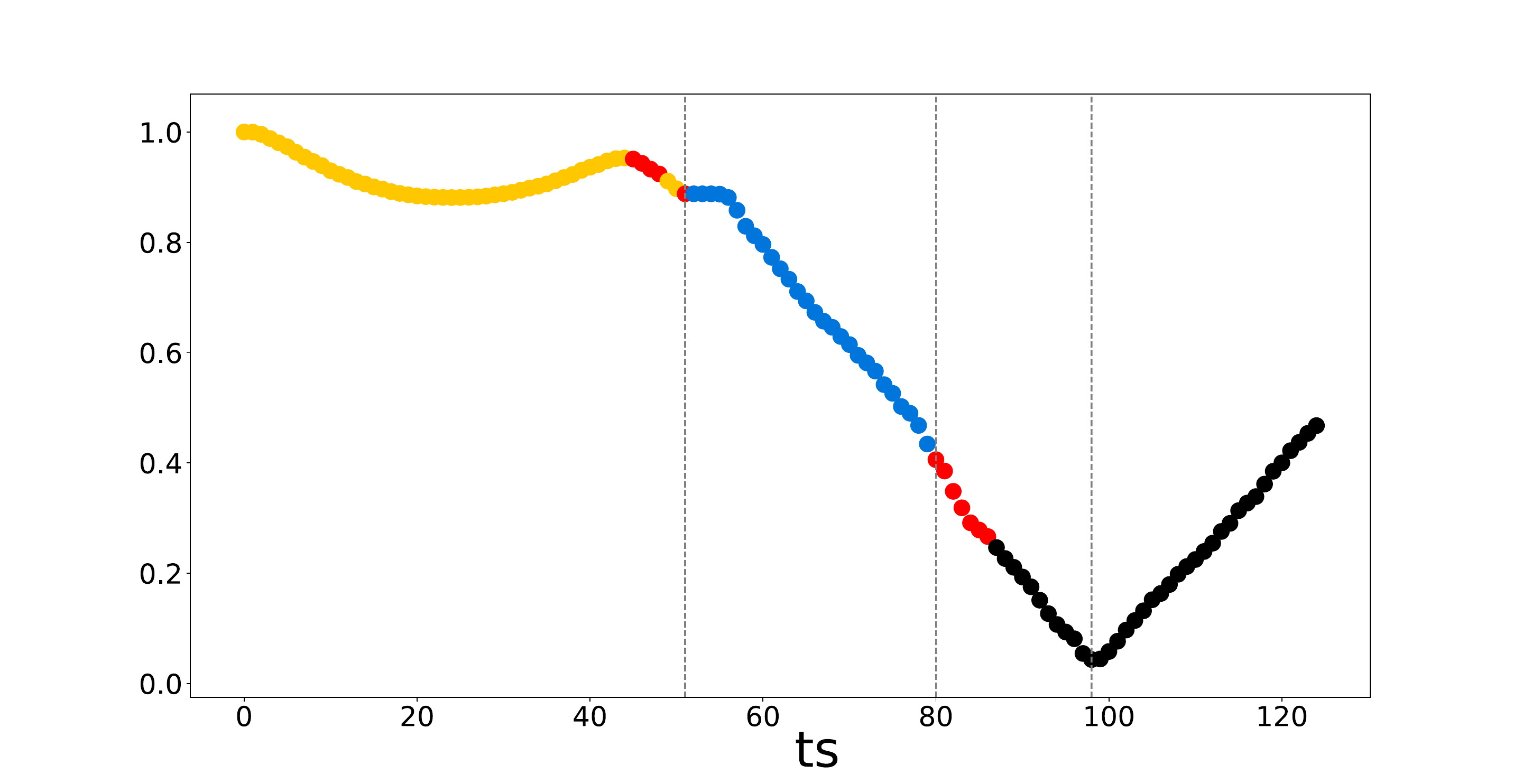} & 
    \includegraphics[clip, trim=120 0 135 80, scale=0.115]{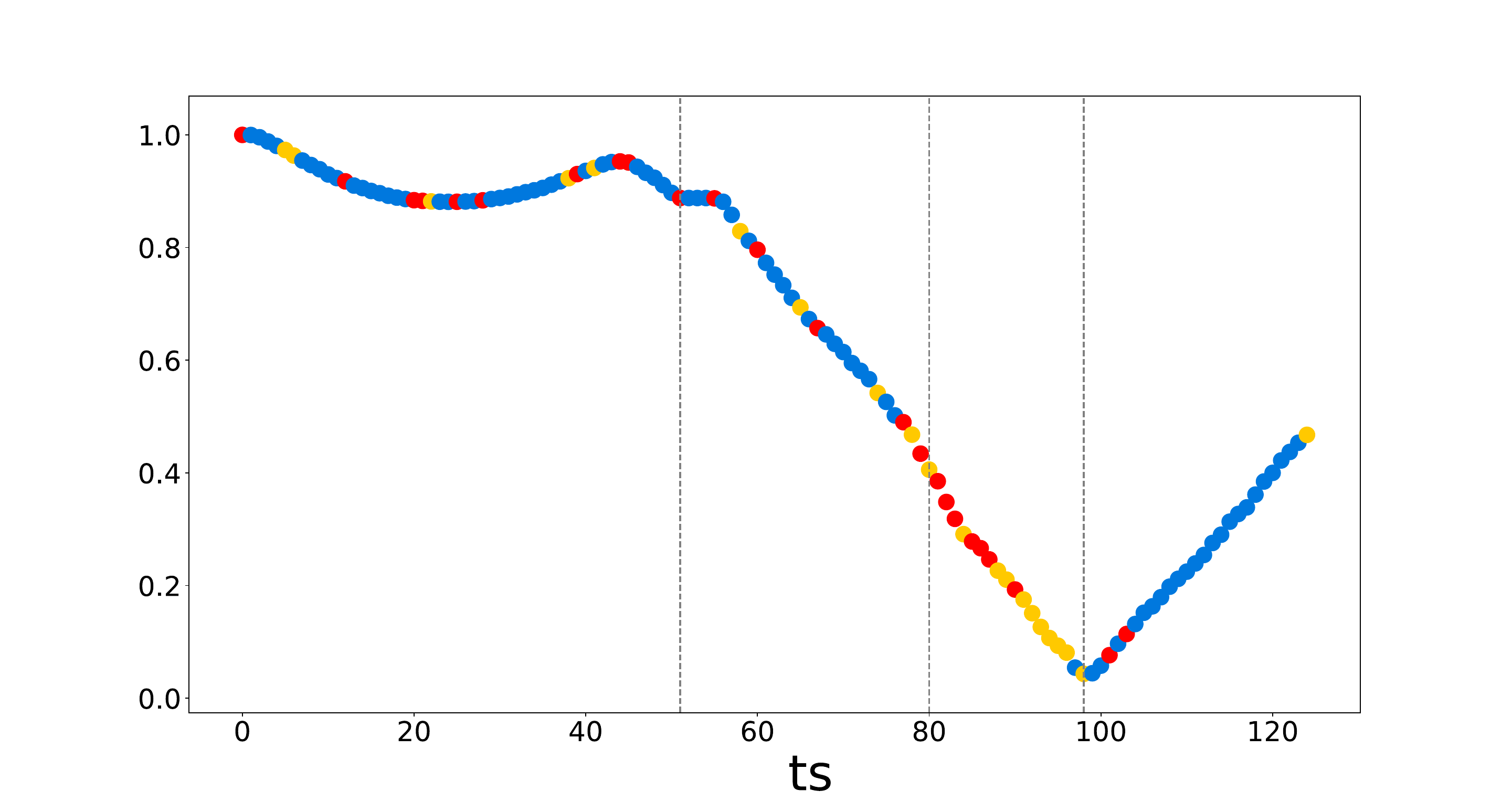}\\
    \rotatebox{90}{\quad Force on EEF}&
    \includegraphics[clip,  trim=120 0 135 80, scale=0.115]{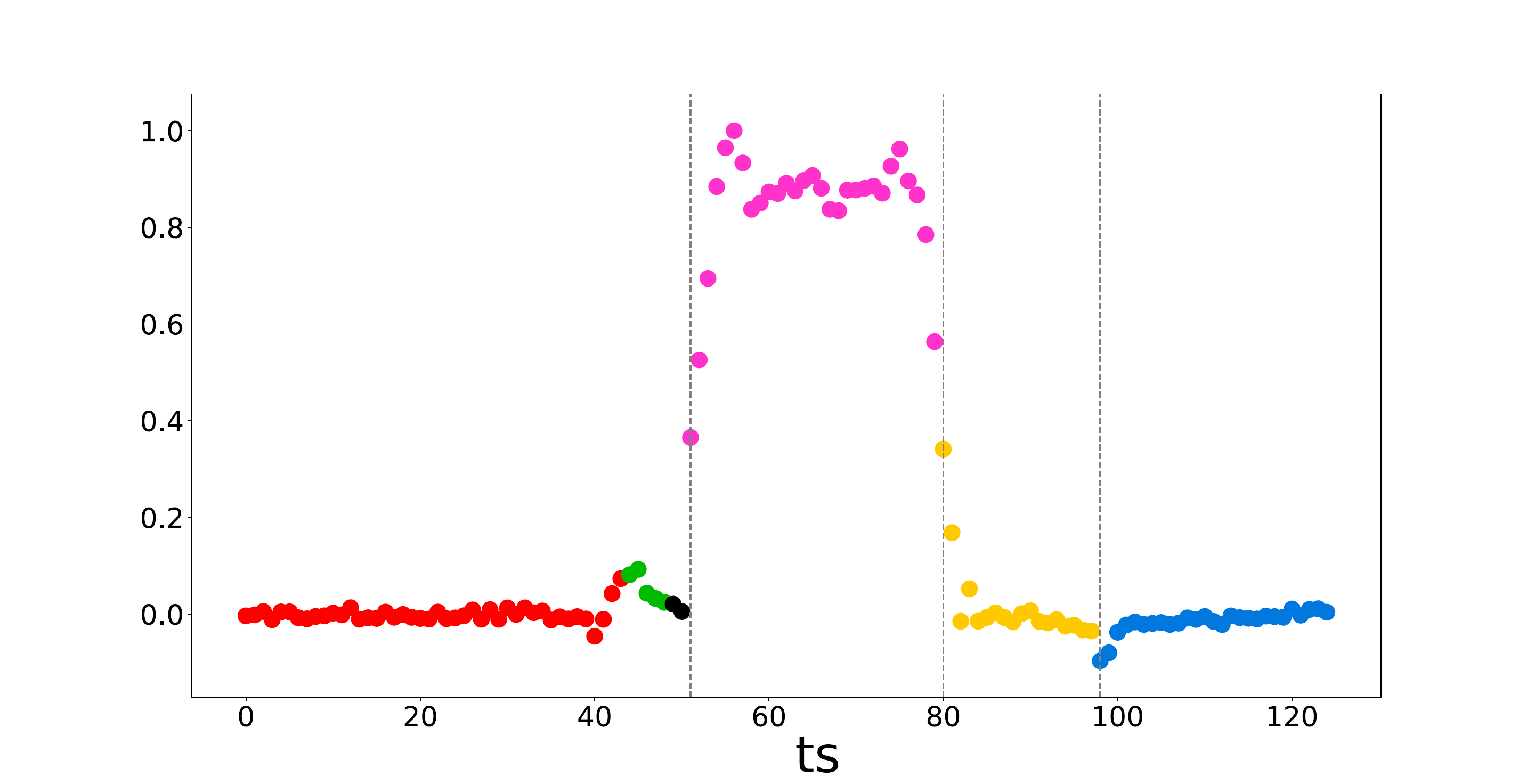} & 
    \includegraphics[clip,  trim=120 0 135 80, scale=0.115]{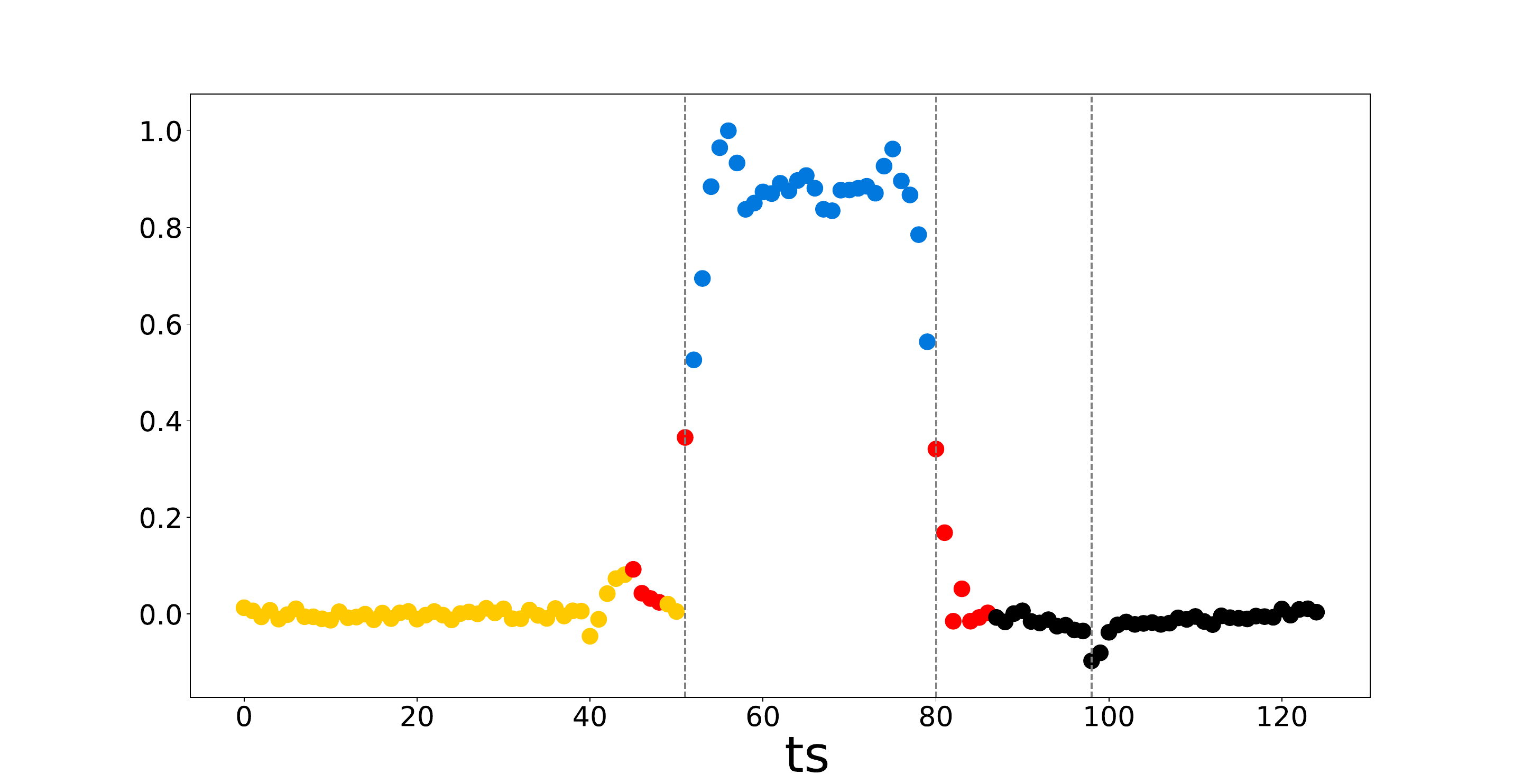} & 
    \includegraphics[clip,  trim=120 0 135 80, scale=0.115]{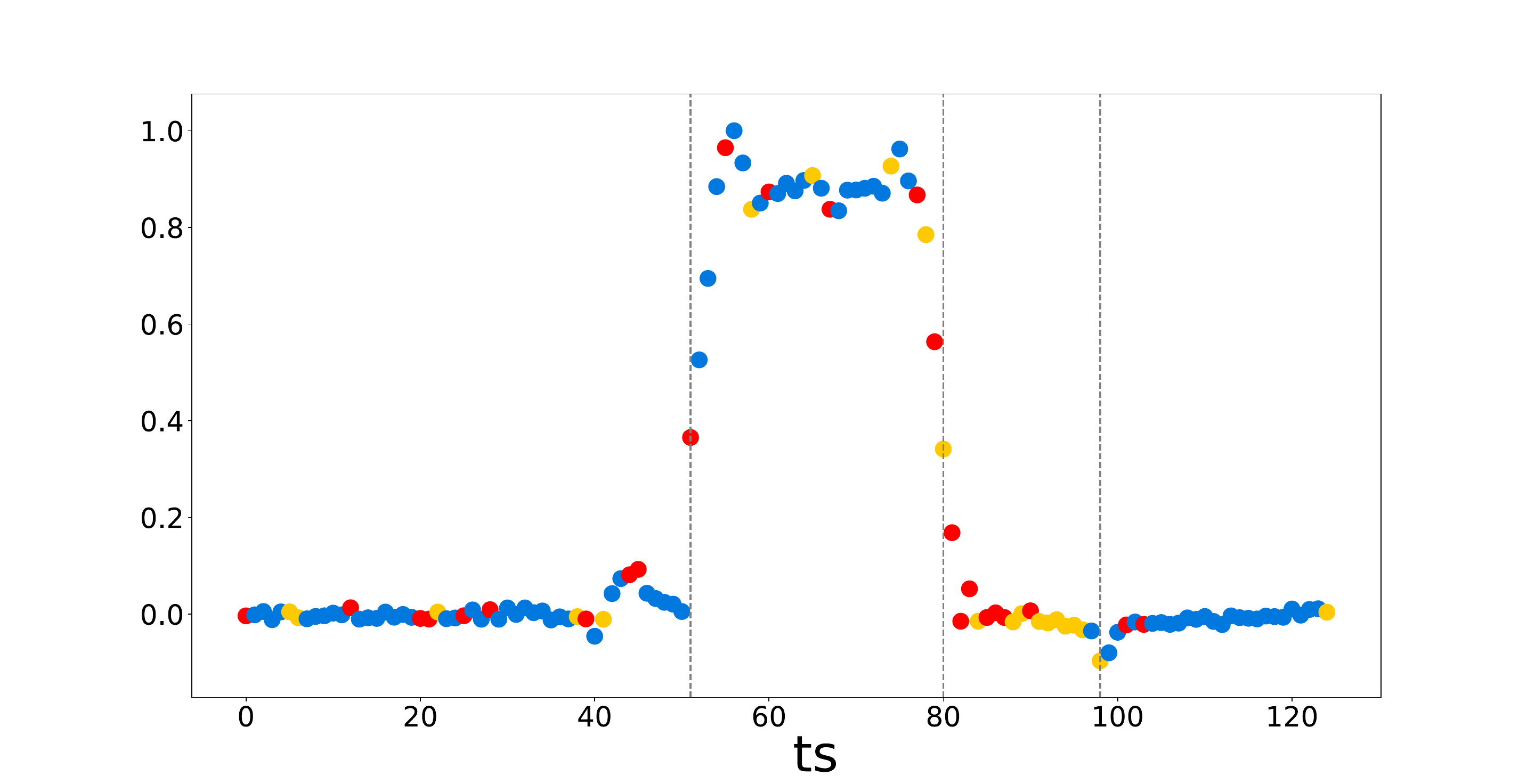}\\
  \end{tabular}
  \captionsetup{belowskip=-10pt}
  \caption{The top row shows the segmentation result of the three compared methods BNG-IRL, BNGMM \citep{rasmussen1999infinite}, and BN-IRL \citep{michini2015bayesian}, mapped onto the end-effector (EEF) trajectory of the task demonstration. The skill assignment $z$ is represented by the color of each sample. The subgoals, which are only inferred in our approach and BN-IRL, are represented by an X in the skill's color. The next two rows show the segmentation results in feature space, where the normalized distance of the EEF to the edge of the table and the normalized force on the EEF are plotted over time. The three distinct events during the task demonstration: The start and end of the box pushing as well as the turning point after pushing are highlighted in the figures. It can be seen, that by combining the influence of subgoals and feature constraints, only our approach reliably detects the distinct phases of the task and groups them in individual skills. The sampling-based parameter inference leading to the results is shown in Extension~1.}
  \label{tab:comparison_box_pushing}
\end{figure*}

The setup is shown in Fig.~\ref{fig:Push_Box}, where we simulate a user demonstration of a pushing task. 
The robot starts from an arbitrary start configuration and moves toward the box on the table. The box is then pushed towards the edge of the table with the robot until it falls on the floor. After the box has fallen from the table, the robot moves until a turning point and retracts back towards the table. The demonstrated end-effector (EEF) trajectory can be seen in the upper row of Fig.~\ref{tab:comparison_box_pushing}. The state space $\bm{S}$ of this task is the Cartesian space $[x, y, z] \in \mathbb{R}^3$ and the feature space $\bm{\mathcal{F}}\in \mathbb{R}^3$ is defined by the Euclidean distances of the robot's EEF to the box and to the edge of the table over which the box will be pushed as well as the force acting on the EEF. The latter two features recorded during the demonstration can also be seen in Fig.~\ref{tab:comparison_box_pushing}.
\subsubsection{Unsupervised Task Segmentation Baselines}
\label{sec:comparison_box_pushing}

\textbf{Bayesian nonparametric GMM (BNGMM)} \citep{rasmussen1999infinite} is a probabilistic approach to clustering multivariate observations. The Bayesian nonparametric formulation allows to adapt the complexity of the model, i.e. the number of clusters, to the data.

\textbf{BN-IRL} \citep{michini2015bayesian} tries to explain a task as a sequence of subgoals, which are important states encountered during the user demonstration. BN-IRL focuses only on inferring subgoals in the state space and does not take the features $\bm{\mathcal{F}}$ into account.

\textbf{Automatic Waypoint Extraction (AWE)} \citep{shi2023waypoint} uses dynamic programming to automatically extract a demonstration’s minimal set of waypoints that approximate the demonstrated trajectory well enough when linearly interpolating between them such that the reconstruction error lies below a predefined error threshold. The reconstruction error is defined as the maximum over the shortest distances of any point on the original trajectory to the reconstructed trajectory.

\textbf{Bayesian Online Changepoint Detection (BOCPD)} \cite{sugawara2023unsupervised} apply BOCPD to the time derivative of measured force and torque signals to detect characteristic changes in the contact situation between a robot with its environment.

\subsubsection{Metrics}
\label{sec:segmentation_metrics}
To evaluate the segmentation performance of the different methods, we compute the frame-wise accuracy (Acc), the edit score (Edit), and F1 scores at overlap thresholds of 10\%, 25\%, and 50\% (F1@{10, 25, 50}). The accuracy evaluates performance at the frame level, while the edit score and F1 scores assess segmentation quality at the segment level \citep{liu2023diffusion}. We calculate the overlap between each detected segment and the ground truth segments and assign ground truth labels to maximize the overall intersection rate across the entire task. 

\subsubsection{Results and Discussion}
\label{sec:results_discussion_box_pushing}

The most likely segmentation results for BNG-IRL, BNGMM, and BN-IRL after 1000 Gibbs sampling steps are shown in Fig.~\ref{tab:comparison_box_pushing}. BNG-IRL is the only implementation to reliably identify the different skills of the task, which are represented by the different colors and corresponding subgoals marked as \textbf{X}. The task consists of four phases: 1) approaching the box and aligning the EEF, 2) pushing the box from the table, 3) moving forward to the turning point without load, and 4) retracting back towards the table. As reported in Table~\ref{tab:segmentation_evaluation_box_pushing}, BNG-IRL achieves the highest frame-wise accuracy. While all ground truth segments are correctly detected, false positive segmentations before the pushing phase slightly lower the edit and F1 scores. Nevertheless, BNG-IRL still outperforms all baselines in average segment-level evaluation metrics.

Due to the similarity in the force domain, BNGMM groups the samples before and after pushing the box in one cluster, see red samples in the center column of Fig.~\ref{tab:comparison_box_pushing}. The turning point is also not identified with this approach, since without considering the intent of the actions, the black samples appear close in feature space. However, this method reliably detects the approach and the push phase, which leads to the second-highest quantitative evaluation results after BNG-IRL reported in Table~\ref{tab:segmentation_evaluation_box_pushing}.

\begin{table}[thpb]
\centering
\caption{Quantitative evaluation of our BNG-IRL segmentation approach against the four baselines for the box-pushing task. The accuracy measures the performance at the sample level, while the edit and F1 scores assess the performance at the segment level.}
\label{tab:segmentation_evaluation_box_pushing}
\begin{tabular}{p{2.13cm}rrrr}
\toprule
Method & Acc & Edit & F1@\{10, 25, 50\} & Avg\\
\midrule
BNG-IRL\color{gray}{(our)} & \cellcolor{green!20}89.3 & 66.7 & \cellcolor{green!20}80.0 / 80.0 / 80.0 & \cellcolor{green!20}79.2\\
BNGMM & 79.5 & 57.1 & 100 / 75.0 / 75.0 & 77.3\\
BN-IRL & 41.8 & 4.7 & 85.7 / 57.1 / 0.00 & 37.9\\
BOCPD & 77.0 & \cellcolor{green!20}80.0 & 66.7 / 66.7 / 66.7 & 71.4\\
AWE & 73.0 & 50.0 & 75.0 / 75.0 / 75.0 & 69.6\\
\bottomrule
\end{tabular}
\end{table}
BN-IRL, on the other hand, which solely considers the subgoal's influence on the actions when subdividing the demonstration has difficulties inferring coherent skills in this setup, leading to both the lowest frame-wise and segment-wise performance scores. The majority of observations (blue samples in the right column of Fig.~\ref{tab:comparison_box_pushing}) are assigned to the same skill, whose subgoal is the last state of the demonstration. With the proposed observation likelihood in \cite{michini2015bayesian} $p(\bm{o}_i\vert z, \bm{g}_{k}) = \frac{e^{\alpha Q^*(\bm{s}_i, \bm{a}_i, \bm{g}_k)}}{\sum_a e^{Q^*(\bm{s}_i, a, \bm{g}_k)}}$, that only considers the observed action $\bm{a}_i$ with respect to the optimal policy to reach $\bm{g}_k$, the actions of the blue samples indeed appear to be directed towards reaching the blue subgoal. As mentioned in Sec.~\ref{sec:subgoal-driven-intention}, our proposed optimality criterion for action $\bm{a}_i$ and subgoal $\bm{g}_k$ also considers the actions after $\bm{a}_i$ and can therefore exclude the possibility that the actions before reaching the turning point are targeted towards the blue subgoal. Since we aim to find a sequence of separable skills, our observation likelihood is advantageous in this case over the one proposed in BN-IRL. Since AWE does not consider the force signals, the box-pushing phase can not be correctly distinguished from the phases 1) and 3). BOCPD correctly detects the start and end of the pushing phase but suffers from over-segmentation in the end of phase 2 and fails to detect the turning point.

\subsection{Manipulation Task with DLR LWR IV}
\label{sec:Manipulation_Real_World}
\subsubsection{Experimental Setup}
\label{sec:Ex_RW_Setup}
\begin{figure}[thpb]
  \centering
  \captionsetup{belowskip=-5pt}
  \includegraphics{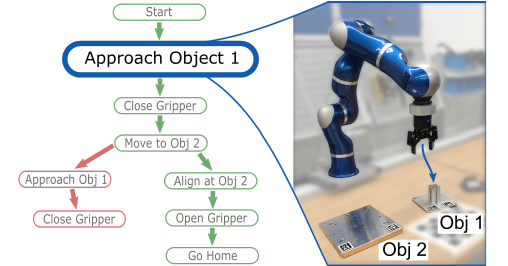}
  \caption{The task graph structures the low-level skills on a higher level of abstraction. The green sequence of skills represents the intended task flow from the initial demonstration. The recovery behavior in red restores a situation from which the robot can continue with the intended task flow. The monitored execution of the 'Approach Object 1' skill is shown on the right.}
  \label{fig:Setup_Manipulation}
\end{figure}
As seen in Fig.~\ref{fig:Setup_Manipulation}, a DLR LWR IV robot \citep{albu2007dlr}, equipped with a 2-finger gripper and a force-torque (FT) sensor, is mounted on a linear axis. The FT sensor measures the forces and torques acting on the end-effector. An Intel Realsense camera tracks the 6D poses of objects. Two pedals near the workspace allow the user to start and stop the demonstration and operate the gripper. When the demonstration mode is activated, the robot compensates for its own weight, enabling kinesthetic teaching. The task for the robot is to pick up object 1 and to place it on top of object 2 such that their edges are aligned. Both objects can have arbitrary initial 6D poses. Additionally, the robot should detect and recover from a task anomaly, where the robot loses the object during transport. A video of the experiment setup, as well as the task learning procedure and autonomous error recovery, are provided in Extension~2. The feature space $\bm{\mathcal{F}}$ for this task includes the following seven features: the Euclidean distance between the robot's end-effector and objects 1 and 2, the relative orientation between the end-effector and objects 1 and 2, the force acting on the end-effector, the gripper finger distance, and the gripper's grasp status $\in \{-1, 0, 1\}$.

\subsubsection{Incremental Task Learning Procedure}
\label{sec:Exp_Task_Learning}
\begin{figure}[thpb]
\centering
\includegraphics[clip, trim=30 16 20 11, scale=1.1]{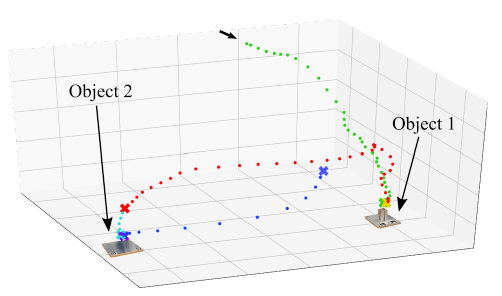}
\caption{The Cartesian end-effector trajectory during the user demonstration of the manipulation task. Samples with the same color are assigned to the same skill. The state subgoals of all six skills are depicted with an X. Several subgoals are inferred in the vicinity of the objects, as objects are grasped or released here. The features between the skills therefore only differ in the gripper finger distance and grasp status.}
\label{fig:Traj_Seq_2}
\end{figure}

The teaching procedure starts with an initial demonstration of the task, where the user picks up object 1 with the robot and places it on top of object 2 in the desired goal configuration. The segmentation result can be seen in Fig.~\ref{fig:Traj_Seq_2}, where the important subgoals of the task are correctly identified. The inferred sequence of skills is then used to construct an initial task graph (see the green sequence in Fig.~\ref{fig:Setup_Manipulation} and Extension~2), where the skills are encoded as DMPs.

\begin{figure}[b]
\centering
\includegraphics[clip, trim=0 0 0 0, scale=1]{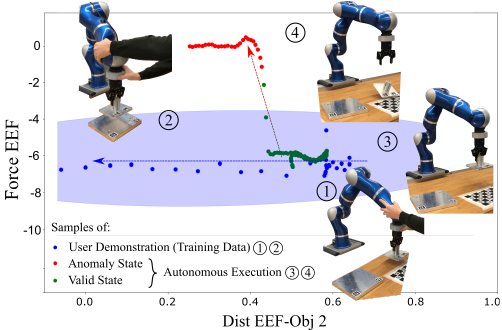}
\caption{Anomaly detection mechanism illustrated with the expected force during the transportation skill. In blue are the samples collected during the user demonstration \ciwn{1} - \ciwn{2} which were used to infer the skill's expected feature region, represented by a 2D Gaussian (light blue area). The green and red samples are recorded during the robot execution \ciwn{3}, \ciwn{4}. First, the features lie within the expected region (green samples) but as soon as the object falls out of the gripper and the force on the end-effector suddenly decreases, the samples are classified as anomolous (red samples) which eventually triggers an anomaly.}
\label{fig:Anomaly_Dist}
\end{figure}

The system then switches to the autonomous execution phase, in which the robot performs the skills from the initial task graph. Using the skills' DMPs, the robot's EEF trajectories can be generalized to varying subgoal configurations. In case a task anomaly occurs, like losing the object during manipulation, the anomaly detection component automatically registers a deviation of the features from their expected region. The anomaly detection mechanism is visualized for the measured end-effector force in Fig.~\ref{fig:Anomaly_Dist}. It can be seen that the measured samples before the task anomaly (green) were confidently classified with the two-step anomaly detection approach as belonging to the executed skill. As soon as the measured force leaves the expected constraint region, the Mahalanobis distance exceeds the anomaly threshold (\ref{eq:P_A}). After 300~ms, an anomaly is confidently detected and the robot stops. Since no recovery behavior is available in the task graph, the robot requests a user demonstration to resolve the anomaly.

\begin{figure*}[th]
  \centering
   \includegraphics[scale=0.965]{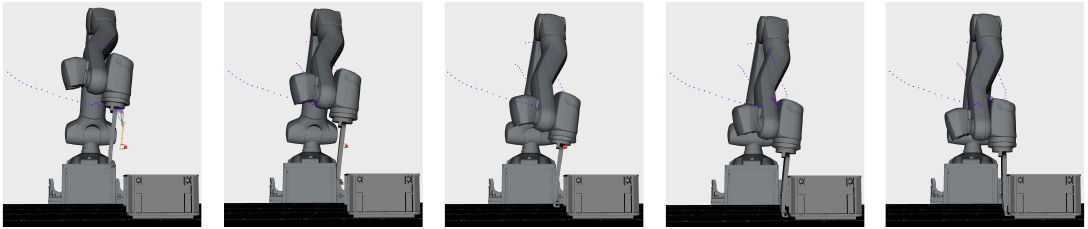}
  \caption{The mean EEF poses $\bm{\mu}_{G,k}$ of the skills' subgoals in the sequence of occurrence during the task. The subgoals are defined as regions of 6D EEF poses. If the robot reaches the subgoal region of a skill, the next intended task flow skill with a new subgoal is scheduled.}
  \label{fig:Box_grasping_description_SGs}
\end{figure*}

The demonstrated recovery behavior is segmented, encoded analog to the initial demonstration, and appended to the skill in the task graph during which the anomaly was detected (see sequence connected with red arrows in Fig.~\ref{fig:Setup_Manipulation}). The robot can now leverage this recovery behavior to automatically resolve similar situations occurring at any phase of the transportation skill.

\subsection{Box Grasping and Locking with DLR SARA}
\label{sec:box_grasping}
This task stands for a variety of specialized contact tasks that require precise coordination of end-effector pose and applied force in the different phases. The phases that make up this task cannot be described by common robot skills, but require precise coordination of end-effector pose and applied force and therefore must be learned through demonstration on the real system. We utilize this task to demonstrate the different capabilities of our proposed framework, which include A) the three variants of teaching sequences, B) the hierarchical task decomposition based on a few user demonstrations, and C) the autonomous task execution, where new anomalies can be detected and autonomously resolved after a recovery behavior has been learned. A video showcasing these capabilities is provided in Extension~3.
\subsubsection{Experimental Setup}
\label{sec:setup_box_grasping}

The DLR SARA robot \citep{iskandar2020joint}, is equipped with a 6-DOF FT sensor in the wrist that can measure task forces and torques during kinesthetic demonstration which need to be reproduced during the execution \citep{iskandar2021collision}. On the last robot link are buttons and a display that the user can interact with during the demonstrations. The display provides the user with information about the robot's status and informs which buttons to press to navigate the desired teaching sequence. As shown in Fig.~\ref{fig:Teaser_Box_grasping} and Fig.~\ref{fig:Gripper_Design}, a passive gripper is mounted on the robot which is designed to pick and lock the standardized euro boxes using a sequence of specific motions with the robot. A euro box with a known pose relative to the robot's base coordinate frame is located in the robot's workspace.
\subsubsection{Recorded Data}
\label{sec:Box_Grasping_Data}
The state vector $\bm{s} \in \bm{S}$ for this task is defined by the 6D robot end-effector pose in the box coordinate frame $\bm{s} = [\bm{p}_{\mathrm{EEF}}, \bm{q}_{\mathrm{EEF}}]$ represented by the Cartesian position $\bm{p}_{\mathrm{EEF}} = [x, y, z]$ and orientation in unit quaternions $\bm{q}_{\mathrm{EEF}} = [q_w, q_x, q_y, q_z]$. The considered features $\bm{\mathcal{F}}$ are the measured external forces $\bm{F}_{\mathrm{ext}} = [ f_x, f_y, f_z]$, the torques at the wrist $\bm{T}_{\mathrm{wrist}} = [t_x, t_y, t_z]$, the distance  $\bm{D}_{\mathrm{EEF}} = [d_x, d_y, d_z]$ and the orientation in Euler angles $\bm{O}_{\mathrm{EEF}} = [\alpha, \beta, \gamma ]_{XYZ}$ of the EEF relative to the box to pick. To preprocess the recorded data for learning, the features are normalized with element-wise mean normalization $f_{\mathrm{norm}} = \frac{f - \mu_f}{max(f) - min(f)}$. Using Riemannian geometry, unit quaternions $\bm{q} \in \bm{\mathcal{S}}^3$ can be mapped into a tangent space that locally linearizes the manifold $\bm{\mathcal{S}}^3$. We follow the proposed formulation of \cite{simo20173d} and \cite{calinon2020gaussians} that leverages Riemannian geometry to extend GMMs to non-Euclidean data. That allows us to encode the 6D end-effector pose and velocity as well as the contact force in one GMM for learning motion primitives and constraints as described in Sec.~\ref{sec:motion-generation-anomaly-detection}.

\subsubsection{Task Description}
\label{sec:box_gripping_task_description}
To better understand how the grasping and locking mechanism of the box gripper works, which has to be learned from demonstration, we first describe the gripper hardware and functionality before explaining the sequence of motions and difficulties during the task.
\paragraph{Gripper Design}
\label{sec:gripper}
\begin{figure}[thpb]
  \centering
  \includegraphics[scale=0.965]{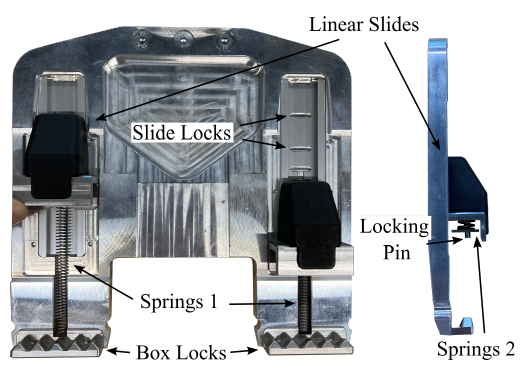}
  \caption{Passive box gripper design with movable linear slides to grasp and lock a euro box.}
  \label{fig:Gripper_Design}
\end{figure}
The gripper \citep{eiberger2022kistengreifer}, as depicted in Fig.~\ref{fig:Gripper_Design}, has two movable linear slides. When moving the linear slides up, springs 1 are tensioned and pull the slides back into the neutral configuration if the slides are released. When the locking pin in the linear slides is pushed, a latch at the back of the slides retracts and engages in one of the slide locks. This blocks the linear movement of the slides, which enables the grasping of boxes of different heights. To fixate a box inside the gripper, springs 2 are compressed to maintain a constant force between the linear slides and the box locks. The box locks engage at its counterpart at the bottom of the box and prevent movement perpendicular to the linear motion of the slides.
\paragraph{Kinematic Sequence to Grasp and Lock the Box}
\label{sec:kinematic_sequence}
The task, as depicted in Fig.~\ref{fig:Teaser_Box_grasping} and Fig.~\ref{fig:Box_grasping_description_SGs}, consists of five different phases. In the first phase, the robot moves from a start configuration closer to the box, while slightly tilting the end-effector to avoid contact between the box locks of the gripper and the box. In the second phase, the gripper moves closer to the box until both linear slides evenly contact the side wall of the box (see column 2 of Fig.~\ref{fig:Teaser_Box_grasping}). After that, while maintaining contact between the slides and the box, the gripper is pushed down along the side wall of the box, tensioning springs 1 until the third configuration in the upper row of Fig.~\ref{fig:Teaser_Box_grasping} is reached. In the next phase, the box gripper is moved closer to the box, such that the locking pins in the slides are pushed and the linear movement of the slides is blocked. In the final phase, springs 2 are compressed, while rotating the box gripper into a vertical configuration, such that the box locks can engage at the bottom of the box.
\begin{figure}[thpb]
  \centering
  \begin{tabular}{cc}
      Constraint Regions $\mathcal{N}(\bm{\theta}_{C,k})$ & Subgoal Regions $\mathcal{N}(\bm{\theta}_{G,k})$
      \vspace{0.2cm}\\
  \end{tabular}  
  \includegraphics[clip, trim=4 0 0 0, scale=1]{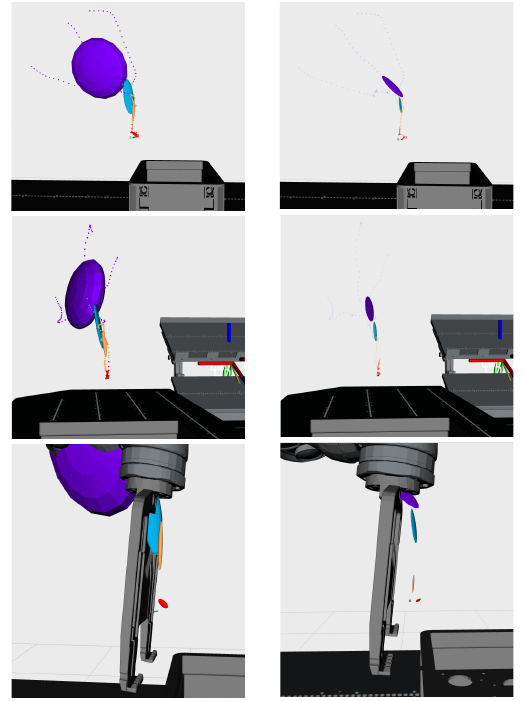}
  \caption{Position part of the subgoal and constraint regions inferred with BNG-IRL. The constraint regions $\bm{C}_k$ and the subgoal regions $\bm{G}_k$ are both defined as multivariate Gaussian distributions with parameters $\bm{\theta}_{C,k}$ and $\bm{\theta}_{G,k}$, respectively. However, the constraints are defined in feature space $\bm{\mathcal{F}}$ and need to be met during the execution of a skill, while the subgoal regions are defined in state space $\bm{S}$ and are postconditions that must be reached for successful skill termination. The ellipsoids represent the observed correlation and variation of the training data of each skill. An EEF pose within the constraint region of skill 2 is depicted in the lower left image, while the lower right image shows an EEF pose in the subgoal region of skill 1. The corresponding mean EEF poses $\bm{\mu}_{G,k}$ of all subgoal regions are depicted in Fig.~\ref{fig:Box_grasping_description_SGs}. The characteristic features that define the constraints of each skill can be seen in Fig.~\ref{fig:feats_box_grasping}.}
  \label{fig:SG_Constraints_Box_Grasping}
\end{figure}
\paragraph{Difficulties}
\label{sec:box_gripping_difficulties}
The task poses several difficulties that can prevent successful grasping and locking of the box. The challenges of every phase are depicted in the bottom row of Fig.~\ref{fig:Teaser_Box_grasping}. In the first phase, the box locks of the gripper must not collide with the box, while the slides are already located over the wall of the box so that in phase 2, only the front part of the slides contact the wall of the box. If the gripper moves too close to the box in phases 2 or 3, the locking pin is pushed too early and the sliding mechanism is blocked. If the gripper moves too far away from the box and the slides lose contact with the wall, the springs 1 pull the slides into the neutral configuration. In both cases, configuration 3 can not be reached. If configuration 3 has been reached, the gripper can move closer to the box, such that the locking pins are pushed and the springs 2 can be compressed. Before rotating into the vertical configuration, the robot must push the gripper further down to avoid a collision between the box locks and the lower part of the box. Precise coordination of applied force in the direction of spring 2 and rotation of the gripper is required during this phase since there is only a very small clearing between the gripper and the box. If the robot does not exert enough force or rotates too early, the box locks collide with the box and the final configuration can not be reached.
\subsubsection{Initial Teaching Sequence}
\label{sec:box_grasping_initial_teaching}
\begin{figure}[thpb]
  \centering  
  \begin{tabular}{cccc}
    &  demo 1 &  demo 2 &  demo 3\\
    \hspace{-15pt}
    \rotatebox{90}{\quad{pos z (m)}} &
    \hspace{-15pt}
    \includegraphics[clip, trim=10 5 40 15,scale=0.19]{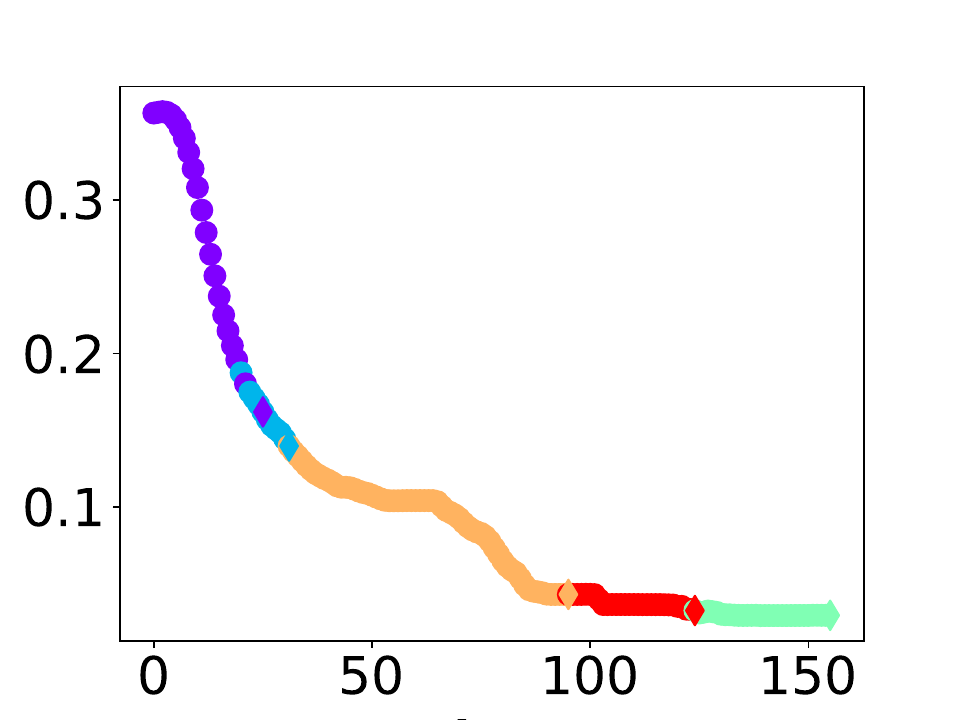} & 
    \hspace{-15pt}
    \includegraphics[clip, trim=10 5 40 15,scale=0.19]{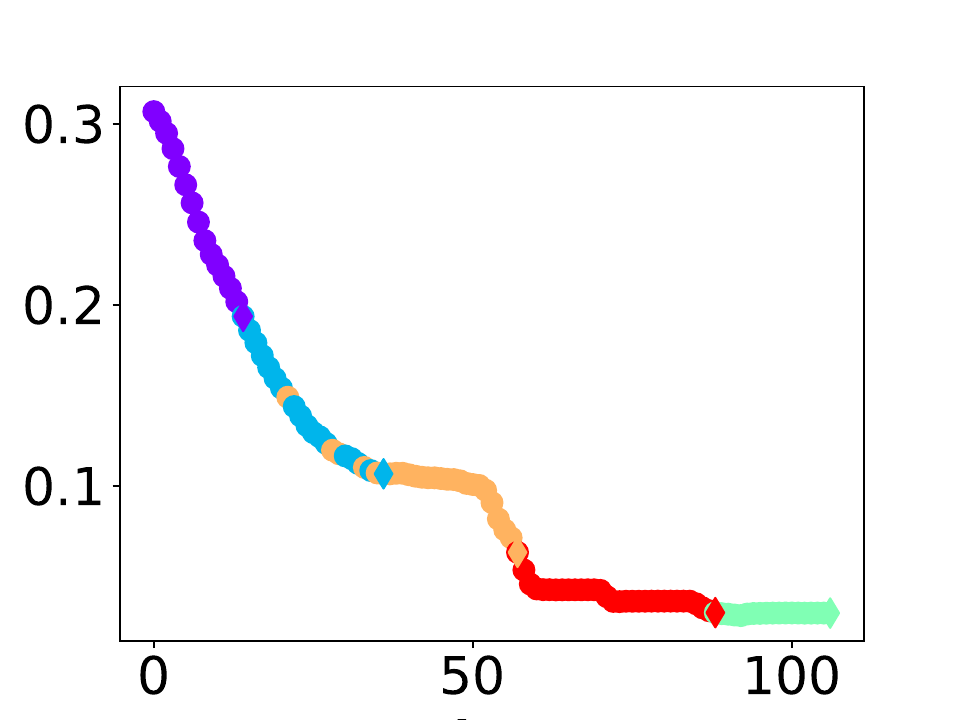} & 
    \hspace{-15pt}
    \includegraphics[clip, trim=10 5 40 15,scale=0.19]{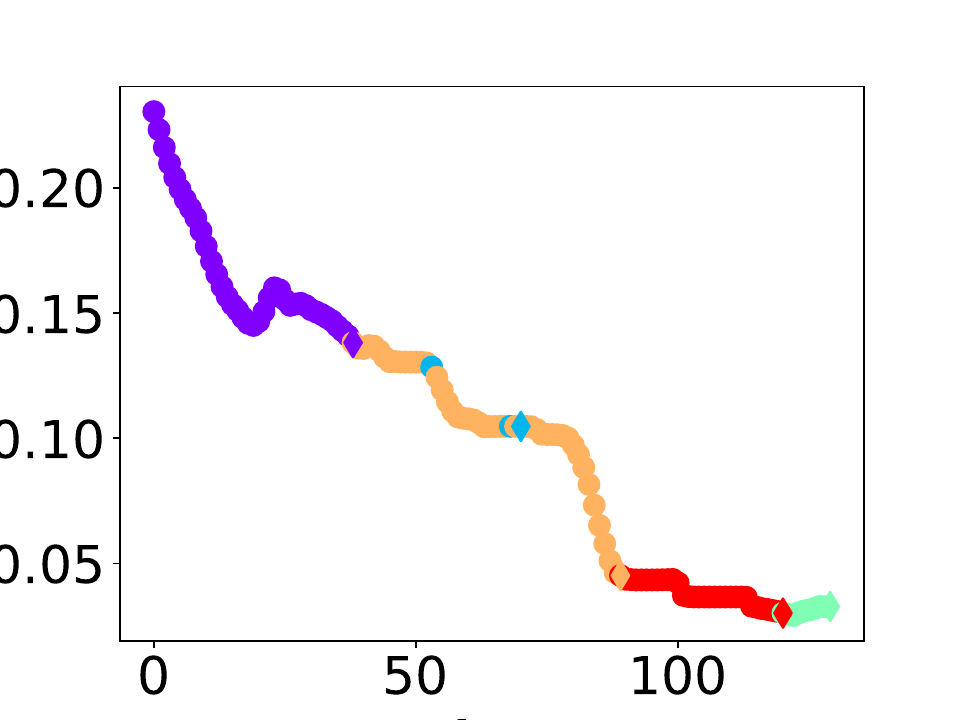}\\    
    \hspace{-15pt}
    \rotatebox{90}{\quad{force z (N)}} &
    \hspace{-15pt}
    \includegraphics[clip, trim=10 5 40 15,scale=0.19]{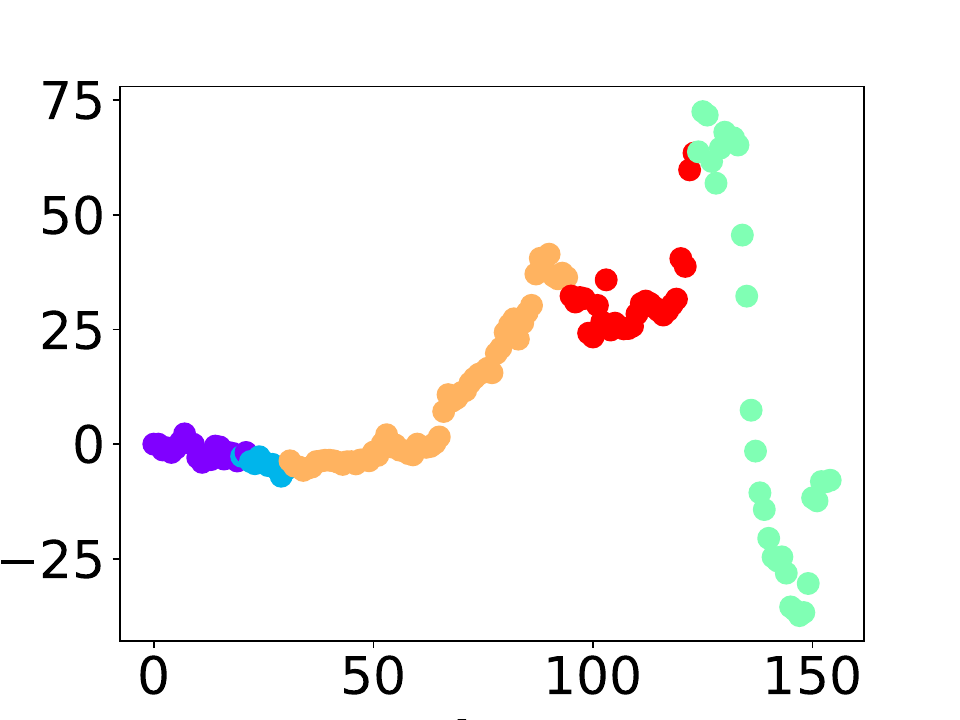} & 
    \hspace{-15pt}
    \includegraphics[clip, trim=10 5 40 15,scale=0.19]{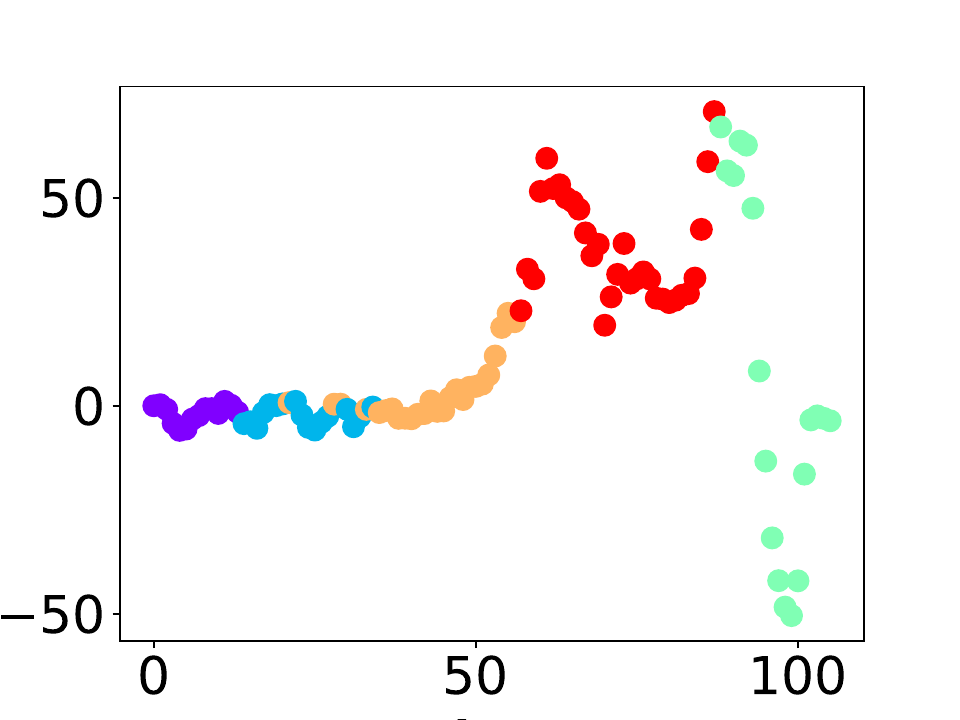} & 
    \hspace{-15pt}
    \includegraphics[clip, trim=10 5 40 15,scale=0.19]{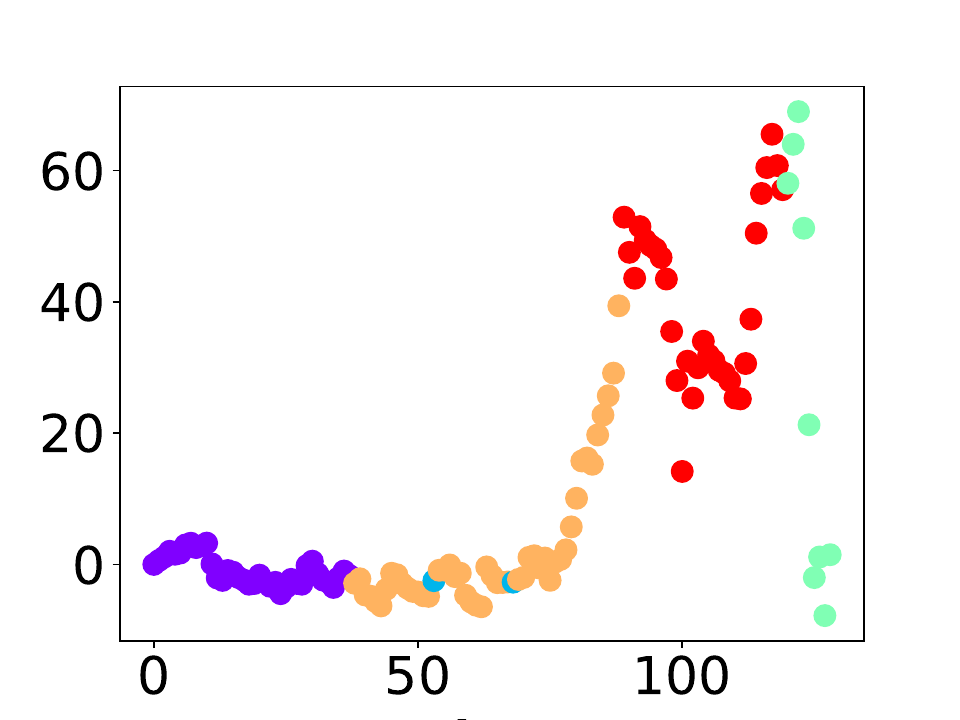} \\
    \hspace{-15pt}
    \rotatebox{90}{\quad{ang y (rad)}} &
    \hspace{-15pt}
    \includegraphics[clip, trim=10 5 40 15,scale=0.19]{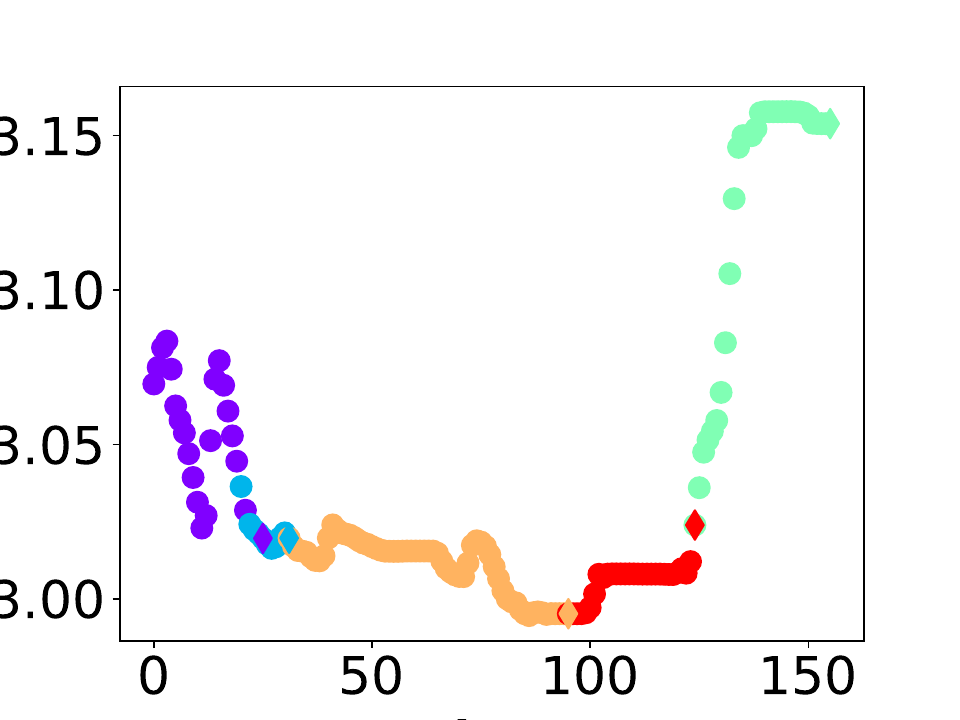} & 
    \hspace{-15pt}
    \includegraphics[clip, trim=10 5 40 15,scale=0.19]{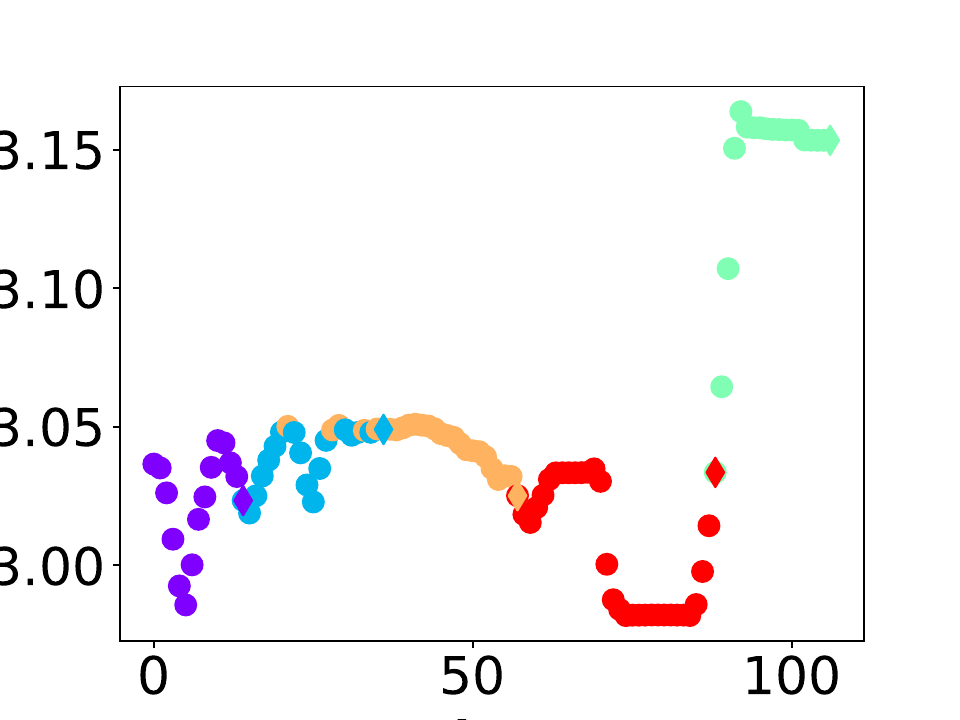} & 
    \hspace{-15pt}
    \includegraphics[clip, trim=10 5 40 15,scale=0.19]{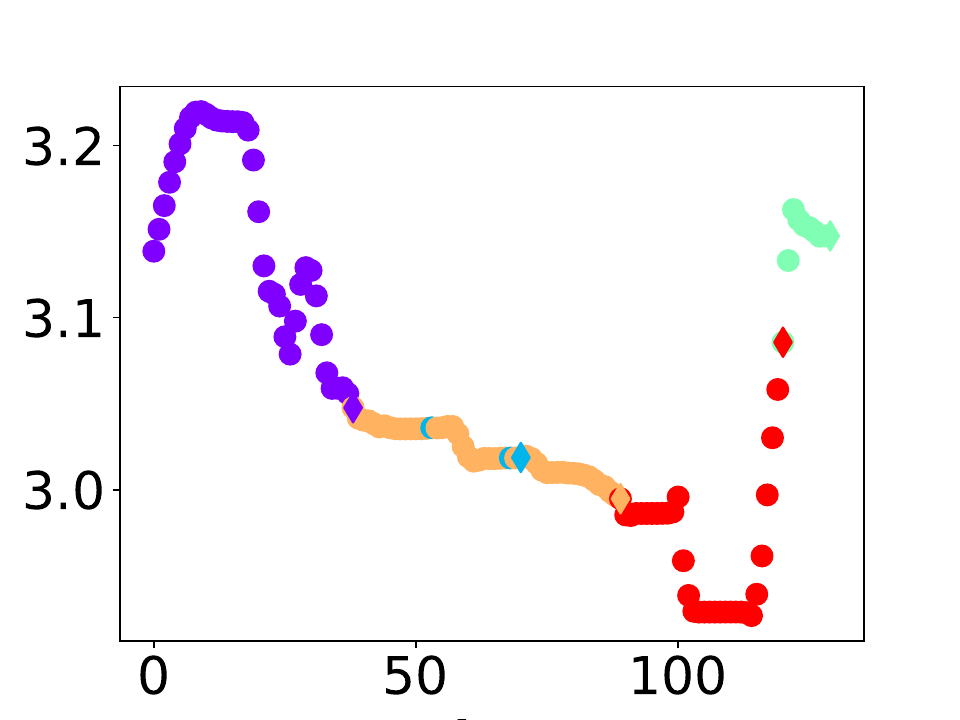} \\
    \hspace{-15pt}
    \rotatebox{90}{\quad{pos x (m)}} &
    \hspace{-15pt}
    \includegraphics[clip, trim=10 5 40 15,scale=0.19]{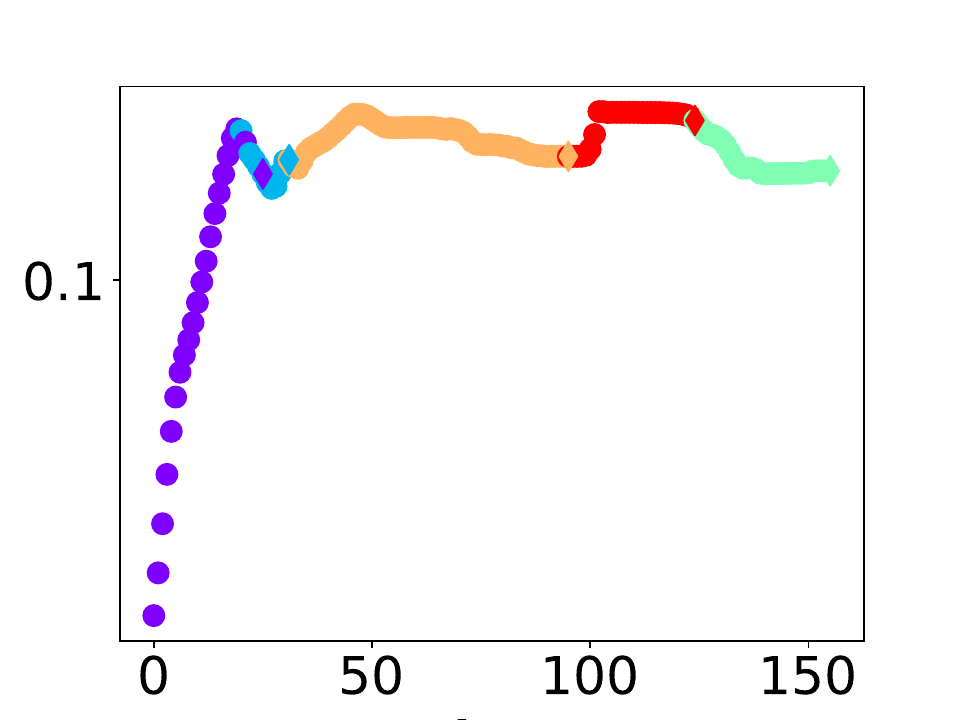} & 
    \hspace{-15pt}
    \includegraphics[clip, trim=10 5 40 15,scale=0.19]{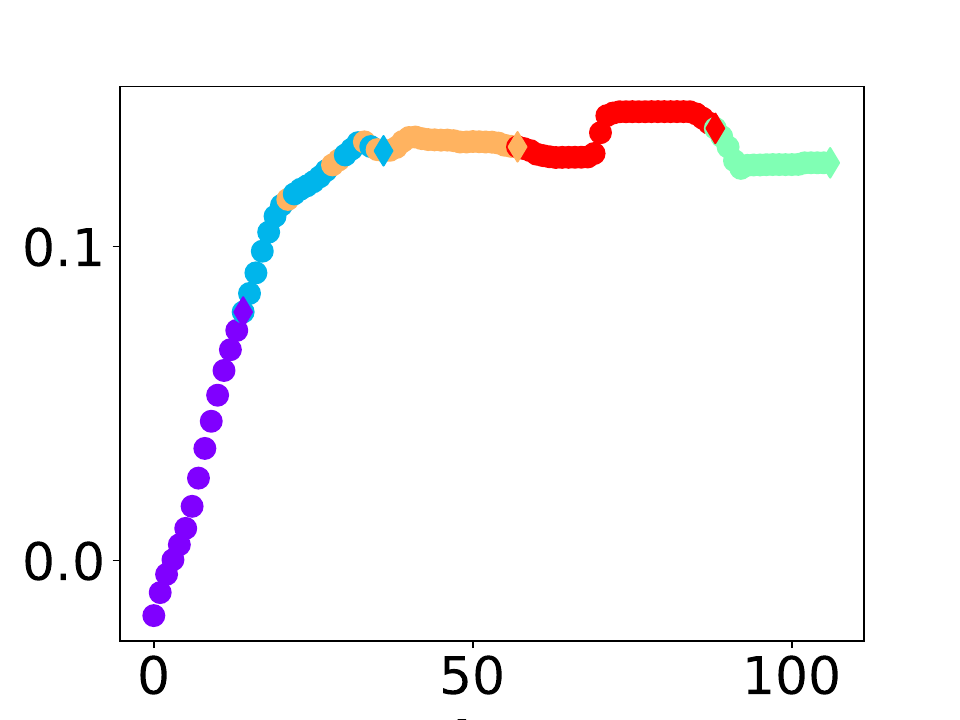} & 
    \hspace{-15pt}
    \includegraphics[clip, trim=10 5 40 15,scale=0.19]{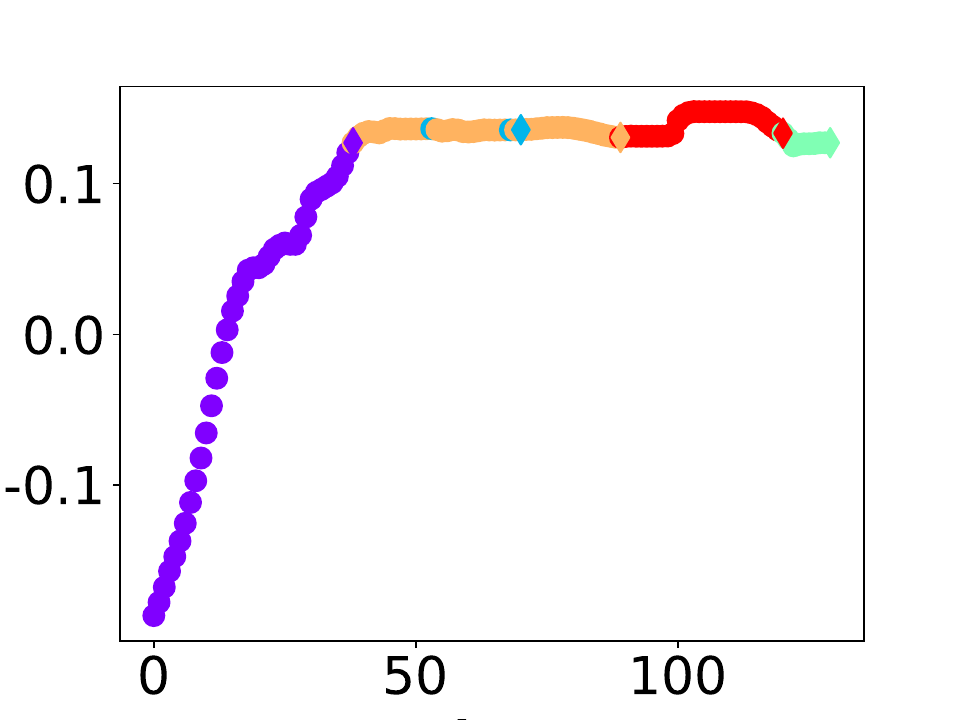} \\
    \hspace{-15pt}
    \rotatebox{90}{\quad{force x (N)}} &
    \hspace{-15pt}
    \includegraphics[clip, trim=10 5 40 15,scale=0.19]{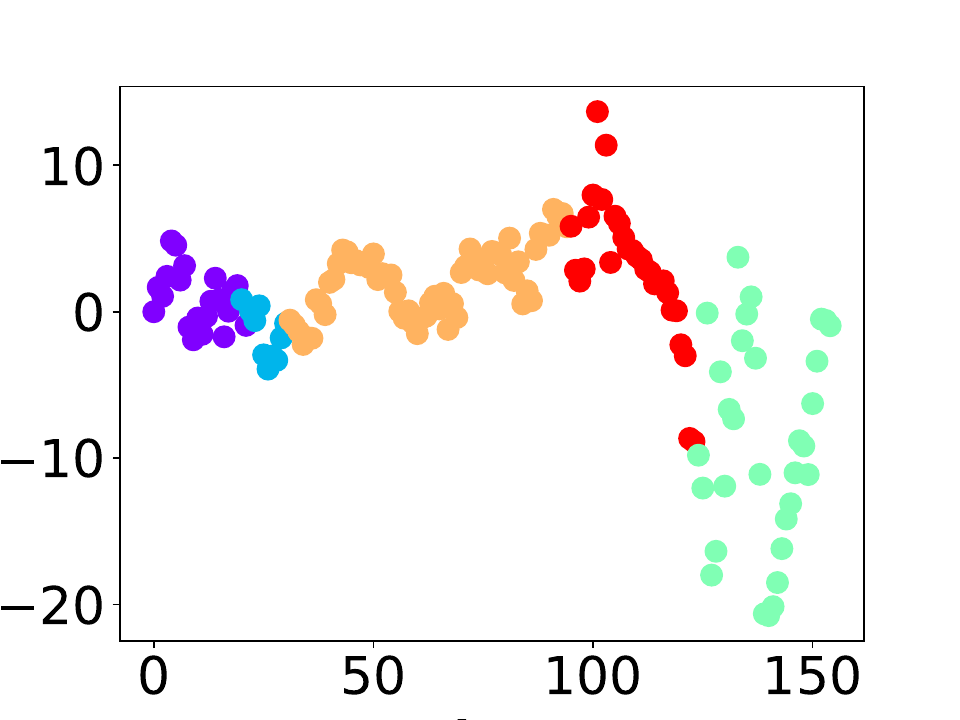} & 
    \hspace{-15pt}
    \includegraphics[clip, trim=10 5 40 15,scale=0.19]{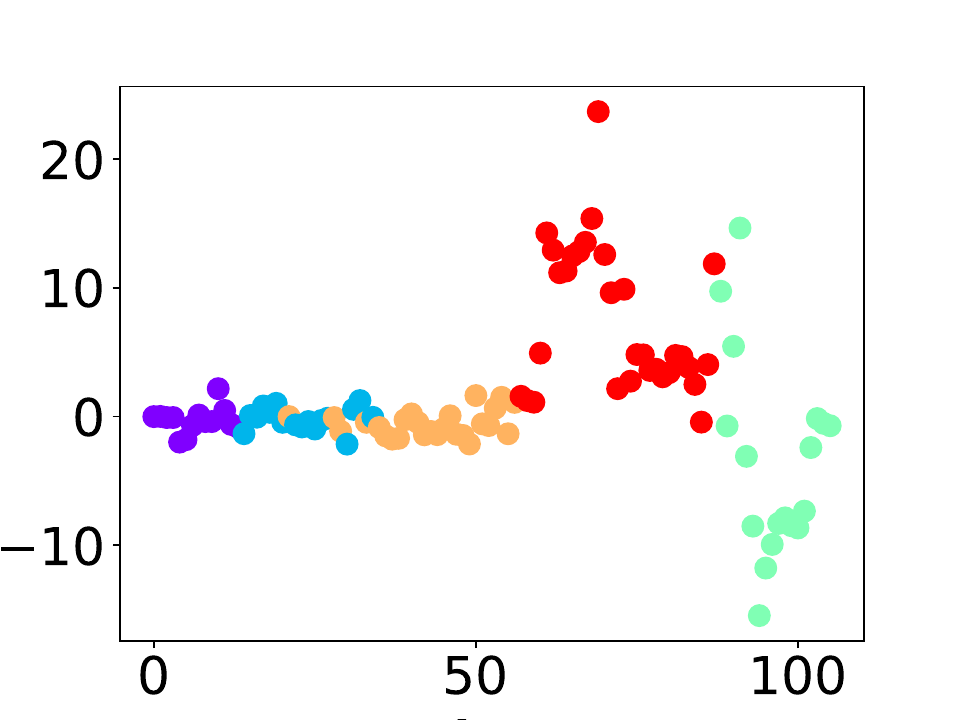} & 
    \hspace{-15pt}
    \includegraphics[clip, trim=10 5 40 15,scale=0.19]{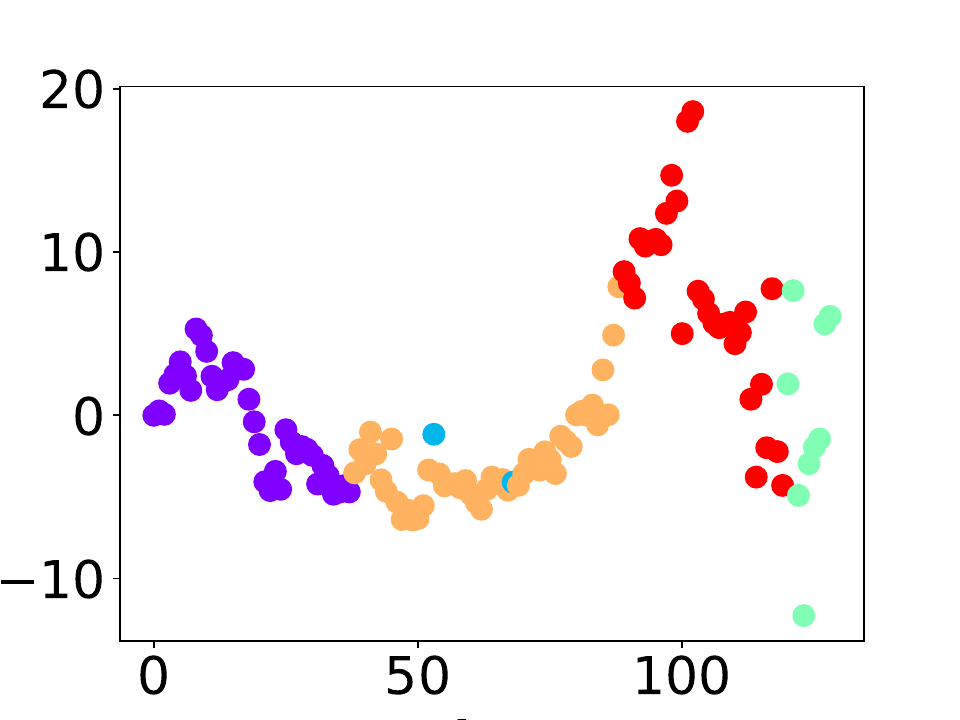} \\
  \end{tabular}
  \captionsetup{belowskip=-10pt}
  \caption{Characteristic features during the initial task demonstrations, which are utilized to segment the task into skills based on consistent correlations among the demonstrations. The features define the constraint regions $\bm{C}_k$ of each skill. The colors represent the different skills.}
  \label{fig:feats_box_grasping}
\end{figure}

To learn an initial model of the task of grasping and locking the box with the gripper, a user provides three demonstrations using kinesthetic teaching. These demonstrations are then segmented with our BNG-IRL approach. The results of the sampling procedure with the highest MAP likelihood are shown in Fig.~\ref{fig:SG_Constraints_Box_Grasping} and Fig.~\ref{fig:feats_box_grasping}. The results indicate a segmentation of the task into the five skills, which were described above. As shown in Fig.~\ref{fig:SG_Constraints_Box_Grasping}, the further the task progresses, the more restricted the subgoal (right) and constraint regions (left) become, because the process requires more precision once contact between the robot and the box is established. As expected, the subgoal regions are located at the end of each skill. The skills' actions are targeted toward reaching its subgoal configuration. In the upper two rows of Fig.~\ref{fig:SG_Constraints_Box_Grasping}, only the 3D position component of the subgoal region is depicted, however, the subgoal regions are defined in state space, which also includes a 3D orientation component. To illustrate this, Fig.~\ref{fig:Box_grasping_description_SGs} depicts the mean 6D EEF pose $\bm{\mu}_{G,k}$ of every subgoal region. The first row of Fig.~\ref{fig:Teaser_Box_grasping} shows the EEF poses during execution of the task that fulfill condition (\ref{eq:subgoal_reached_condition}), which trigger a transition to the next intended task flow skill in the task graph. Fig.~\ref{fig:feats_box_grasping} shows that the similarities and correlations among the features are considered in the segmentation process as well. Skill 3 (orange) encodes the linear correlation between force and position in z-direction, caused by springs 1. As seen in the first two rows of Fig.~\ref{fig:feats_box_grasping}, a deflection in z from the neutral configuration at 0.1~m causes a linear increase of force in z-direction. In the last skill (light green), a significant reduction of force in the z-direction is correlated with a rotation around the y-axis and a motion in the negative x-direction. This motion causes the box locks to engage, which compensates for the force of the compressed springs 2.

\paragraph{Quantitative Evaluation of Unsupervised Segmentation}

\label{sec:box_grasping_segmentation_evaluation}
We evaluate the task segmentation performance of BNG-IRL against the two baseline approaches AWE \citep{shi2023waypoint} and BOCPD \cite{sugawara2023unsupervised} described in Sec.~\ref{sec:comparison_box_pushing}. We compute accuracy, edit and F1@{10, 25, 50} scores, as explained in Sec.~\ref{sec:segmentation_metrics}.

\begin{figure}
    \centering
    \includegraphics[clip, trim=100 15 100 80, width=\linewidth]{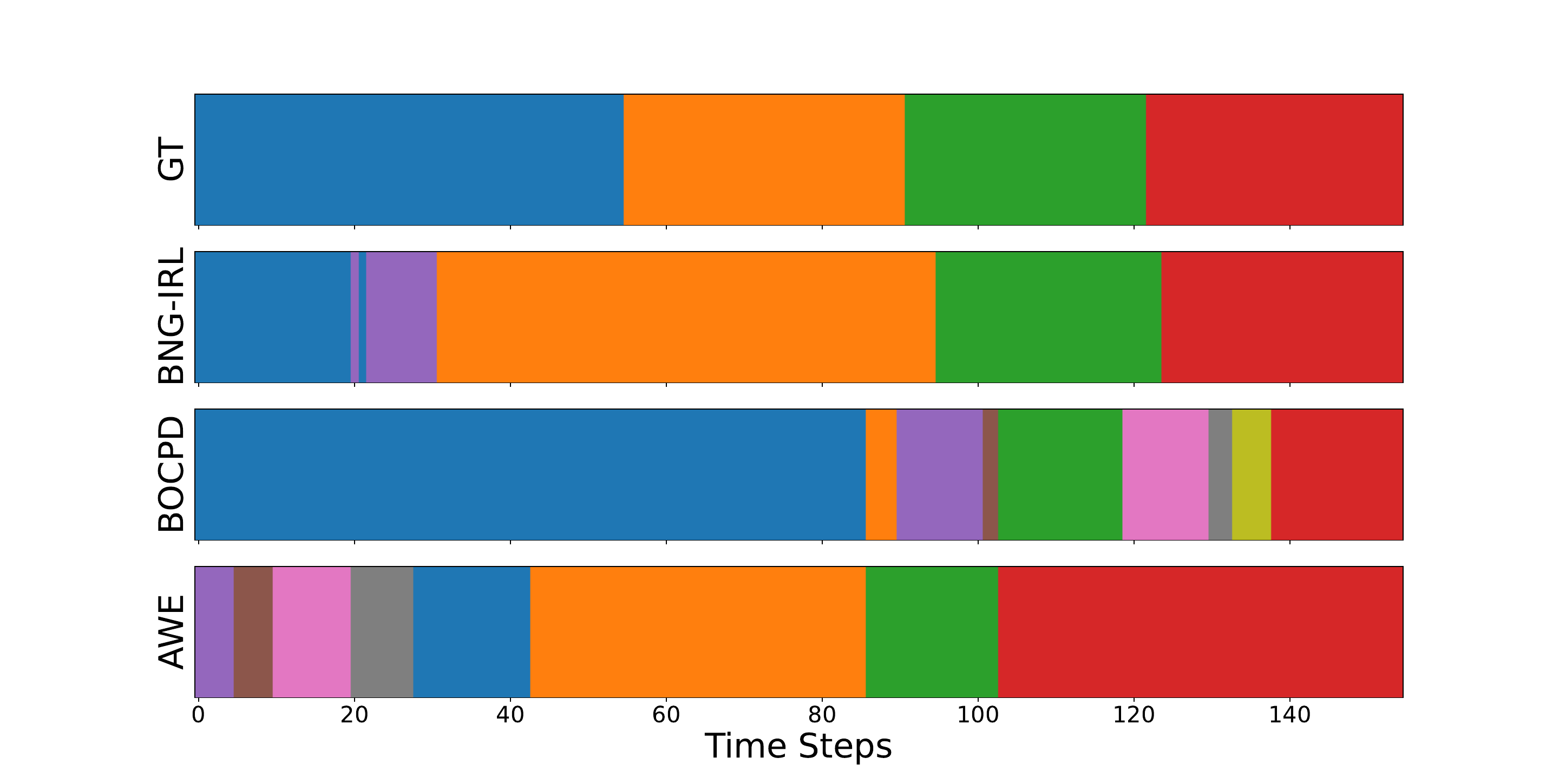}
    \caption{Comparison of the ground truth segmentation of demonstration 1 with the task segmentation results of BNG-IRL, BOCPD, and AWE. BOCPD suffers from over segmentation in the final phase of the task, while AWE detects too many segments in the beginning.}
    \label{fig:boc_grasping_segmentation_ts}
\end{figure}

\paragraph{Results}
Table~\ref{tab:segmentation_evaluation_box_grasping} shows that BNG-IRL outperforms both baselines in frame- and segment-level metrics averaged over three task demonstrations. As illustrated in Fig.~\ref{fig:boc_grasping_segmentation_ts}, AWE’s chosen error threshold of 10 mm, necessary for accuracy during the contact phase, causes over-segmentation at the beginning of the task. Because the robot does not approach the box along a linear path, the trajectory is unnecessarily subdivided, although this is not relevant for the task. In contrast, BNG-IRL does not segment each task demonstration individually but finds a combined segmentation result for all demonstrations. It leverages the similarities across different demonstrations, which leads to a more consistent result. BOCPD, however, suffers from over-segmentation during the contact phase while failing to detect the transition from free space to contact. This occurs because the contact force only increases slowly during the first contact skill but exhibits larger fluctuations in later stages due to higher impact forces. Consequently, applying BOCPD to the time derivative of force and torque measurements leads to this segmentation result.

\begin{table}[thpb]
\centering
\caption{Quantitative evaluation of our BNG-IRL segmentation approach and the two baselines BOCPD \citep{sugawara2023unsupervised} and AWE \citep{shi2023waypoint} for the box-grasping task. The accuracy measures the performance at the sample level, while the edit and F1 scores assess the performance at the segment level.}
\label{tab:segmentation_evaluation_box_grasping}
\begin{tabular}{p{2.13cm}rrrr}
\toprule
Method & Acc & Edit & F1@\{10, 25, 50\} & Avg\\
\midrule
BNG-IRL\color{gray}{(our)} & \cellcolor{green!20}74.9 & \cellcolor{green!20}67.9 & \cellcolor{green!20}88.9 / 88.9 / 74.1 & \cellcolor{green!20}78.9\\
BOCPD & 61.5 & 46.8 & 58.8 / 44.1 / 34.5 & 49.1\\
AWE & 55.6 & 44.8 & 64.2 / 64.2 / 32.1 & 52.2\\
\bottomrule
\end{tabular}
\end{table}

\subsubsection{Skill Refinement Teaching Sequence}
\label{sec:refinement_box_grasping}
\begin{figure}[thpb]
  \centering
  \begin{tabular}{cc}
      \hspace{0.2cm} Skill 4 Model & \hspace{1.8cm} Forces in z
      \\
  \end{tabular}  
   \includegraphics[scale=0.965]{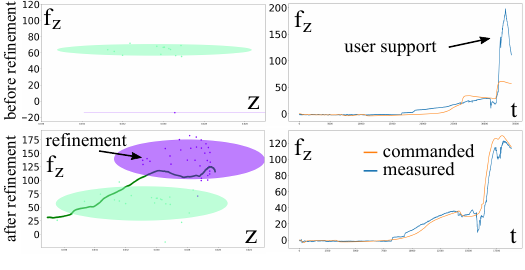}
  \caption{Force Refinement of skill 4. The relation between z-position and commanded force in z-direction for skill 4, is illustrated by the 2D excerpt of the GMM for motion- and force generation (left). The measured and commanded forces in z-direction during the entire task are depicted on the right. As shown in the upper row, the low-level skill model from the initial user demonstrations generates a force command in z-direction, which is not enough to fully compress springs 2. After collecting new user support training data, the low-level skill model is updated, which results in a higher commanded force during skill 4 for the same EEF configuration.}
  \label{fig:Box_grasping_Force_Refinement}
\end{figure}

The inferred skills from the previous teaching sequence are encoded as dynamical systems for motion generation and anomaly detection as described in Sec.~\ref{sec:motion-generation-anomaly-detection}, where we set the number of mixture components per skill equal to two. The robot is already capable of autonomously executing the learned skills, however, incremental refinements in the force domain might still be required to complete the task successfully. We additionally utilize the refinement sequence to increase the variation in the training data for each skill's low-level model in a combined robot execution and user support phase. Since the autonomous execution does not simply replicate the demonstration, the contact forces during autonomous execution typically differ slightly from the ones recorded during the user demonstrations. As seen in the upper right graph of Fig.~\ref{fig:Box_grasping_Force_Refinement}, the commanded force in the last part of the skill 4 is not enough to fully compress springs 2 in the commanded EEF configuration. That is why a user supports during this phase by applying additional force in z-direction so that the task can be successfully completed. As seen in the lower row of Fig.~\ref{fig:Box_grasping_Force_Refinement}, the recorded force is used to update the low-level model, which results in a higher commanded force for that EEF configuration in the next autonomous execution. 

\begin{figure}[thpb]
  \centering
  \begin{tabular}{cc}
      \hspace{0.2cm} Skill 3 Model & \hspace{1.3cm} Anomaly Detection 
      \\
  \end{tabular}
   \includegraphics[scale=0.965]{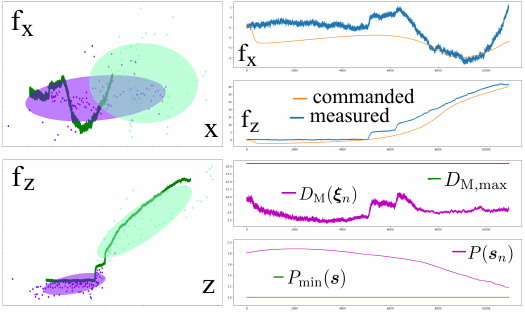}
  \caption{Nominal execution of skill 3. The relation between $f_x$ and $x$, as well as $f_z$ and $z$ for skill 3, is illustrated by the 2D excerpts of the GMM for motion- and force generation (left). As seen on the left, the measured forces during the execution $f_x$ and $f_z$ are within the expected constraint region for the measured EEF pose. The deviation between the commanded- and measured forces (upper right) is in a tolerable region, which is why the computed Mahalanobis distance $D_{\mathrm{M}}(\bm{\xi}_{n})$ does not exceed the upper boundary $D_{\mathrm{M, max}}$. Since the robot operates in a region of the training data with low epistemic uncertainty, the approach can confidently predict anomalies (see (\ref{eq:P_C}) and lower right figure).}
  \label{fig:Box_grasping_No_Anomaly}
\end{figure}

\subsubsection{Unsupervised Anomaly Detection}
\label{sec:anomaly_detection_box_grasping}
\begin{figure*}[thpb]
  \centering
  \begin{tabular}{p{0.25\textwidth}p{0.25\textwidth}p{0.25\textwidth}p{0.25\textwidth}}
     \hspace{0.6cm} I) push against EEF & \hspace{0.4cm} II) slide locked & \hspace{-1.05cm} III) pull away during contact & \hspace{-0.3cm} IV) box missed\\
    \end{tabular}
   \includegraphics[scale=0.955]{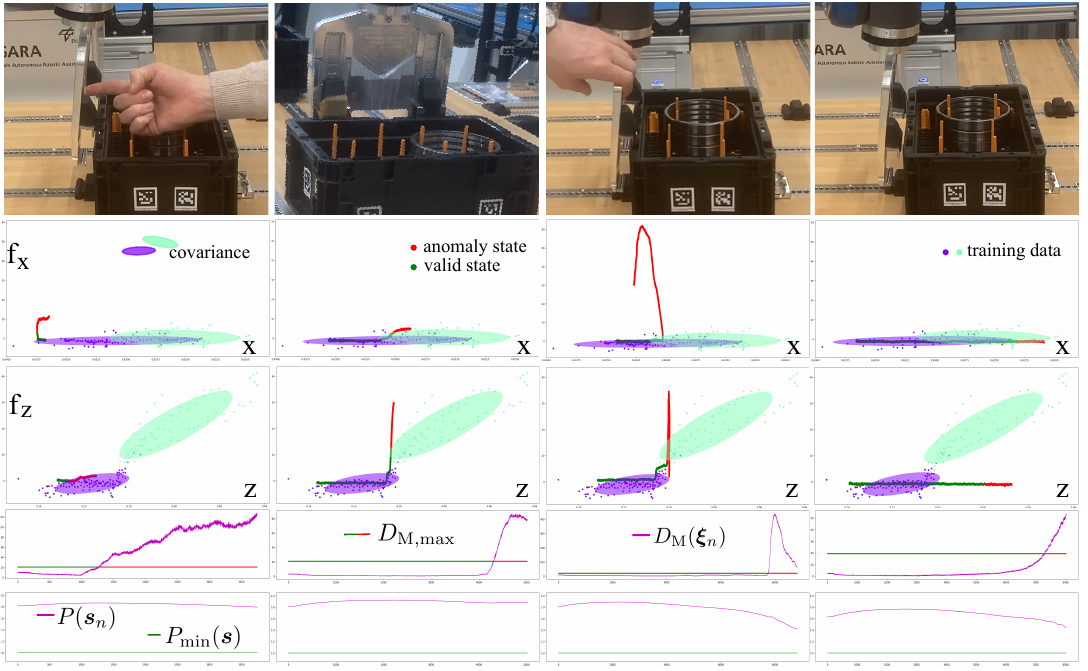}
  \caption{When interfering with the task, our method consistently detects anomalies before potentially dangerous situations can occur. As shown in rows 2 and 3, when the measured forces $f_x$ and $f_z$ are outside the expected range with respect to the measured EEF pose, our unsupervised anomaly detection approach classifies the measurements as anomalous (red samples). As more variance in the force $f_z$ is present in the training data for the phase where the robot tensions springs 1 (turquoise region), the anomaly detection is less sensitive towards deviations in $f_z$ in that phase during the execution. If the computed Mahalanobis distance $D_{\mathrm{M}}(\bm{\xi}_{n})$ exceeds the upper boundary $D_{\mathrm{M, max}}$ for more that 300~ms and the robot is within the training region with low epistemic uncertainty (lower two rows), an anomaly is confidently detected. Detailed videos of all anomaly cases are provided in Extension~3.}
  \label{fig:Box_grasping_Anomalies}
\end{figure*}

The refined skill sequence can now be executed on the robot with active anomaly detection. To test the anomaly detection and recovery capability of our approach, we simulate four different anomalies during the third skill of the task, depicted in Fig.~\ref{fig:Box_grasping_Anomalies}. The anomalies are I) pushing against the end-effector before tensioning springs 1, II) prematurely locked linear slides, III) pulling the gripper away from the box during tensioning springs 1, and IV) missed contact between the slides and the box. Anomalies I) and III) simulate user interference with the task, II) simulates a hardware defect, and IV) a perception error, that causes a wrongly predicted box configuration resulting in a gripper offset relative to the box.

As shown in Fig.~\ref{fig:Box_grasping_Anomalies}, our approach successfully detects all anomalies before a potentially dangerous situation can occur. The second and third row of Fig.~\ref{fig:Box_grasping_Anomalies} show the measured force in x and z-direction over the EEF position in the same direction as well as the corresponding training data for the skill. The expected force region with respect to the EEF pose is depicted by the ellipsoids representing the covariance of the training data. As shown on the left in Fig~\ref{fig:Box_grasping_No_Anomaly}, the measured forces during the execution are expected to lie within the region of the training data. If the measured and commanded forces do not deviate more than expected, the Mahalanobis distance computed with (\ref{eq:anomaly_mahal_dist}) stays below the skill's anomaly threshold (see right side of Fig.~\ref{fig:Box_grasping_No_Anomaly}). However, if the measured forces deviate from their expected region, and the Mahalanobis distance successively exceeds the threshold for more than 300~ms, an anomaly is triggered (see second last row in Fig.~\ref{fig:Box_grasping_Anomalies}). Since the variation in the training data is smaller in the first part of the skill, represented by the purple ellipsoids, the anomaly detection is more sensitive to deviations in that phase. As shown in the bottom row of Fig.~\ref{fig:Box_grasping_Anomalies}, the model can confidently predict anomalies for the measured EEF poses, since condition (\ref{eq:P_C}) is continuously met.

\paragraph{Quantitative Evaluation}

We evaluate the performance of our anomaly detection against several state-of-the-art baselines. The characteristics of the compared approaches are summarized in Table~\ref{tab:characteristics_anomaly_baselines}. ConditionNET \citep{sliwowski2024conditionnet} and FinoNET \citep{inceoglu2021fino} are supervised anomaly detection approaches, trained on successful and unsuccessful videos of the task execution, whereas our unsupervised anomaly detection approach is only trained on the end-effector poses and contact forces recorded from three successful user demonstrations and three successful robot executions of the task.

\textbf{ConditionNET} \citep{sliwowski2024conditionnet} is a vision-language model designed to learn the preconditions and effects of skills. It frames anomaly detection as a state prediction problem, where given an image and a natural language description of the skill, the model classifies whether the image represents a precondition, effect, or neither. An anomaly is detected if the predicted state does not match the actual state of the skill.

\begin{table*}[thpb]
\centering
\caption{Characteristics of the compared anomaly detection approaches during the training and prediction phase.}
\label{tab:characteristics_anomaly_baselines}
\begin{tabular}{p{1.38cm}|p{1.7cm}p{2.8cm}p{2.9cm}|p{3.3cm}p{1.4cm}p{1cm}}
\toprule
 & \multicolumn{3}{|c|}{Training}  & \multicolumn{3}{|c}{Prediction} \\
Method &  Setting &  Data &  Examples &  Data & Evaluation &  Resp t \\
\midrule
GMR\color{gray}{(our)}  &unsupervised&$\bm{p}_{\mathrm{EEF}}, \bm{q}_{\mathrm{EEF}}, \bm{F}_{\mathrm{ext}}$&6 success examples&skill + meas. $\bm{p},\bm{q},\bm{F}$&per frame&5ms\\
GMRwo                   &unsupervised&$\bm{p}_{\mathrm{EEF}}, \bm{q}_{\mathrm{EEF}}, \bm{F}_{\mathrm{ext}}$&6 success examples&meas. $\bm{p},\bm{q},\bm{F}$&per frame&5ms\\
VLM QA                  &pretrained&-&-&nl prompt, latest frames&per frame&10-15s\\
CondNET                 &supervised&annot video frames&30 success, 49 anom&video frame, skill phase&per frame   &20ms\\
FINO                    &supervised&annot video frames&30 success, 49 anom&8 video frames (skill)&per skill   &40ms\\

\bottomrule
\end{tabular}
\end{table*}

\begin{table*}[thpb]
\centering
\caption{Evaluation of the frame-wise detection performance of our GMR-based anomaly detection approach against the baselines ConditionNET \citep{sliwowski2024conditionnet}, VLM-based CoT-QA \citep{agia2024unpacking}, GMR without segmentation. For all cases except the "box missed" anomaly, our approach shows the best detection performance.}
\label{tab:anomaly_frame_evaluation}
\begin{adjustbox}{max width=\textwidth}
\begin{tabular}{p{1.38cm}|p{0.389cm}p{0.388cm}p{0.388cm}p{0.388cm}m{0.34cm}|p{0.388cm}p{0.388cm}p{0.388cm}p{0.388cm}p{0.34cm}|p{0.388cm}p{0.388cm}p{0.388cm}p{0.388cm}p{0.385cm}|p{0.388cm}p{0.388cm}p{0.388cm}p{0.388cm}p{0.34cm}}
\toprule
 & \multicolumn{5}{|c|}{push against EEF}  & \multicolumn{5}{|c|}{slide locked} & \multicolumn{5}{|c|}{pull away during contact} & \multicolumn{5}{|c}{box missed} \\
Method &  Acc &  Pre &  Rec &  F1 & Del &  Acc &  Pre &  Rec &  F1 & Del &  Acc &  Pre &  Rec &  F1 & Del &  Acc &  Pre &  Rec &  F1 & Del \\
\midrule
GMR\color{gray}{(our)} &\cellcolor{green!20}97.6&\cellcolor{green!20}100&\cellcolor{green!20}96.6&\cellcolor{green!20}98.3&\cellcolor{green!20}0.2  &\cellcolor{green!20}97.6&\cellcolor{green!20}100&\cellcolor{green!20}96.0&\cellcolor{green!20}98.0&\cellcolor{green!20}0.3   &\cellcolor{green!20}98.5&\cellcolor{green!20}97.4&\cellcolor{green!20}99.5&\cellcolor{green!20}98.4&\cellcolor{green!20}0.04   &60.0&100&34.6&51.5&4.3   \\
CondNET   &68.6&98.9&55.1&70.8&1.2   &64.4&98.6&50.2&66.5&3.0   &94.0&88.8&94.3&91.5&0.2   &\cellcolor{green!20}66.9&\cellcolor{green!20}95.3&\cellcolor{green!20}53.8&\cellcolor{green!20}68.8&\cellcolor{green!20}2.2\\
VLM QA   &38.0&85.7&11.2&19.9&3.4   &64.0&91.7&53.2&67.3&5.7   &75.0&72.4&38.9&50.6&2.7   &53.9&100&31.2&47.5&2.3\\
GMRwo   &50.8&100&29.5&45.5&3.4   &41.2&0&0&0&8.3   &61.9&95.5&17.7&29.8&0.1      &38.8&0&0&0&6.6\\

\bottomrule
\end{tabular}
\end{adjustbox}
\end{table*}

\textbf{FinoNET} \citep{inceoglu2021fino} is a deep-neural network-based model to detect manipulation failures by classifying an observed skill recording as \textit{success} or \textit{failure}. Four frames from the beginning and four frames from the end phase of the skill are randomly sampled and used as input for the classification. The model provides a success assessment after observing the complete skill. 

\textbf{VLM CoT-QA} A LVLM (Pixtral-12B-2409) \citep{agrawal2024pixtral} is queried to assess the task progress and detect anomalies based on the skill's video frames up to the current time step and a natural language prompt. The anomaly detection problem is framed as a chain-of-thought (CoT), video question answering (QA) task. We reduce the time horizon of the anomaly detection to the performed skill and design the VLM prompts according to \cite{agia2024unpacking}, which include a comprehensive description of the current skill, an explanation of the VLM's role as an anomalies detector, and the remaining time until expected skill completion.

\textbf{GMR without segmentation} We use the same GMR-based probabilistic anomaly detection mechanism as in our proposed approach. However, the task is not segmented into skills, and the entire training data is encoded as one GMM.

\paragraph{Dataset and Metrics} We collected a dataset that contains videos, robot end-effector trajectories, and contact forces from  22 successful and 40 unsuccessful executions of the box grasping task. The unsuccessful executions contain 10 instances for each anomaly type. For ConditionNET and FinoNet, the video frames corresponding to each skill are segmented into pre-, core-, and effect phases. Each video is annotated with natural language descriptions of the performed skills, a temporal segmentation mask for the three phases of every skill, and a success label. We partition the dataset into training (70\%) and validation (30\%) sets while preserving the distribution of successful and unsuccessful executions. We report frame-wise accuracy, precision, recall, F1 scores, and mean anomaly detection delay for the models performing online anomaly detection at each time step (Table~\ref{tab:anomaly_frame_evaluation}). The mean detection delay is the average time in seconds between the first occurrence of an anomaly and its detection by a given approach. This metric does not include model response time, which varies significantly across models (see Table~\ref{tab:characteristics_anomaly_baselines}). Table~\ref{tab:anomaly_task_evaluation} shows the average task assessment accuracy over executions, indicating whether a model correctly detected an anomaly at any point during the complete execution.

\begin{table}[thpb]
\centering
\caption{Average prediction accuracy over complete task executions for the four anomaly cases and the successful case.}
\label{tab:anomaly_task_evaluation}
\begin{tabular}{p{1.38cm}|ccccc}
\toprule
 & \multicolumn{5}{|c}{Prediction Accuracy} \\
Method &  I &  II &  III & IV & no \\
\midrule
GMR\color{gray}{(our)}  &\cellcolor{green!20}100&\cellcolor{green!20}100&\cellcolor{green!20}100&\cellcolor{green!20}100&\cellcolor{green!20}100\\
GMRwo                   &100&0&100&0&0\\
VLM QA                  &50.0&66.7&75.0&66.7&44.4\\
CondNET                 &100&100&100&100&33.3\\
FINO                    &100&100&80.0&100&66.7\\

\bottomrule
\end{tabular}
\end{table}

\paragraph{Results} As shown in Table~\ref{tab:anomaly_frame_evaluation}, our approach outperforms all other online detection baselines in frame-wise prediction performance and detection delay for anomaly cases I–III. Unlike other methods, our detector identifies subtle force deviations before anomalies become visually observable. For the "box missed" case, ConditionNET achieves better detection performance, as this anomaly is visually observable before anomalous force readings occur. Table~\ref{tab:anomaly_task_evaluation} further demonstrates that our approach is the only one that confidently detects all anomalies without triggering false positives during successful executions. Methods relying on vision are sensitive to camera viewpoints and degrade in performance under occlusions. GMR without segmentation performs the worst in this setting, as its mixture components are unevenly distributed across skills, and the anomaly detection threshold in (\ref{eq:P_A}) remains fixed throughout the task, reducing the overall sensitivity to deviations. The VLM CoT-QA baseline correctly describes observed video frames and focuses on relevant questions for anomaly detection. However, it often struggles to determine whether a described situation constitutes an anomaly. Most detected anomalies with this method result from exceeding the skill’s time limit. Even in such cases, the VLM fails to maintain consistent predictions across multiple time steps.

\subsubsection{Task Decision Teaching Sequence}
\label{sec:box_grasping_task_decision_sequence}
Finally, to autonomously recover from the detected anomalies, the anomaly cases need to be classified to select the appropriate recovery behaviors. For this, we utilize the anomalous observations $\{\bm{\xi}_n\}_{n=1}^{\epsilon}$ of EEF velocities and contact forces (red samples in Fig.~\ref{fig:Box_grasping_Anomalies}) of each anomaly case to learn a Support Vector Machine with a sliding time window of length 100. We constantly update the supervised anomaly classifier with new training data to improve the classification of known anomalies and to extend the model with new anomaly classes as they occur.
\begin{figure}[thpb]
  \centering
   \includegraphics[scale=0.965]{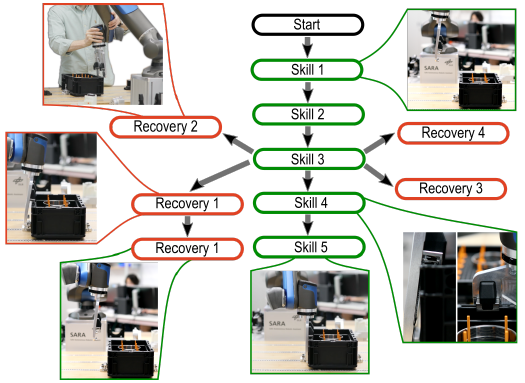}
  \caption{The Task Graph for the box grasping and locking task, including recovery behaviors for the different anomalies I) - IV) of skill 3. After the initial demos, the TG consists of the intended task flow skills in green. As new anomalies are identified, the TG can be incrementally extended with recovery behaviors. They recover from anomalies, such that the robot can continue with the intended task flow.}
  \label{fig:Box_Grapsing_Task_Graph}
\end{figure}

If a new anomaly is detected with Algorithm~\ref{alg:determine_anomaly_label}, the task graph in Fig.~\ref{fig:Box_Grapsing_Task_Graph} is extended with a new recovery behavior. Just like the initial task flow skills, recovery behaviors are learned from demonstration, however, they are appended to the skill in the task graph during which the anomaly was detected. If, during the execution, the recovery behavior's final subgoal is reached, we assume that the robot can continue with the nominal execution of the task (green skills in Fig.~\ref{fig:Box_Grapsing_Task_Graph}). To choose the next skill after a recovery behavior, we select the nominal skill whose low-level motion model according to Sec.~\ref{sec:motion-generation-anomaly-detection} maximizes $P(\bm{s}_n) = \sum_{e=1}^{E}\pi_e\mathcal{N}(\bm{s}_n \vert \bm{\mu}_e^s, \bm{\Sigma}_e^{ss})$ for the measured EEF pose $\bm{s}_n$, i.e. the skill with the lowest epistemic uncertainty for generating an action. This skill is best suited to continue with the execution from the current robot configuration.

\subsection{Discussion and Limitations}
\label{sec:discussion_limitation}
For input far away from the training data, the GMM/GMR-based motion generation approach struggles to produce meaningful output commands, which may instead converge to spurious attractors. To increase the generalization capabilities of our framework in areas beyond the observed EEF poses during the demonstration, we propose to distinguish the skills in contact and free-space motion skills using e.g. the classification proposed in \cite{eiband2023online}. For free-space motion skills, the only aim is to reach their subgoal configurations, which are by design within the training region of the next skill. A motion planner can thus be used to generate a collision-free trajectory to the subgoal region, from which a contact skill can continue with the execution. Dynamical Systems learned from user demonstrations can furthermore suffer from local minima in absolute velocity in the training data, which can cause the robot to get stuck in these regions during the execution. States during the demonstration, where low EEF velocities close or equal to zero are recorded, are more likely to be important states where high precision in the EEF configuration is required. Our approach can identify these states as subgoals. If such a subgoal is reached during the execution, our system transitions to the next skill in the task graph, using a new low-level skill model that can escape from the local minimum of the previous skill.

Another advantage when distinguishing between contact and free-space motions concerns safety during the execution. Since unintended contacts with the environment usually trigger a collision stop of the robot, this safety feature needs to be deactivated during contact tasks to avoid false positive collision detection. However, our anomaly detection approach still registers unintended forces that exceed the expected process forces and thus increases user safety in contact situations. When commanding a robot using impedance control, the presence of unmodelled contact forces and torques causes a stiffness-dependent offset between the commanded and measured EEF pose of the robot. However, the box-grasping mechanism requires precise EEF configurations to complete the task. To compensate for this offset, we actively command the configuration-dependent force in every cycle needed to counteract the force resulting from the springs of the gripper.

Since we assume, that the features of every skill follow a multivariate Gaussian distribution, we are limited to inferring linear correlations between the features when segmenting skills with our BNG-IRL approach. We demonstrated that we can solve challenging tasks with this approach, however, there may exist tasks, where nonlinear relations between the features play an important role. Additionally, when computing the epistemic uncertainty for step 1 of the anomaly detection, the likelihood for states in the beginning and end of the skills that lie closer to the boundary of the training data are closer to the confidence threshold (\ref{eq:P_C}). This means that the anomaly detection is by design less confident for states in the beginning and the end of the skill. Lastly, the selection of a nominal skill after a recovery behavior does not consider high-level or semantic task information. Instead, the skill best suited to generate a motion based on the current EEF configuration is chosen. However, incorporating semantic information could be beneficial in narrowing down candidate skills for transitions. Similarly, augmenting recovery behaviors with semantic information could enable their reuse across different skills, allowing for automatic recovery from similar anomalies without explicitly demonstrating the recovery behavior. An automatic evaluation of whether the current situation meets the precondition of another skill is proposed by \cite{sliwowski2024conditionnet} to select an appropriate recovery behavior from known skills. Instead of demonstrating a new recovery behavior, users could assess whether an existing skill in the task graph is suitable for recovery and select it via a user interface, similar to the proposed approach in our previous works \citep{willibald2020collaborative, eiband2023collaborative}. This would allow the system to gradually add new connections between existing skills, without the need for semantic annotation. Future research could explore those ideas to reuse existing skills to improve automatic recovery from anomalies.
\section{Conclusion}
\label{sec:Conclusion}
We introduced a novel incremental learning framework designed for complex contact-based tasks composed of multiple sequential sub-steps, which are challenging to learn with existing LfD methods. The initial task demonstrations are segmented using our unsupervised BNG-IRL segmentation approach to learn a nominal task model. Our framework facilitates incremental learning at both high and low levels, simplifying the teaching process for users by eliminating the need to anticipate anomalies or new scenarios. Our unsupervised anomaly detection technique identifies deviations from the intended task execution without prior knowledge of potential anomaly cases. Only if the approach detects a new anomaly, the user is queried to provide a recovery behavior, which can then be used to automatically recover from that anomaly in the future. Additionally, the low-level model is updated with new training data collected during execution to continuously refine the existing skills.

Our segmentation approach shows improved performance over four baseline methods by combining the advantages of subgoal-based inverse reinforcement learning with probabilistic feature clustering in one model. Furthermore, we demonstrated the applicability of the framework with delicate tasks performed on two robotic systems. Notably, only three demonstrations were needed to learn a robust initial model of the box grasping task, capable of identifying several different anomaly cases based on the expected contact force depending on the robot-environment interaction. Our unsupervised anomaly detection approach outperforms all other supervised visual anomaly detection baselines in three out of four anomaly detection cases and is the only one to confidently detect all anomalies, while not triggering any false positive detection during successful executions.

\begin{acks}
This work has been supported by Helmholtz Association and by the DLR internal projects "Factory of the Future Extended" and ASPIRO. The authors would like to thank Daniel Sliwowski for his support with the experimental evaluation.
\end{acks}

\begin{dci}
The authors declared no potential conflicts of interest with respect to the research, authorship, and/or publication of this article
\end{dci}

\begin{funding}
The authors received no financial support for the research, authorship, and/or publication of this article.
\end{funding}

\begin{sm}
Index to Multimedia Extensions
\begin{table}[h]
    \begin{tabular}{lll}
       1  &  Video & Video to experiment in Sec.\ref{sec:Box_Pushing_Exp}\\
       2  & Video & Video to experiment in Sec.\ref{sec:Manipulation_Real_World}\\
       3  & Video & Video to experiment in Sec.\ref{sec:box_grasping}\\
    \end{tabular}

\end{table}
\end{sm}

\bibliographystyle{SageH}
\bibliography{main}

\end{document}